\documentclass[10pt,twocolumn,letterpaper]{article}
\usepackage{standalone}

\usepackage[accsupp]{axessibility}
\usepackage{wacv}
\usepackage{times}
\usepackage{epsfig}
\usepackage{graphicx}
\usepackage{amsmath}
\usepackage{amssymb}
\usepackage{booktabs}
\usepackage{multirow}
\usepackage{cuted} % for text box across two columns
\usepackage[utf8]{inputenc} % for german

\graphicspath{{figs}}
\usepackage[pdftex,dvipsnames]{xcolor}  % Coloured text etc.

\usepackage{lipsum}  % only for dummy text! To be deleted !
\usepackage[numbers,sort,compress]{natbib} % reference in increasing order
\usepackage{pifont} % for checkmark and cross symbol
\usepackage[abs]{overpic}
\usepackage{tikz}
\usepackage{float}
\usepackage{tikz-imagelabels}
\imagelabelset{
coarse grid color = red,
fine grid color = gray,
image label font =\small,%\sffamily\bfseries\small,
image label distance = 0.0mm,
image label back = white,
image label text = black,
coordinate label font = \small, %\sffamily\bfseries
coordinate label distance = 0.0mm,
coordinate label back = white,
coordinate label text = black,
annotation font = \normalsize, %\normalfont\small,
arrow distance = 1.5mm,
border thickness = 0.0pt,
arrow thickness = 0.4pt,
tip size = 0.0mm,
outer dist = 0.0cm,
}

\usepackage[pagebackref=true,breaklinks=true,colorlinks,bookmarks=false,hypertexnames=false]{hyperref}
\usepackage[capitalize]{cleveref}
\crefname{section}{Sec.}{Secs.}
\Crefname{section}{Section}{Sections}
\Crefname{table}{Table}{Tabs.}
\crefname{table}{Table}{Tabs.}
\newcommand{\red}[1]{\textcolor{red}{#1}}
\newcommand{\newchange}[1]{\textcolor{black}{#1}}

\renewcommand{\paragraph}[2][.]{\vspace{4pt}\noindent{\bf #2#1}}
\newcommand{\xmark}{\ding{55}}
\newcommand{\cmark}{\ding{51}}

\newcommand*{\newcite}[1]{~\cite{#1}}
\newcommand{\norm}[1]{\left\lVert#1\right\rVert}

\newcommand{\hrnet}{\newcite{wang2020hrnet}}
\newcommand{\segformer}{\newcite{xie2021segformer}}
\newcommand{\deeplabvtwo}{\newcite{chen2017deeplabv2}}
\newcommand{\acdc}{\newcite{sakaridis2021acdc}}
\newcommand{\ourstyle}{$\mathrm{ISSA}$}
\newcommand{\ourstylebf}{{$\mathbf{ISSA}$}}

\newcommand{\wplus}{$\mathcal{W^+}$}
\usepackage{xargs}  
\usepackage[disable]{todonotes}
\newcommandx{\com}[2][1=]{\todo[linecolor=Plum,backgroundcolor=Plum!25,bordercolor=Plum,#1]{#2}}
\newcommandx{\changed}[2][1=]{\todo[linecolor=Blue,backgroundcolor=Blue!25,bordercolor=Blue,#1]{#2}}
\newcommand{\commentOut}[1]{}

\def\wacvPaperID{1455} % Enter the WACV Paper ID here

\wacvalgorithmstrack   % Uncomment this line if you are submitting to the Algorithms Track.
\wacvfinalcopy % *** Uncomment this line for the final submission

\ifwacvfinal
\usepackage[breaklinks=true,bookmarks=false,hypertexnames=false]{hyperref}
\else
\usepackage[pagebackref=true,breaklinks=true,colorlinks,bookmarks=false,hypertexnames=false]{hyperref}
\fi

\makeatother
\begin{document}

%%%%%%%%% TITLE
%ISSA
\title{Intra-Source Style Augmentation for Improved Domain Generalization}
%ISSA

\author{Yumeng Li$^{1,2}$\hspace{1.7em}
Dan Zhang$^{1,4}$\hspace{1.7em}
Margret Keuper$^{2,3}$\hspace{1.7em}
Anna Khoreva$^{1,4}$\\
$^1$Bosch Center for AI
\hspace{0.3em}$^2$ University of Siegen 
\hspace{0.3em}$^3$MPI for Informatics
\hspace{0.3em}$^4$University of T\"ubingen 
\\
{\tt\small \{yumeng.li, dan.zhang2, anna.khoreva\}@de.bosch.com \ 
{margret.keuper}@uni-siegen.de
}
}

\maketitle
\thispagestyle{empty}

%%%%%%%%% ABSTRACT
\begin{abstract}
The generalization with respect to domain shifts, as they frequently appear in applications such as autonomous driving, is one of the remaining big challenges  for deep learning models. Therefore, we propose an intra-source style augmentation ({\ourstyle}) method to improve domain generalization in semantic segmentation. Our method is based on a novel masked noise encoder for StyleGAN2 inversion. The model learns to faithfully reconstruct the image preserving its semantic layout through noise prediction. Random masking of the estimated noise enables the style mixing capability of our model, i.e.~it allows to alter the global appearance without affecting the semantic layout of an image. Using the proposed masked noise encoder to randomize style and content combinations in the training set, {\ourstyle} effectively increases the diversity of training data and reduces spurious correlation. As a result, we achieve up to $12.4\%$ mIoU improvements on driving-scene semantic segmentation under different types of data shifts, i.e., changing geographic locations, adverse weather conditions, and day to night. {\ourstyle} is model-agnostic and straightforwardly applicable with CNNs and Transformers. 
It is also complementary to other domain generalization techniques, e.g., it improves the recent state-of-the-art solution RobustNet by $3\%$ mIoU in Cityscapes to Dark Z\"urich. Code is available at \url{https://github.com/boschresearch/ISSA}.
\vspace{-0.6em}
\end{abstract}

%%%%%%%%% BODY TEXT
\section{Introduction}
\begin{figure}[t]
    \begin{centering}
    \setlength{\tabcolsep}{0.0em}
    \renewcommand{\arraystretch}{0}
    \par\end{centering}
    \begin{centering}
    \vspace{-0.5em}
    \hfill{}%
	\begin{tabular}{@{}c@{}c}
        \centering
		Unseen domain (snow) & Ground truth \tabularnewline
	    \includegraphics[width=0.47\linewidth]{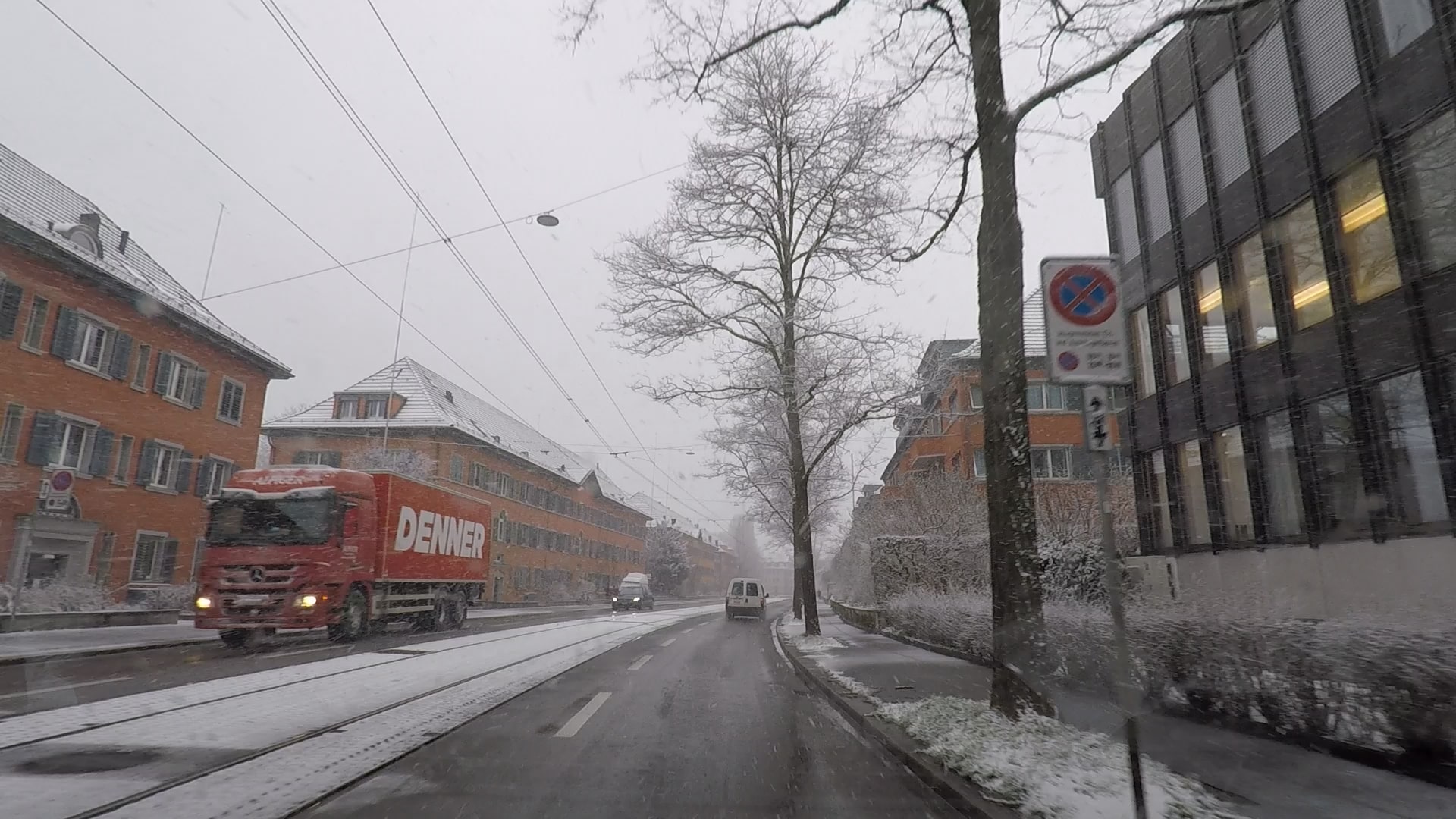} 
		&{\footnotesize{}} 
		\begin{tikzpicture}
            \node [
	        above right,
	        inner sep=0] (image) at (0,0) {\includegraphics[width=0.47\linewidth]{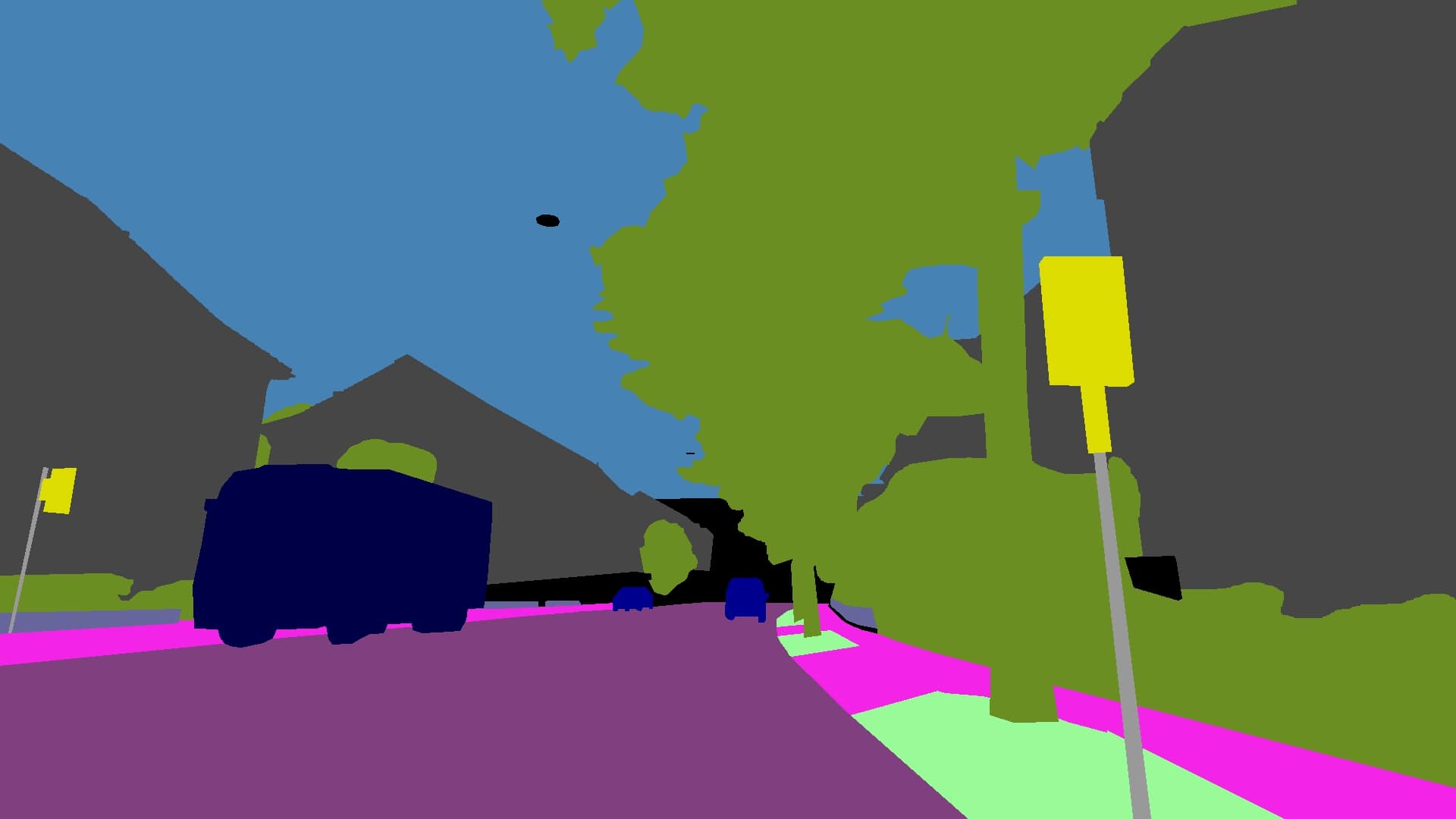} };
            \begin{scope}[
            x={($0.1*(image.south east)$)},
            y={($0.1*(image.north west)$)}]
            \draw[thick,green] (7,0.5) rectangle (9.8,7.2);
            \draw[thick,green] (1.1,2) rectangle (3.6,4.6);
        \end{scope}
    \end{tikzpicture}
		\tabularnewline
		Baseline & Ours \tabularnewline
	\begin{tikzpicture}
            \node [
	        above right,
	        inner sep=0] (image) at (0,0) {\includegraphics[width=0.47\linewidth]{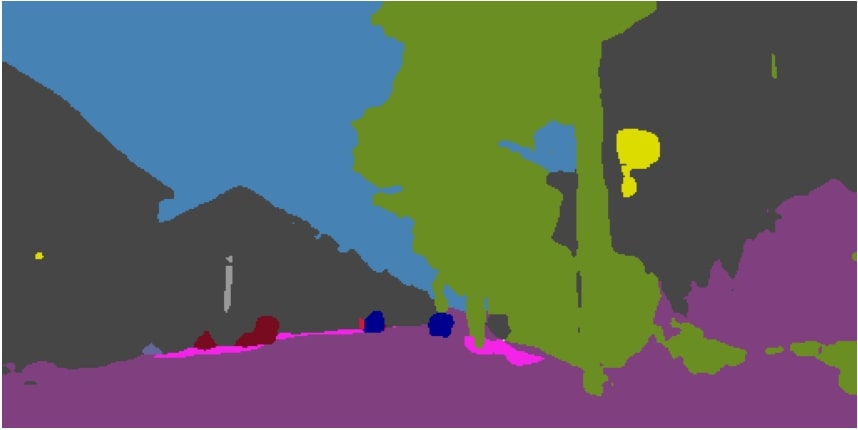}};
            \begin{scope}[
            x={($0.1*(image.south east)$)},
            y={($0.1*(image.north west)$)}]
            \draw[thick,red] (7,0.5) rectangle (9.8,7.2);
            \draw[thick,red] (1.1,2) rectangle (3.6,4.6);
        \end{scope}
    \end{tikzpicture} 
		&{\footnotesize{}} 
    \begin{tikzpicture}
            \node [
	        above right,
	        inner sep=0] (image) at (0,0) {\includegraphics[width=0.47\linewidth]{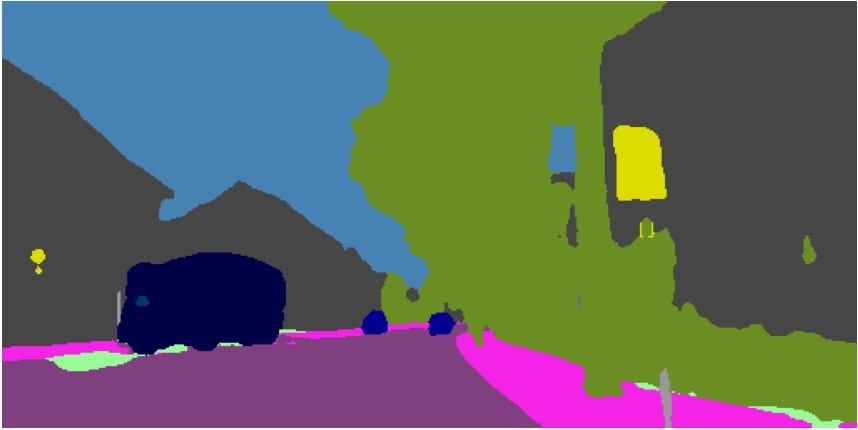}};
            \begin{scope}[
            x={($0.1*(image.south east)$)},
            y={($0.1*(image.north west)$)}]
            \draw[thick,green] (7,0.5) rectangle (9.8,7.2);
            \draw[thick,green] (1.1,1.5) rectangle (3.6,4.6);
        \end{scope}
    \end{tikzpicture}	
		\end{tabular}
\hfill{}
\par\end{centering}
\caption{
Semantic segmentation results of HRNet{\hrnet} on unseen domain (snow), trained on Cityscapes\newcite{cordts2016cityscapes} and tested on ACDC\newcite{sakaridis2021acdc}. The model trained with our {\ourstyle} can successfully segment the truck, while the baseline model fails completely.}
\label{fig:intro-semseg}
\vspace{-1.0em}
\end{figure}

The varying environment with potentially diverse illumination and adverse weather conditions makes challenging the deployment of deep learning models in an open-world\newcite{sakaridis2021acdc,zhang2021autonomous}.  
Therefore, improving the generalization capability of neural networks is crucial for safety-critical applications such as autonomous driving (see for example \cref{fig:intro-semseg}). While generally the target domains can be inaccessible or unpredictable at training time, it is important to train a generalizable model, based on the known (source) domain, which may offer only a limited or biased view of the real world\newcite{burton2017safety,shafaei2018uncertainty}.

Diversity of the training data is considered to play an important role for domain generalization, including natural distribution shifts\newcite{taori2020robustness}. Many existing works assume that multiple source domains are accessible during training\newcite{hu2020mda,li2018learning,balaji2018metareg,li2018domain,li2020domain,jin2020feature,zhou2020optimaltransport}. For instance, 
Li~\etal\newcite{li2018learning} %applied 
divide source domains into meta-source and meta-target to simulate domain shift for learning; 
Hu~\etal\newcite{hu2020mda} propose multi-domain discriminant analysis to learn a domain-invariant feature transformation.
%meta-learning to better generalize to unseen domains. %, where source domains are divided into meta-source and meta-target domains to simulate domain shift.
However, for pixel-level prediction tasks such as semantic segmentation, collecting diverse training data involves a tedious and costly annotation process\newcite{caesar2018coco}. Therefore, improving generalization from a \emph{single source domain} is exceptionally compelling, particularly for semantic segmentation.
%of great interest, especially for semantic segmentation.

One pragmatic way to improve data diversity is by applying data augmentation. It has been widely adopted in solving different tasks, such as image classification\newcite{zhang2018mixup,zhou2021mixstyle,hendrycks2019augmix,verma2019manifold,hong2021stylemix}, GAN training with limited data\newcite{stylegan2ada,jiang2021deceivegan}, or pose estimation\newcite{peng2018humanpose,bin2020adv_pose,wang2021human}. %For dense prediction tasks such as semantic segmentation, the data collection and annotation process are prohibitive, increasing the need for data augmentation. %
%\newcite{zhang2018mixup,hendrycks2019augmix,verma2019manifold,hong2021stylemix,baek2021gridmix,hendrycks2018benchmarking,devries2017cutout}. This solution is model agnostic and can thus be reused across different network architectures. 
%Despite the larger need of data augmentation  for semantic segmentation due to its prohibitive labelling, it is essentially harder
%to apply augmentation on more diverse scene-centric dataset than single object-centric datasets.
%Only a few works\newcite{yun2019cutmix, hendrycks2018benchmarking} allow for direct application in such task. 
\iffalse
% Its effectiveness has been shown \commentOut{the primary use case of data augmentation approaches} on image-level prediction, e.g., image classification.\changed{less critized.}
% %on single-object-centric datasets and
% Nevertheless, only a few works\newcite{yun2019cutmix, hendrycks2018benchmarking} allow for direct application in pixel-level tasks such as semantic segmentation,  despite the larger need for semantic segmentation due to the prohibitive labelling.
\fi
%Yet, the generalization in such pixel-level tasks depends heavily on data augmentation techniques because of the high costs for manual annotation of new data.
One line of data augmentation techniques focuses on increasing the content diversity in the training set, such as geometric transformation (e.g., cropping or flipping), %MixUp\newcite{zhang2018mixup},
CutOut\newcite{devries2017cutout}, and CutMix\newcite{yun2019cutmix}. However, CutOut and CutMix are ineffective on natural domain shifts as reported in\newcite{taori2020robustness}. Style augmentation, on the other hand, only modifies the style - the non-semantic appearance such as texture and color of the image\newcite{gatys2016image} - while preserving the semantic content. %, while preserving the semantic content. %which is more effective on improving the robustness against domain shifts\newcite{jackson2019style}. , where style primarily refers to texture and color\newcite{gatys2016image}
By diversifying the style and content combinations, style augmentation can reduce overfitting to the style-content correlation in the training set, improving robustness against domain shifts. 
%Acquiring styles from extra datasets or directly from the target domain are not always possible at training. % can be acquired from extra datasets or directly from the target domain such as in [ref], which however are not always accessible at training. 
Hendrycks corruptions\newcite{hendrycks2018benchmarking} provide a wide range of synthetic styles, including weather conditions. However, they are not always realistic looking, thus being still far from resembling natural data shifts. %It is helpful to reduce the overfitting in the source domain, thus improving robustness against domain shifts. Hendrycks corruptions\newcite{hendrycks2018benchmarking} includes synthetic weather 
In this work, we propose an intra-source style augmentation (\ourstyle) strategy for semantic segmentation, aiming to improve the style diversity in the training set without extra labeling effort or using extra data sources. %Color jittering and Hendrycks corruptions\newcite{hendrycks2018benchmarking} increase the style diversity by synthesizing new yet non-realistic looking styles. 
%{\ourstyle} mixes up the styles of training samples in the source domain and delivers realistic looking data augmentations.  % also aim at diversifying the global image style. Different to them, we  %However, the augmented images are not necessarily natural-looking and are therefore often ineffective on natural distribution shifts\newcite{taori2020robustness}. %
%In this work, we explore an intra-source data augmentation strategy for semantic segmentation, such as to improve data diversity without extra labeling effort. %
% It utilizes a high-quality data generation model for realistic looking data augmentation, and exploits styles available in the training set. 
%We propose to close this gap by using a high-quality data generation model for realistic looking data augmentation with diverse styles, resulting in intra-source style augmentation (\ourstyle). %

Our augmentation technique is based on the inversion of StyleGAN2\newcite{stylegan2}, which is the state-of-the-art unconditional Generative Adversarial
Network (GAN) and thus ensures high quality and realism of synthetic samples. %
GAN inversion allows to encode a given image to latent variables, and thus facilitates faithful reconstruction with style mixing capability. To realize {\ourstyle}, we learn to separate semantic content from style information based on a single source domain. This allows to alter the style of an image while leaving the content unchanged. Specifically, we make use of the styles extracted within the source domain and mix them up. 
%\com{i think here we can make "intra-source" in the name more reflected. the way of alternating the style is to extract styles in the source domain and mix them up.} 
Thus, we can increase the data diversity and alleviate the spurious correlation %\newcite{wiles2021finegrain,schott2021visual} 
in the given training data.  %Empirically, we observed a semantic segmentation model trained on Cityscapes. On the same training set with style mixing between different training samples, the model suffers from ## mIoU drop, indicating ...
%Note that we exploit the styles within the source domain and mix them up, thus our method is dubbed as Intra-Source Style Augmentation(ISSA).

\begin{figure*}[t]
    \begin{centering}
    \setlength{\tabcolsep}{0.0em}
    \renewcommand{\arraystretch}{0}
    \par\end{centering}
    \begin{centering}
    \vspace{-0.5em}
    \hfill{}%
	\begin{tabular}{@{}c@{}c@{}c@{}c}
        \centering
		Input
		& pSp${}^\dagger$ & Feature-Style Encoder\newcite{feature_style} & Masked Noise Encoder (Ours) \tabularnewline
	\begin{tikzpicture}
            \node [
	        above right,
	        inner sep=0] (image) at (0,0) {\includegraphics[width=0.235\textwidth]{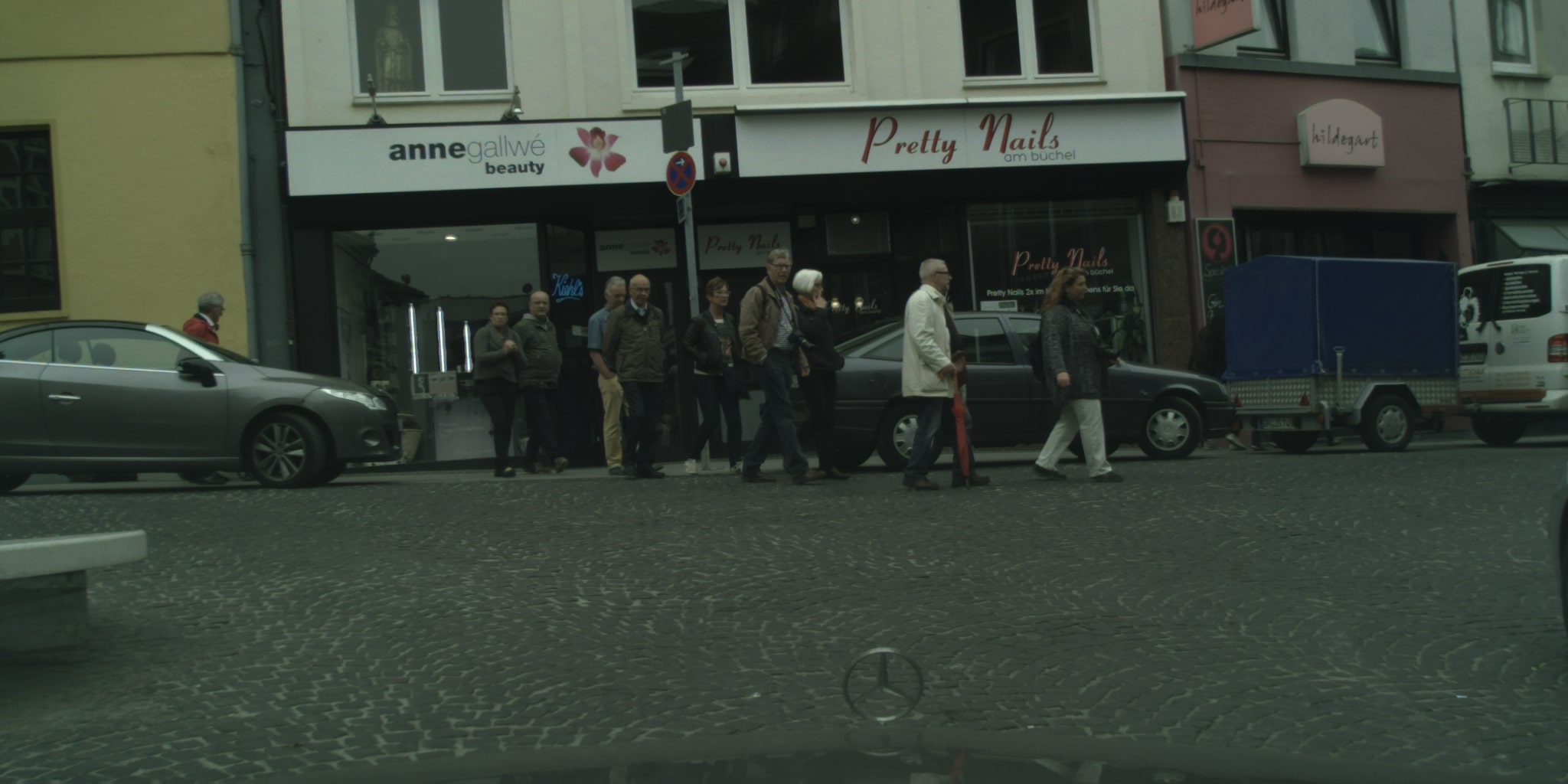}};
            \begin{scope}[
            x={($0.1*(image.south east)$)},
            y={($0.1*(image.north west)$)}]
            \draw[thick,green] (3,3.5) rectangle (9.5,8) ;
        \end{scope}
    \end{tikzpicture}
		& {\footnotesize{}}
	\begin{tikzpicture}
            \node [
	        above right,
	        inner sep=0] (image) at (0,0) {\includegraphics[width=0.235\textwidth]{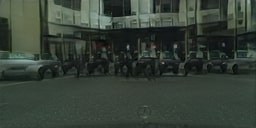} };
            \begin{scope}[
            x={($0.1*(image.south east)$)},
            y={($0.1*(image.north west)$)}]
            \draw[thick,red] (3,3.5) rectangle (9.5,8) ;
        \end{scope}
    \end{tikzpicture}	
		
		& {\footnotesize{}}
	\begin{tikzpicture}
            \node [
	        above right,
	        inner sep=0] (image) at (0,0) {\includegraphics[width=0.235\textwidth]{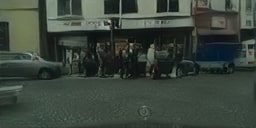} };
            \begin{scope}[
            x={($0.1*(image.south east)$)},
            y={($0.1*(image.north west)$)}]
            \draw[thick,red] (3,3.5) rectangle (9.5,8) ;
        \end{scope}
    \end{tikzpicture}
		& {\footnotesize{}}
	\begin{tikzpicture}
            \node [
	        above right,
	        inner sep=0] (image) at (0,0) {\includegraphics[width=0.235\textwidth]{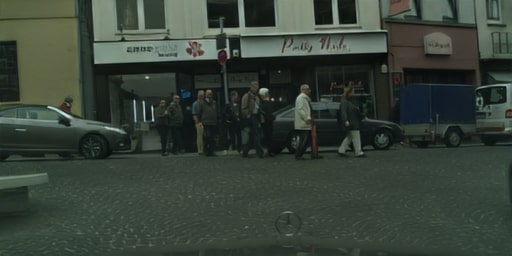}};
            \begin{scope}[
            x={($0.1*(image.south east)$)},
            y={($0.1*(image.north west)$)}]
            \draw[thick,green] (3,3.5) rectangle (9.5,8) ;
        \end{scope}
    \end{tikzpicture}
		\tabularnewline
		\begin{tikzpicture}
            \node [
	        above right,
	        inner sep=0] (image) at (0,0) {\includegraphics[width=0.235\textwidth]{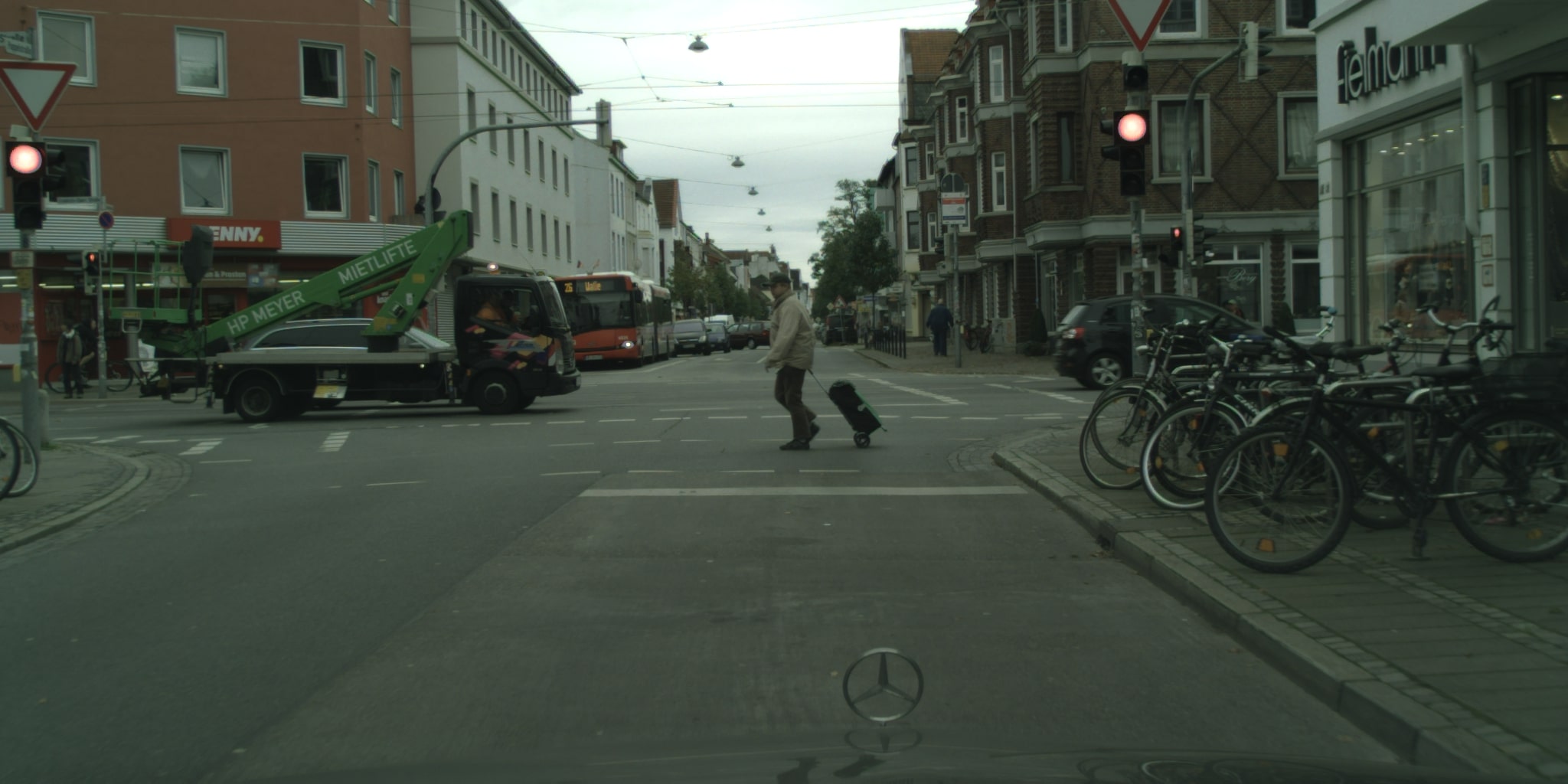}};
            \begin{scope}[
            x={($0.1*(image.south east)$)},
            y={($0.1*(image.north west)$)}]
            \draw[thick,green] (7,6) rectangle (7.5,10) ;
            \draw[thick,green] (0,6.5) rectangle (0.6,9.5) ;
            \draw[thick,green] (4.6,4) rectangle (5.9,7) ;
        \end{scope}
    \end{tikzpicture}
		& {\footnotesize{}}
	\begin{tikzpicture}
            \node [
	        above right,
	        inner sep=0] (image) at (0,0) {\includegraphics[width=0.235\textwidth]{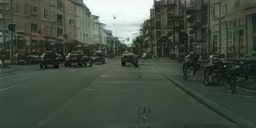} };
            \begin{scope}[
            x={($0.1*(image.south east)$)},
            y={($0.1*(image.north west)$)}]
            \draw[thick,red] (7,6) rectangle (7.5,10) ;
            \draw[thick,red] (0,6.5) rectangle (0.6,9.5) ;
            \draw[thick,red] (4.6,4) rectangle (5.9,7) ;
        \end{scope}
    \end{tikzpicture}
		& {\footnotesize{}}
    \begin{tikzpicture}
            \node [
	        above right,
	        inner sep=0] (image) at (0,0) {\includegraphics[width=0.235\textwidth]{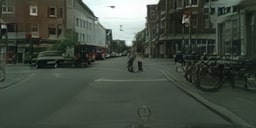}};
            \begin{scope}[
            x={($0.1*(image.south east)$)},
            y={($0.1*(image.north west)$)}]
            \draw[thick,red] (7,6) rectangle (7.5,10) ;
            \draw[thick,red] (0,6.5) rectangle (0.6,9.5) ;
            \draw[thick,red] (4.6,4) rectangle (5.9,7) ;
        \end{scope}
    \end{tikzpicture}
		& {\footnotesize{}}
	\begin{tikzpicture}
            \node [
	        above right,
	        inner sep=0] (image) at (0,0) {\includegraphics[width=0.235\textwidth]{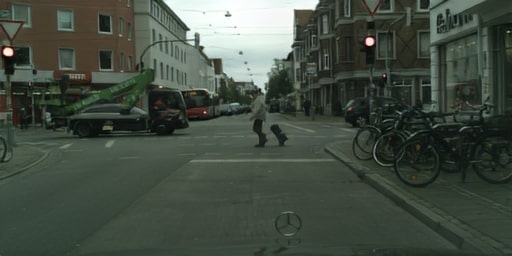}};
            \begin{scope}[
            x={($0.1*(image.south east)$)},
            y={($0.1*(image.north west)$)}]
            \draw[thick,green] (7,6) rectangle (7.5,10) ;
            \draw[thick,green] (0,6.5) rectangle (0.6,9.5) ;
            \draw[thick,green] (4.6,4) rectangle (5.9,7) ;
        \end{scope}
    \end{tikzpicture}
		\tabularnewline
	\begin{tikzpicture}
            \node [
	        above right,
	        inner sep=0] (image) at (0,0) {\includegraphics[width=0.235\textwidth]{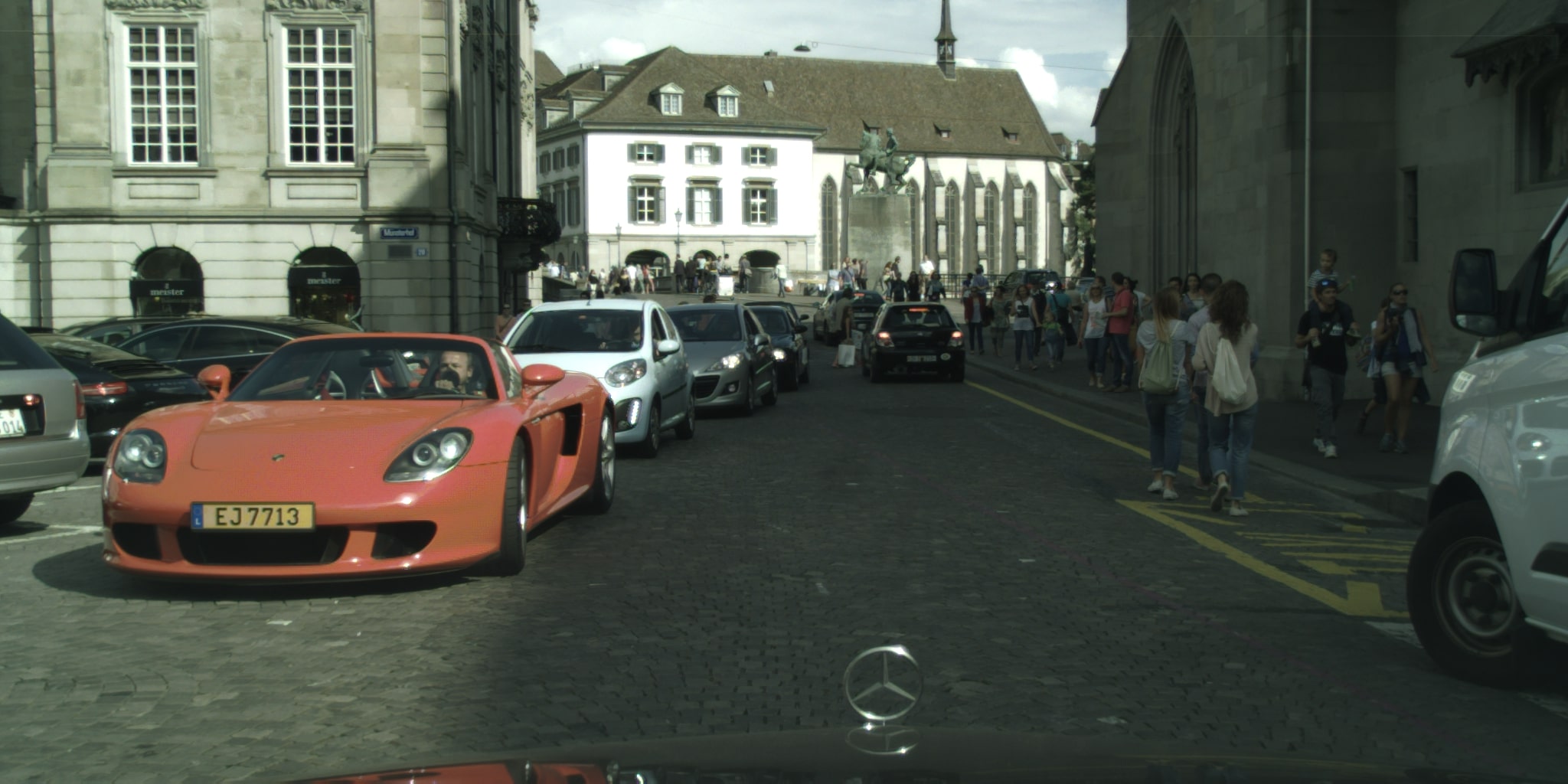}};
            \begin{scope}[
            x={($0.1*(image.south east)$)},
            y={($0.1*(image.north west)$)}]
            \draw[thick,green] (0.5,2.1) rectangle (4.1,6.4) ;
        \end{scope}
    \end{tikzpicture}
		& {\footnotesize{}}
	\begin{tikzpicture}
            \node [
	        above right,
	        inner sep=0] (image) at (0,0) {\includegraphics[width=0.235\textwidth]{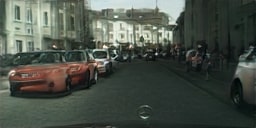} };
	        \begin{scope}[
            x={($0.1*(image.south east)$)},
            y={($0.1*(image.north west)$)}]
            \draw[thick,red] (0.5,2.1) rectangle (4.1,6.4) ;
        \end{scope}
    \end{tikzpicture}
		& {\footnotesize{}}
    \begin{tikzpicture}
            \node [
	        above right,
	        inner sep=0] (image) at (0,0) {\includegraphics[width=0.235\textwidth]{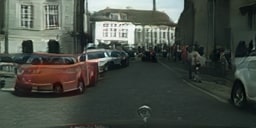}};
            \begin{scope}[
            x={($0.1*(image.south east)$)},
            y={($0.1*(image.north west)$)}]
            \draw[thick,red] (0.5,2.1) rectangle (4.1,6.4) ;
        \end{scope}
    \end{tikzpicture}
		& {\footnotesize{}}
	\begin{tikzpicture}
            \node [
	        above right,
	        inner sep=0] (image) at (0,0) {\includegraphics[width=0.235\textwidth]{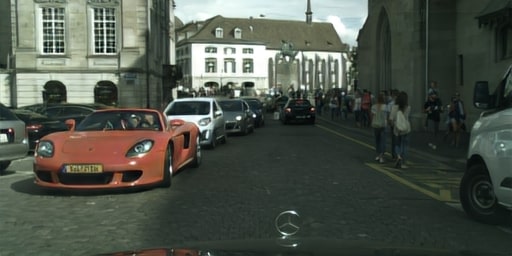}};
            \begin{scope}[
            x={($0.1*(image.south east)$)},
            y={($0.1*(image.north west)$)}]
            \draw[thick,green] (0.5,2.1) rectangle (4.1,6.4) ;
        \end{scope}
    \end{tikzpicture}
		\tabularnewline
		\end{tabular}
\hfill{}
\par\end{centering}
\caption{Qualitative results (best view in color and zoom in) of StyleGAN2 inversion methods on Cityscapes, i.e., pSp${}^\dagger$, Feature-Style encoder\newcite{feature_style} and our masked noise encoder. Note, pSp${}^\dagger$ is an improved version of pSp\newcite{pspencoder} introduced by us.
pSp${}^\dagger$ can reconstruct the rough layout of the scene but still struggles to preserve details. The Feature-Style encoder shows a better reconstruction quality, yet it cannot faithfully reconstruct small objects (e.g. pedestrian), and some objects (e.g. the vehicle, bicycle) are rather blurry. Our masked noise encoder has highest image fidelity, preserving finer details in the inverted image. More visual examples, including the original pSp results, can be found in \cref{fig:encoder-visual-appendix} in the supp. material.}
\label{fig:encoder-visual}
\vspace{-1.0em}
\end{figure*}

The faithful reconstruction of images with complex structures such as driving scenes is non-trivial. Prior methods\newcite{pspencoder,feature_style,roich2021pti,alaluf2022hyperstyle,dinh2022hyperinverter} are mainly tested on simple single-object-centric datasets, e.g., CelebA-HQ\newcite{progressivegan}, FFHQ\newcite{stylegan}, or LSUN\newcite{yu2015lsun}. 
As shown in \newcite{abdal2020image2stylegan++}, extending the native latent space of StyleGAN2 with a stochastic noise space can lead to improved inversion quality. However, all style \emph{and} content information will be embedded in the noise map, leaving the latent codes inactive in this setting.
Therefore, to enable the precise reconstruction of complex driving scenes as well as style mixing, we propose a masked noise encoder for StyleGAN2. The proposed random masking regularization on the noise map encourages the generator to rely on the latent prediction for reconstruction. Thus, it allows to effectively separate content and style information and facilitates realistic style mixing, as shown in \cref{fig:encoder-visual}.

In summary, we make the following contributions:
\begin{itemize}
    \item We propose a masked noise encoder for GAN inversion, which enables high quality reconstruction and style mixing of complex scene-centric datasets. 
    \vspace{-0.2em}
    \item We explore GAN inversion for intra-source data augmentation, which can improve generalization under natural distribution shifts on semantic segmentation. 
    \vspace{-0.2em} 
    %\item Extensive experiments demonstrate that our proposed augmentation method {\ourstyle} can significantly promote domain generalization performance, i.e., achieve up to $12.4\%$ mIoU improvements on driving-scene semantic segmentation,  even with limited diversity in the source data and without access to the target domain.
    \item Extensive experiments demonstrate that our proposed augmentation method {\ourstyle} consistently promotes domain generalization performance on driving-scene semantic segmentation across different network architectures, achieving up to $12.4\%$ mIoU improvement, even with limited diversity in the source data and without access to the target domain.  \vspace{-0.2em}
\end{itemize}
\vspace{-0.5em}

%%%%%% GAN inversion comparison %%%%%%
\section{Related Work}

%%%%%%%%%%%%%%%%%%%%%%%%%%%%%%%
\vspace{-0.5em}
\paragraph{Domain Generalization}
Domain generalization concerns the generalization ability of neural networks on a target domain that follows a different distribution than the source domain, and is inaccessible at training. %
Various approaches have been proposed, which employ data augmentation\newcite{khirodkar2019domain,somavarapu2020frustratingly,huang2021fsdr,zhou2021mixstyle,li2022dsu}, domain alignment\newcite{hu2020mda,li2018domain,li2020domain,jin2020feature,zhou2020optimaltransport}, meta-learning\newcite{li2018learning,balaji2018metareg,Li_Episodic_2019,Zhao_Sebe_2021}, or ensemble learning\newcite{d2018domain,Mancini_2018, Wu_Gong_2021}.
While the majority focuses on image-level tasks, e.g., image classification or person re-identification, a few recent works \newcite{robustnet_2021,wildnet_2022_CVPR,kim2021wedge,kim2022pin} investigate pixel-level prediction tasks such as semantic segmentation. RobustNet\newcite{robustnet_2021} proposes an instance selective whitening loss to the instance normalization, aiming to selectively remove information that causes a domain shift while maintaining discriminative features. \newcite{kim2022pin} introduces a memory-guided meta-learning framework to capture co-occurring categorical knowledge across domains. \newcite{wildnet_2022_CVPR,kim2021wedge} make use of extra data in the wild. 

Another line of work explores feature-level augmentation \newcite{zhou2021mixstyle,li2022dsu}. MixStyle\newcite{zhou2021mixstyle} and DSU \newcite{li2022dsu} add perturbation at the normalization layer to simulate domain shifts at test time. However, this perturbation can potentially cause a distortion of the image content, which can be harmful for semantic segmentation (see \cref{sec:exp_dg}). Moreover, these methods require a careful adaptation to the specific network architecture. 
In contrast, {\ourstyle} performs style mixing on the image-level, thus being model-agnostic, and can be applied as a complement to other methods in order to further increase the generalization performance. 

%%%%%%%%%%%%%%%%%%%%%%%%%%%%%%%
%\subsection{Data Augmentation}
\paragraph{Data Augmentation}
Data augmentation techniques can diversify training samples by altering their style, content, or both, thus preventing overfitting and improving generalization.
Mixup augmentations\newcite{zhang2018mixup,dabouei2021supermix,verma2019manifold}  
linearly interpolate between two training samples and their labels, regularizing both style and content. Despite effectiveness shown on image-level classification tasks, they are not well suited for dense pixel-level prediction tasks. 
CutMix\newcite{yun2019cutmix} cuts and pastes a random rectangular region of the input image into another image, thus increasing the content diversity. Geometric transformation, e.g., random scaling and horizontal flipping, can also serve this purpose. 
In contrast, Hendrycks corruptions\newcite{hendrycks2018benchmarking} only affect the image appearance without modifying the content.
Their generated images look artificial, being far from resembling natural data, and thus offer limited help against natural distribution shifts\newcite{taori2020robustness}. 

StyleMix\newcite{hong2021stylemix} is conceptually closer to our method, which aims to decompose training images into content and style representations and then mix them up to generate more samples. Nonetheless, their AdaIN\newcite{huang2017adain} based style mixing method 
cannot fulfill the pixel-wise label-preserving requirement (see \cref{fig:stylemix-sample-1}). 
Our {\ourstyle} is also a style-based data augmentation technique. Benefiting from the usage of a state-of-the-art GAN, it can generate natural looking samples, altering only the style of the original images while preserving their content and, thus, enabling the re-use of the ground truth label maps. 

%%%%%%%%%%%%%%%%%%%%%%%%%%%%%%%
\paragraph{GAN Inversion}
Showing good results, GAN inversion has been explored for many applications such as face editing \newcite{abdal2019image2stylegan,abdal2020image2stylegan++,zhu2020indomain}, image restoration\newcite{pan2021exploiting}, and data augmentation\newcite{face_da,golhar2022gan}. 
StyleGANs\newcite{stylegan, stylegan2, stylegan2ada} are commonly used for inversion, as they demonstrate high synthesis quality and appealing editing capabilities. 
Nevertheless, there is a known distortion-editability trade-off\newcite{tov2021e4e}. Thus, it is crucial to achieve a curated performance for a specific use case.  

GAN inversion approaches can be classified into three groups: optimization based methods\newcite{creswell2018inverting,abdal2019image2stylegan,abdal2020image2stylegan++,gu2020image,kang2021BDInvert,collins2020editing}, encoder based models\newcite{pspencoder,feature_style,bartz2020onenoise,tov2021e4e,wei2022e2style} methods, and hybrid approaches\newcite{chai2021ensembling,roich2021pti,alaluf2022hyperstyle,dinh2022hyperinverter}. 
Optimization methods generally have worse editability and need exhaustive optimization for each input. Thus,
in this paper, we use an encoder based method for our style mixing purpose.  
The representative encoder based work pSp encoder\newcite{pspencoder} 
embeds the input image in the extended latent space $\mathcal{W}^+$ of StyleGAN. 
\newchange{The e4e encoder\newcite{tov2021e4e} improves editability of pSp while trading off detail preservation.}
To improve reconstruction quality the Feature-Style encoder\newcite{feature_style} further replaces the lower latent code prediction with a feature map prediction. 
Despite much progress, most prior work only showcases applications on single object-centric datasets, such as 
FFHQ\newcite{stylegan},  LSUN\newcite{yu2015lsun}.
They still fail on more complex scenes, thus restricting their application in practice. 
Our masked noise encoder can fulfil both the fidelity and the style mixing capability requirements, rendering itself well-suited for data augmentation for semantic segmentation. To the best of our knowledge, our approach is the first GAN inversion method which can be effectively applied as data augmentation for the semantic segmentation of complex scenes.

\vspace{-0.3em}
\section{Method}
\begin{figure*}[t]
\centering
\includegraphics[width=0.95\linewidth]{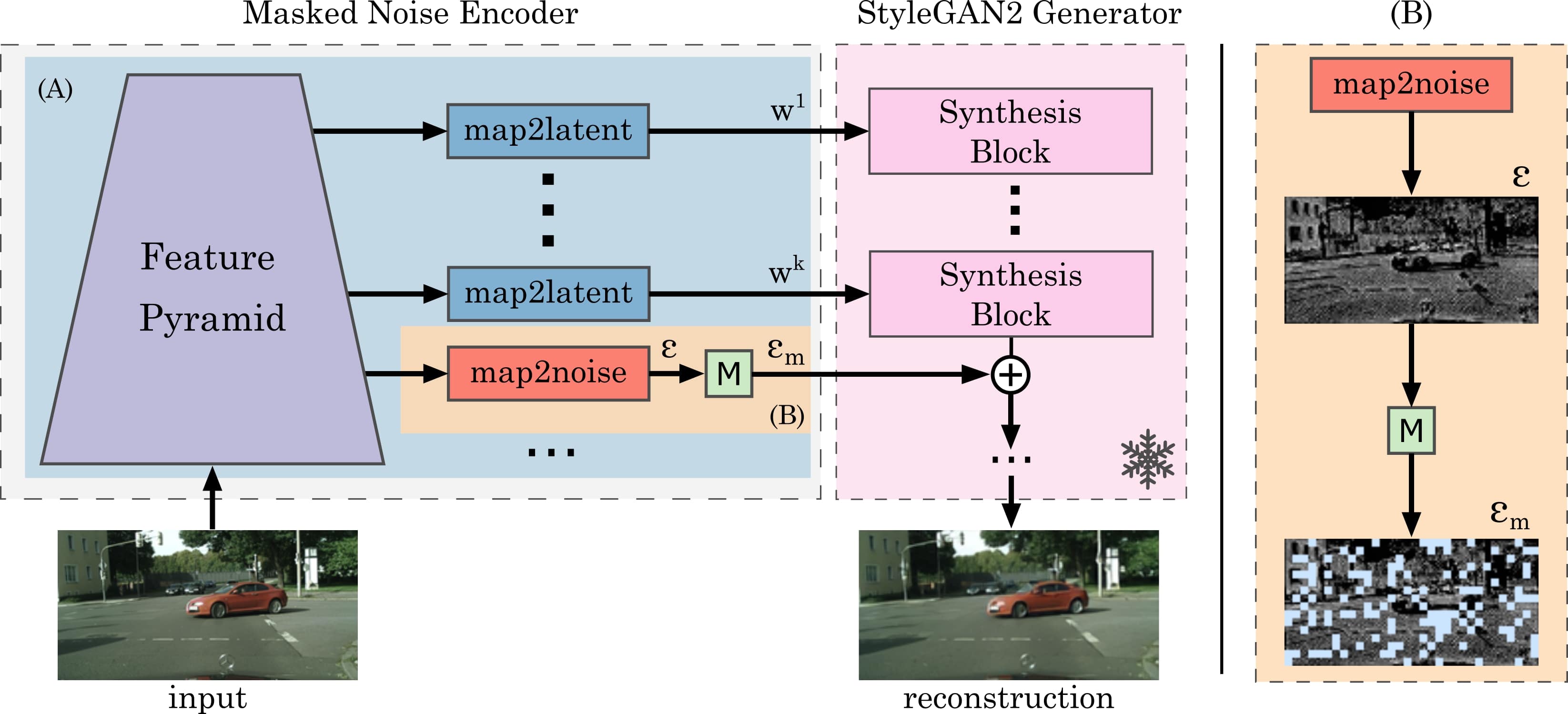}
\caption{\textbf{Method overview.} Our encoder is built on top of the pSp encoder\newcite{pspencoder}, shown in the blue area (A). It maps the input image to the extended latent space $\mathcal{W}^+$ of the pre-trained StyleGAN2 generator. To promote the reconstruction quality on complex scene-centric dataset, e.g., Cityscapes, our encoder additionally predicts the noise map at an intermediate scale, illustrated in the orange area (B). 
$\fbox{M}$ stands for random noise masking, regularization for the encoder training. Without it, the noise map overtakes the latent codes in encoding the image style, so that the latter cannot make any perceivable changes on the reconstructed image, thus making style mixing impossible. 
}
\label{fig:encoder-overview}
\vspace{-1.0em}
\end{figure*}
\vspace{-0.3em}
We introduce our intra-source style augmentation ({\ourstyle}) in \cref{method:style-mixing}, which relies on GAN inversion that can offer faithful reconstruction and style mixing of images. To enable better style-content disentanglement, we propose a masked noise encoder for GAN inversion in \cref{method:gan-inversion}. 
Its detailed training loss is described in \cref{methods: losses}.

%%%%%%%%%%%%%%%%%%%%%%%%%%%
%        Style Mixing     %
%%%%%%%%%%%%%%%%%%%%%%%%%%%
\vspace{-0.3em}
\subsection{Intra-Source Style Augmentation (\ourstyle)}\label{method:style-mixing}
The lack of data diversity and the existence of spurious correlation in the training set often lead to poor domain generalization. To mitigate them, {\ourstyle} aims at modifying styles of the training samples while preserving their semantic content. 
It employs GAN inversion to randomize the style-content combinations in the training set. In doing so, it diversifies the source training set and reduces spurious style-content correlations. Because the content of images is preserved and only the style is changed, the ground truth label maps can be re-used for training, without requiring any further annotation effort.   

{\ourstyle} is built on top of an encoder-based GAN inversion pipeline given its fast inference. GANs, such as StyleGANs\newcite{stylegan,stylegan2,stylegan2ada}, have shown the capability of encoding rich semantic and style information in intermediate features and latent spaces. For encoder-based GAN inversion, an encoder is trained to invert an input image back into the latent space of a pre-trained GAN generator. {\ourstyle} needs an encoder that can separately encode the style and content information of the input image. With such an encoder, it can synthesize new training samples with new style-content combinations, i.e., it can take the content and style codes from different training samples within the source domain and feed them to the pre-trained generator. Since {\ourstyle} modifies only the image style with this encoder, the new synthesized training samples already have a ground truth label map in place.

StyleGAN2 can synthesize scene-centric datasets like Cityscapes\newcite{cordts2016cityscapes} and BDD100K\newcite{yu2020bdd100k}. However, existing GAN inversion encoders cannot provide the desired fidelity to enable {\ourstyle} to improve domain generalization of semantic segmentation via data augmentation. Loss of fine details or %unfaithful  
inauthentic reconstruction of small-scale objects would harm the model's generalization ability. Therefore, we propose a novel encoder design to invert StyleGAN2, termed \emph{masked noise encoder} (see \cref{fig:encoder-overview}).

%%%%%%%%%%%%%%%%%%%%%%%%%%%
%       GAN Inverison     %
%%%%%%%%%%%%%%%%%%%%%%%%%%%
\subsection{Masked Noise Encoder} \label{method:gan-inversion}

We build our encoder upon the pSp encoder\newcite{pspencoder}. It employs a feature pyramid\newcite{lin2017featurepyramid} to extract multi-scale features from a given image, see \cref{fig:encoder-overview}-(A). We improve over pSp by identifying in which latent space to embed the input image for the high-quality reconstruction of the images with complex street scenes. Further, we propose a novel training scheme to enable the style-content disentanglement of the encoder, thus improving its style mixing capability.

\paragraph{Extended Latent Space}
The StyleGAN2 generator takes the latent code $w\in\mathcal{W}$ generated by an MLP network and randomly sampled additive Gaussian noise maps $\{\epsilon\}$ as inputs for image synthesis. As pointed out in\newcite{abdal2019image2stylegan}, it is suboptimal to embed a real image into the original latent space $\mathcal{W}$ of StyleGAN2, due to the gap between the real and synthetic data distributions. A common practice is to map the input image into the extended latent space \wplus. 
The multi-scale features of the pSp feature pyramid are respectively mapped to the latent codes $\{w^k\}$ at the corresponding scales of the StyleGAN2 generator, i.e., $\mathrm{map2latent}$ in \cref{fig:encoder-overview}-(A).

\paragraph{Additive Noise Map}
The latent codes $\{w^k\}$ from the extended latent space {\wplus} alone are not expressive enough to reconstruct images with diverse semantic layouts such as Cityscapes\newcite{cordts2016cityscapes} as shown in \cref{fig:encoder-visual}-(pSp${}^\dagger$). 
The latent codes of StyleGAN2 are one-dimensional vectors that modulate the feature vectors at different spatial positions identically. Therefore, they cannot precisely encode the semantic layout information, which is spatially varying. To address this issue, our encoder additionally predicts the additive noise map $\varepsilon$ of the StyleGAN2 at an intermediate scale, i.e., $\mathrm{map2noise}$ in \cref{fig:encoder-overview}-(B). 
\begin{figure}[t]
    \begin{centering}
    \setlength{\tabcolsep}{0.0em}
    \renewcommand{\arraystretch}{0}
    \par\end{centering}
    \begin{centering}
    \vspace{-0.5em}
    \hfill{}
	\begin{tabular}{@{}c@{}c}
        \centering
		\small Content &  \small Style   \tabularnewline
		\includegraphics[width=0.45\linewidth]{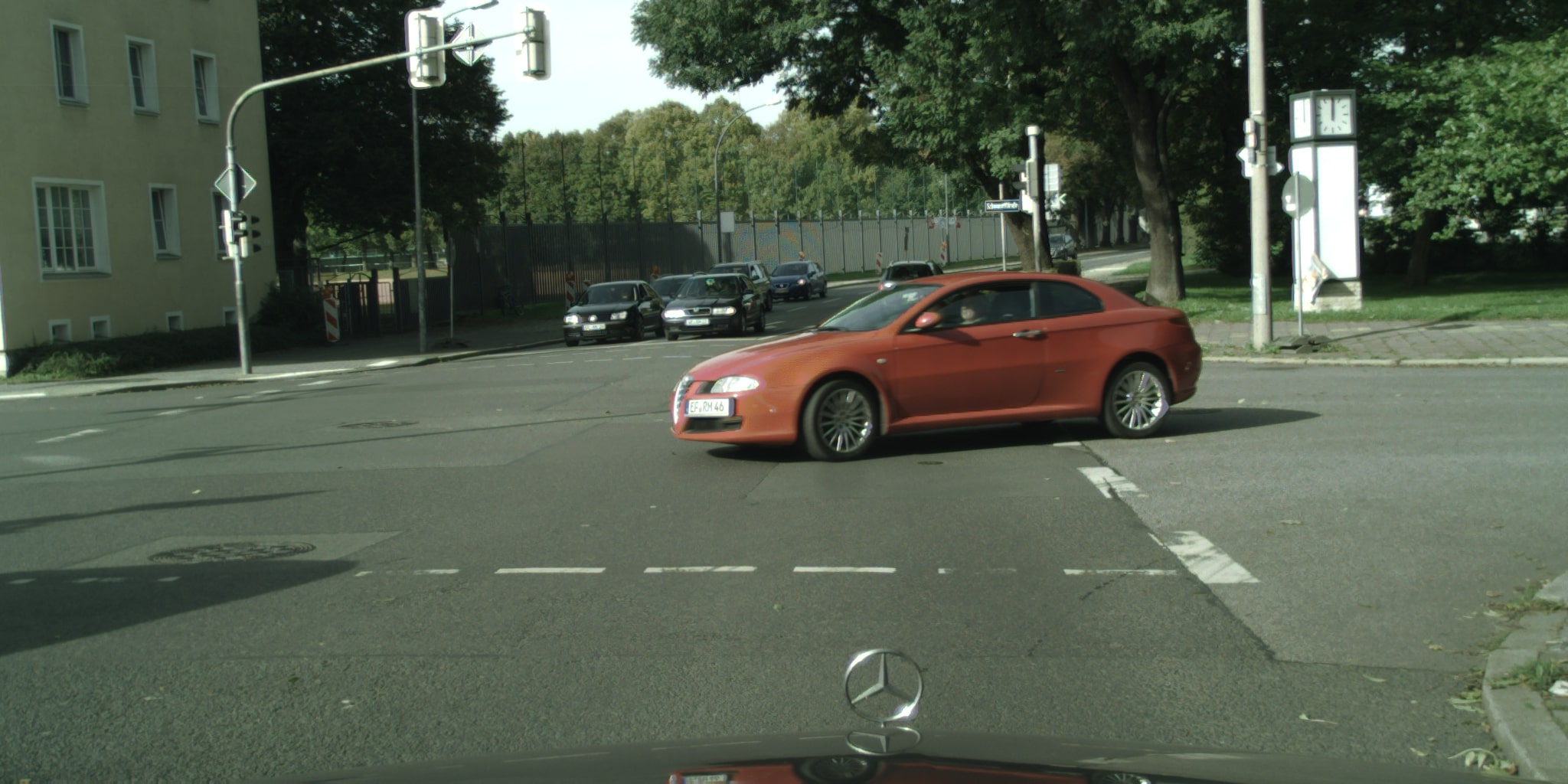} &
		{\hspace{0.5em}} \includegraphics[width=0.45\linewidth]{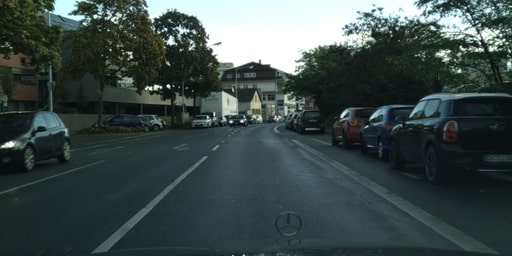} \tabularnewline
	     \small W/o masking & \small W/- masking (Ours)  \tabularnewline
		\includegraphics[width=0.45\linewidth]{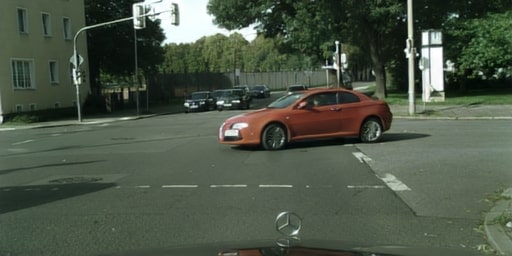} & {\hspace{0.5em}}
		\includegraphics[width=0.45\linewidth]{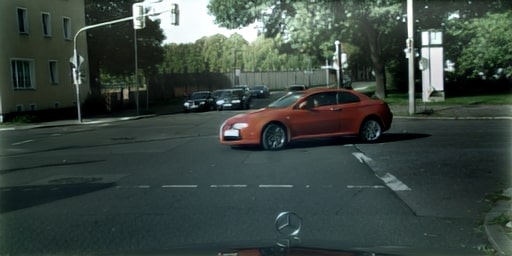} 
		\tabularnewline
		\end{tabular}
\hfill{}
\par\end{centering}
\caption{Style mixing effect enabled by random noise masking (best view in color). Despite the good reconstruction quality, the encoder trained without masking cannot change the style of the given $\mathrm{Content}$ image. In contrast, the encoder trained with masking can modify it using the style from the given $\mathrm{Style}$ image.}
\label{fig:masking-ablation}
\vspace{-0.7em}
\end{figure} 
\begin{figure}[t]
    \begin{centering}
    \setlength{\tabcolsep}{0.0em}
    \renewcommand{\arraystretch}{0}
    \par\end{centering}
    \begin{centering}
    \vspace{-0.5em}
    \hfill{}
	\begin{tabular}{@{}c@{}c@{}c}
        \centering
		 & &  \tabularnewline
		 
		 \includegraphics[width=0.32\linewidth]{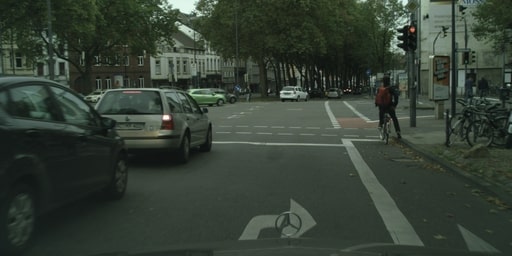} &
		{\footnotesize{}} \includegraphics[width=0.32\linewidth]{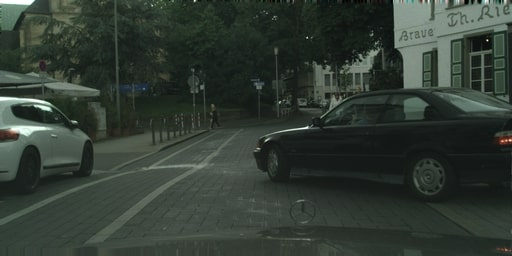} & {\footnotesize{}}
		\includegraphics[width=0.32\linewidth]{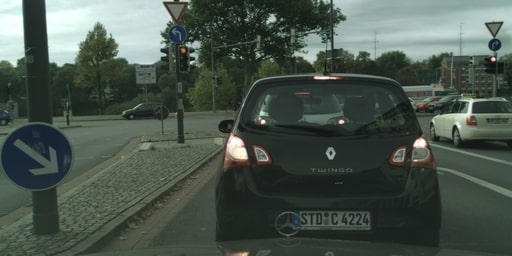} 
		\tabularnewline
		 
		\includegraphics[width=0.32\linewidth]{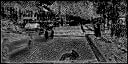} &
		{\footnotesize{}} \includegraphics[width=0.32\linewidth]{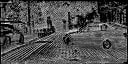} & {\footnotesize{}}
		\includegraphics[width=0.32\linewidth]{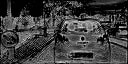} 
		\tabularnewline
		\end{tabular}
\hfill{}
\par\end{centering}
\caption{Noise map visualization of our masked noise encoder. The noise map encodes the semantic content of the image.}
\label{fig:noise-vis}
\vspace{-1.0em}
\end{figure}

\paragraph{Random Noise Masking}
While offering high-quality reconstruction, the additive noise map can be too expressive so that it encodes nearly all perceivable details of the input image. This results in a poor style-content disentanglement and can damage the style mixing capability of the encoder (see \cref{fig:masking-ablation}). 
To avoid this undesired effect, we propose to regularize the noise prediction of the encoder by random masking of the noise map. Note that the random masking as a regularization technique has also been successfully used in reconstruction-based self-supervised learning\newcite{xie2022simmim,he2022mae}. 
In particular, we spatially divide the noise map into non-overlapping $P\times P$ patches, see $\fbox{M}$ in \cref{fig:encoder-overview}-(B). Based on a pre-defined ratio $\rho$, a subset of patches is randomly selected and replaced by patches of unit Gaussian random variables $\epsilon \sim N(0,1)$ of the same size. $N(0,1)$ is the prior distribution of the noise map at training the StyleGAN2 generator. 
We call this encoder \emph{masked noise encoder} as it is trained with random masking to predict the noise map.

The proposed random masking reduces the encoding capacity of the noise map, hence encouraging the encoder to jointly exploit the latent codes $\{w^k\}$ for reconstruction. \cref{fig:masking-ablation} visualizes the style mixing effect. 
The encoder takes the noise map $\varepsilon_c$ and latent codes $\{w_s^k\}$ from the $\mathrm{content}$ image and $\mathrm{style}$ image, respectively. 
Then, they are fed into StyleGAN2 to synthesize a new image, i.e., $G(w_s^k, \varepsilon_c)$. 
If the encoder is not trained with random masking, the new image does not have any perceptible difference with the $\mathrm{content}$ image. This means the latent codes $\{w^k\}$ encode negligible information of the image. In contrast, when being trained with masking, the encoder creates a novel image that takes the content and style from two different images. This observation confirms the enabling role of masking for content and style disentanglement, and thus the improved style mixing capability. The noise map no longer encodes all perceptible information of the image, including style and content. In effect, the latent codes $\{w^k\}$ play a more active role in controlling the style.  
In \cref{fig:noise-vis}, we further visualize the noise map of the masked noise encoder and observe that it captures well the semantic content of the scene.

\begin{figure*}[t]
    \begin{centering}
    \setlength{\tabcolsep}{0.0em}
    \renewcommand{\arraystretch}{0}
    \par\end{centering}
    \begin{centering}
    \vspace{-0.5em}
    \hfill{}
	\begin{tabular}{@{}l@{}c@{}c@{}c}
        \centering
		 &   &  &  \tabularnewline
        \hspace{0.08\textwidth} Content $I_c$ \hspace{0.04\textwidth} \rotatebox{90}{\hspace{1.3em} Style $I_s$}

		& {\footnotesize{}}
		\includegraphics[width=0.235\textwidth]{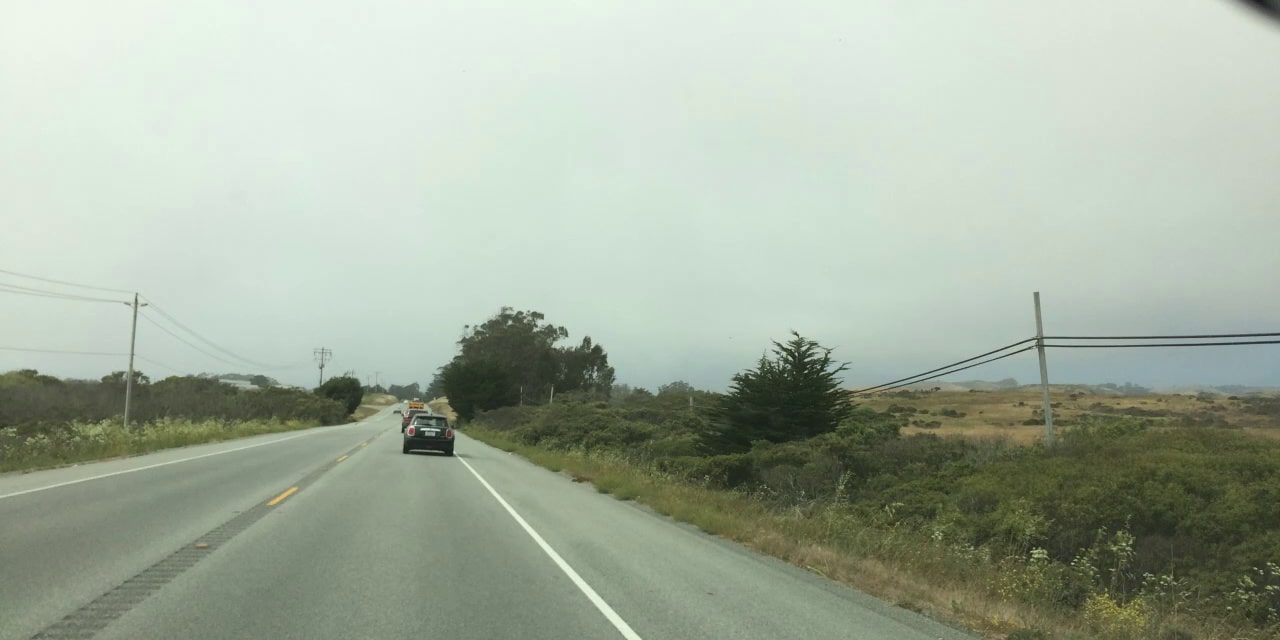} & {\footnotesize{}}
		\includegraphics[width=0.235\textwidth]{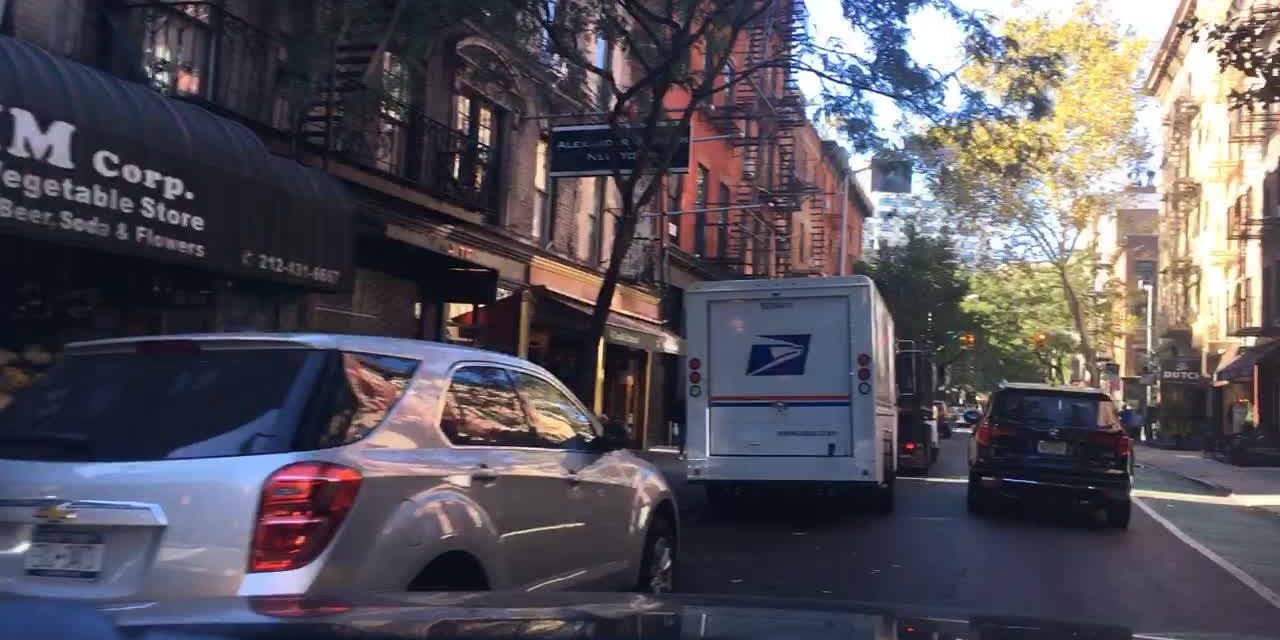} & {\footnotesize{}}
		\includegraphics[width=0.235\textwidth]{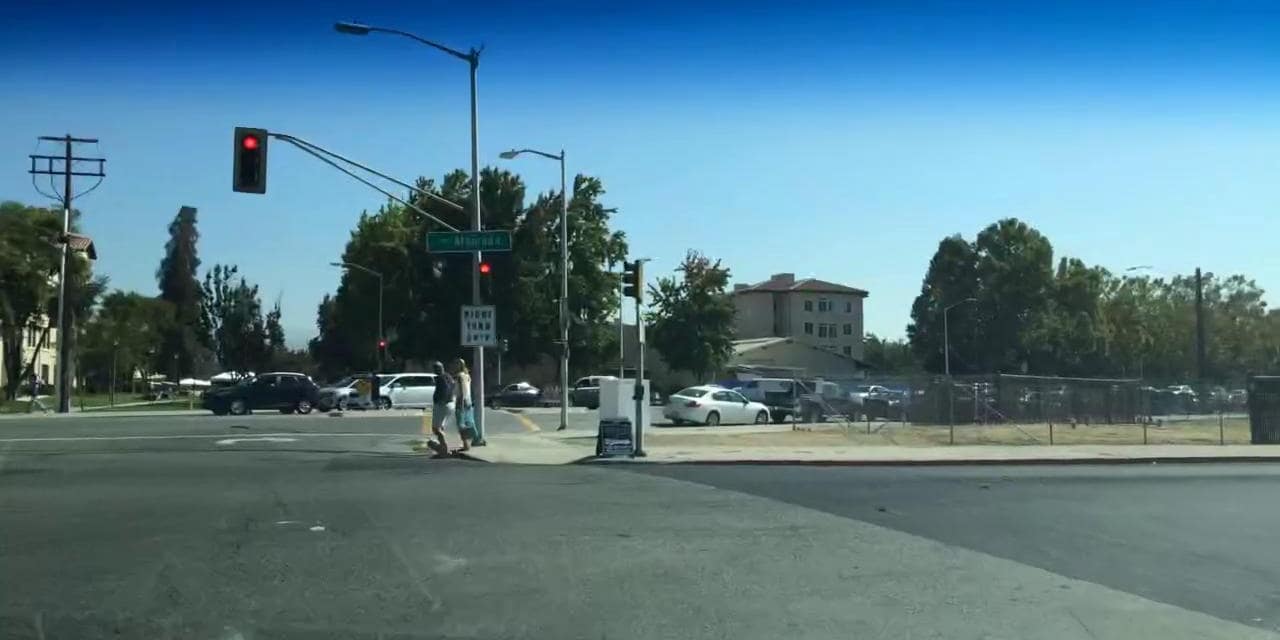}
		\tabularnewline
		\includegraphics[width=0.235\textwidth]{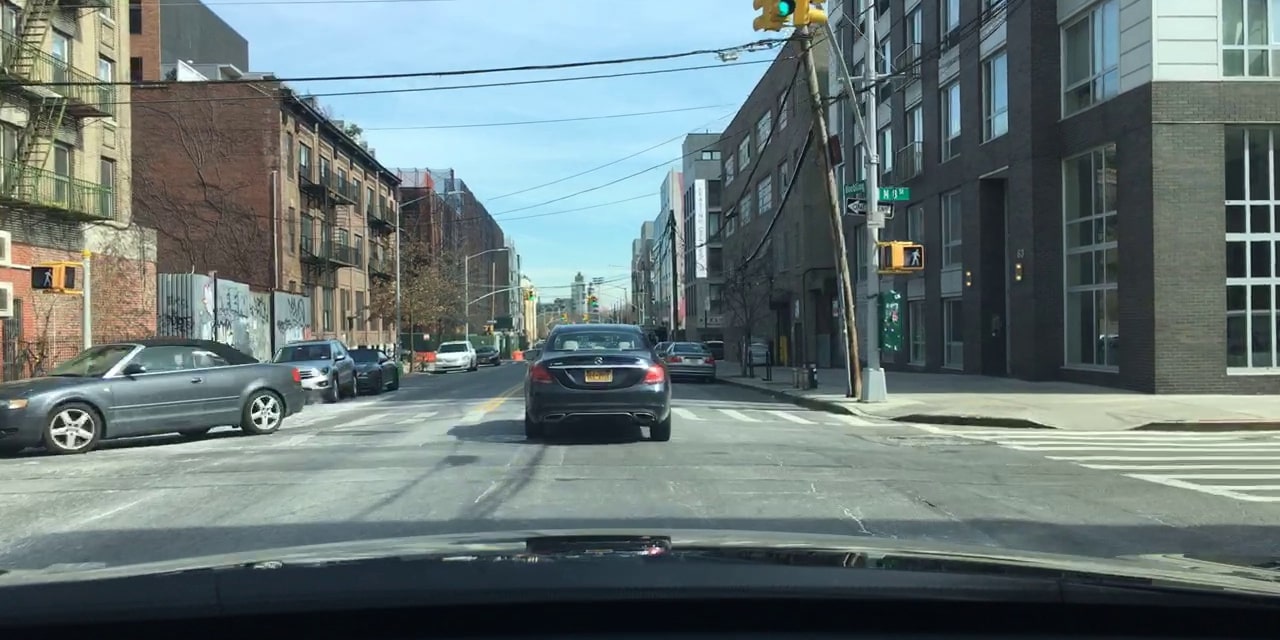} & {\footnotesize{}}
		\includegraphics[width=0.235\textwidth]{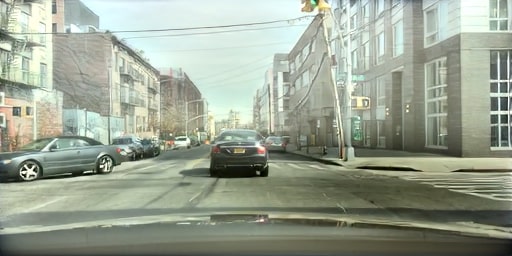} & {\footnotesize{}}
		\includegraphics[width=0.235\textwidth]{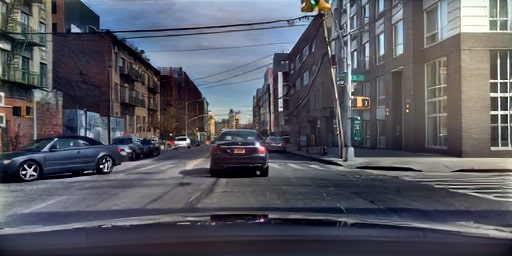} & {\footnotesize{}}
		\includegraphics[width=0.235\textwidth]{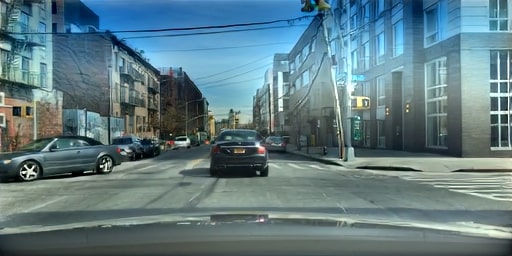}
		\tabularnewline
		\includegraphics[width=0.235\textwidth]{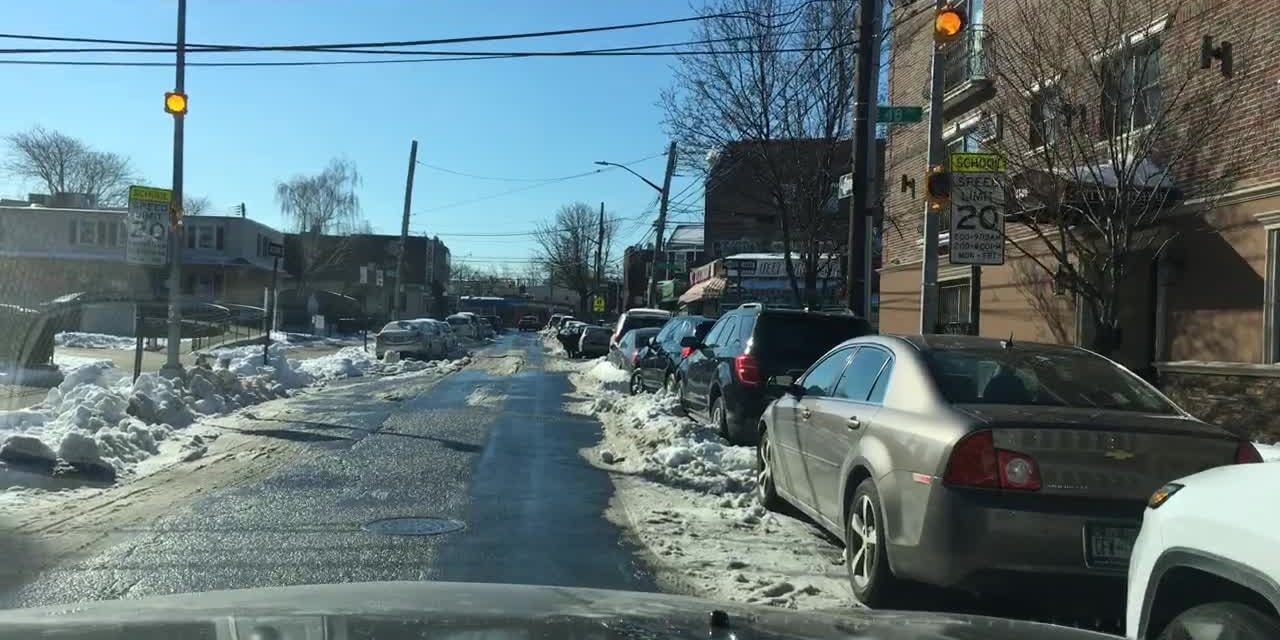} & {\footnotesize{}}
		\includegraphics[width=0.235\textwidth]{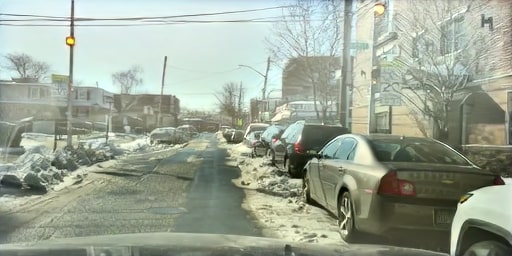} & {\footnotesize{}}
		\includegraphics[width=0.235\textwidth]{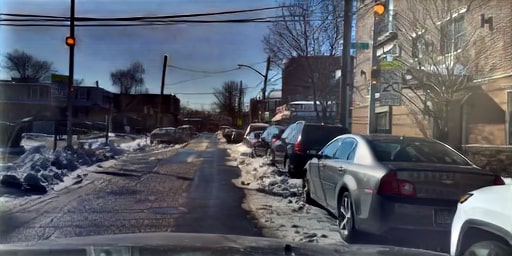} & {\footnotesize{}}
		\includegraphics[width=0.235\textwidth]{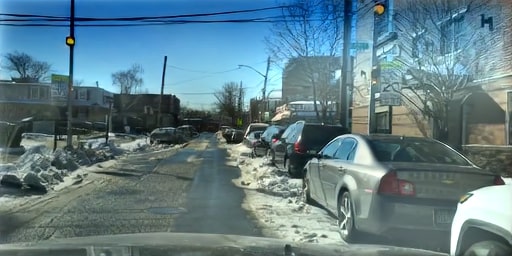}
		\tabularnewline
		\end{tabular}
\hfill{}
\par\end{centering}
\caption{Visual examples of style mixing on BDD100K (best view in color) enabled by our masked noise encoder. 
By combining the latent codes $\{w_s^k\}$ of $I_s$ and the noise map $\varepsilon_c$ of $I_c$, the synthesized images \newchange{$G(w_s^k, \varepsilon_c)$} preserve the content of $I_c$ with a new style resembling $I_s$.
}
\label{fig:intra-mix-example}
\vspace{-1em}
\end{figure*}

%%%%%%%%%%%%%%%%%%%%%%%%%%%%%%%%%%%%%%%%
%%%%%%%%%%%%%%%%%%%%%%%%%%%%%%%%%%%%%%%%%
\subsection{Encoder Training Loss}\label{methods: losses}
Mathematically, the proposed StyleGAN2 inversion with the masked noised encoder $E^M$ can be formulated as
\begin{align}
    \{w^1,\dots, w^K,\varepsilon\} &= E^M(x); \label{eq:encoder-1}\\
    x^* &=G\circ E^M(x)= G(w^1,\dots,w^K, \varepsilon).\nonumber
\end{align}
The masked noise encoder $E^M$ maps the given image $x$ onto the latent codes $\{w^k\}$ and the noise map $\varepsilon$. The StyleGAN2 generator $G$ takes both $\{w^k\}$ and $\varepsilon$ as the input and generates $x^*$. Ideally, $x^*$ should be identical to $x$, i.e., a perfect reconstruction. 

When training the masked noise encoder $E^M$ to reconstruct $x$, the original noise map $\varepsilon$ is masked before being fed into the pre-trained $G$
\begin{align} 
\varepsilon_M &= (1 - M_{noise})  \odot \varepsilon + M_{noise} \odot \epsilon, \\
\tilde{x} &= G(w^1,\dots,w^K, \varepsilon_M), \label{eq:encoder}
\end{align}
where $M_{noise}$ is the random binary mask, $\odot$ indicates the Hadamard product, and $\tilde{x}$ denotes the reconstructed image with the masked noise $\varepsilon_M$. The training loss for the encoder is given as
\begin{align} 
\mathcal{L} =  \mathcal{L}_{mse} + 
    \lambda_1 \mathcal{L}_{lpips} + 
    \lambda_2 \mathcal{L}_{adv} + \lambda_3 \mathcal{L}_{reg} ,\label{eq:loss}
\end{align}
where $\{\lambda_i\}$ are weighting factors. %
The first three terms are the pixel-wise MSE loss, learned perceptual image patch similarity (LPIPS)\newcite{zhang2018lpips} loss and adversarial loss\newcite{goodfellow2014gan}, 
\begin{align} 
\mathcal{L}_{mse} &= \norm{(1 - M_{img})  \odot (x - \tilde{x})}_2, \\
\mathcal{L}_{lpips} &= \norm{(1 - M_{feat}) \odot
(\mathrm{VGG}(x) - \mathrm{VGG}(\tilde{x}))}_2, \\
\mathcal{L}_{adv} &= -\log D(G(E^M(x))).
\label{eq:encoder-loss-rec}
\end{align}
which are the common reconstruction losses for encoder training\newcite{pspencoder,zhu2020indomain}. Note that masking removes the information of the given image $x$ at certain spatial positions, the reconstruction requirement on these positions should then be relaxed. $M_{img}$ and $M_{feat}$ are obtained by up- and down-sampling the noise mask $M_{noise}$ to the image size and the feature size of the VGG-based feature extractor. The adversarial loss is obtained by formulating the encoder training as an adversarial game with a discriminator $D$ that is trained to distinguish between reconstructed and real images.

The last regularization term is defined as
\begin{align}
\mathcal{L}_{reg}=\norm{\varepsilon}_1 +  \norm{E^M_{w}(G(w_{gt}, \epsilon)) - w_{gt}}_2.
\end{align}
The L1 norm helps to induce sparse noise prediction. It is complementary to random masking, reducing the capacity of the noise map. The second term is obtained by using the ground truth latent codes $w_{gt}$ of synthesized images $G(w_{gt}, \epsilon)$ to train the latent code prediction $E^M_{w}(\cdot)$\newcite{feature_style}. It guides the encoder to stay close to the original latent space of the generator, speeding up the convergence.
\vspace{-0.25em}
\section{Experiments}

\cref{sec: exp_gan_inversion} and \cref{sec:exp_dg} respectively report our experiments on the StyleGAN2 inversion and domain generalization of semantic segmentation.

\paragraph{Datasets}
We conduct extensive experiments using the \hbox{following} driving scene datasets: Cityscapes~(CS)\newcite{cordts2016cityscapes}, BDD100K~(BDD)\newcite{yu2020bdd100k}, ACDC\newcite{sakaridis2021acdc} and Dark Z\"urich~(DarkZ)\newcite{sakaridis2019darkzurich}. Cityscapes is collected from different cities primarily in Germany, under good/medium weather conditions during daytime. BDD100K is a driving-scene dataset collected in the US, representing a geographic location shift from Cityscapes. Besides, it also includes more diverse scenes (e.g., city streets, residential areas, and highways) and different weather conditions captured at different times of the day. Both ACDC and Dark Z\"urich are collected in Switzerland. ACDC contains four adverse weather conditions (rain, fog, snow, night) and Dark Z\"urich contains night scenes. The default setting is to use Cityscapes as the source training data, whereas the validation sets of the other datasets represent unseen target domains with different types of natural shifts, i.e., used only for testing. In the supp.~material, we also report the numbers where BDD100K is used as the source set and the remaining datasets are treated as unseen domains. In both cases, we consider a \emph{single source domain} for training.

\paragraph{Training details}
We experiment with two image resolutions: $128\times256$ and $256\times512$. 
The StyleGAN2\newcite{stylegan2ada} model is first trained to \emph{unconditionally} synthesize images and then fixed during the encoder training. To invert the pre-trained StyleGAN2 generator, the masked noise encoder predicts both latent codes in the extended {\wplus} space and the additive noise map. In accordance with the StyleGAN2 generator, {\wplus} space consists of $14$ and $16$ latent code vectors for the input resolution $128\times256$ and $256\times512$, respectively. The additive noise map is always at the intermediate feature space with one fourth of the input resolution. We use the same encoder architecture, optimizer, and learning rate scheduling as pSp\newcite{pspencoder}. Our encoder is trained with the loss function defined in \cref{eq:loss} with $\lambda_1=10$ and $\lambda_2= \lambda_3=0.1$. For our random noise masking, we use a patch size $P$ of $4$ with a masking ratio $\rho= 25\%$. \newchange{A detailed ablation study on the noise map and a computational complexity analysis of the encoder
can be found in \red{S.1}.}

We use the trained masked noise encoder to perform {\ourstyle} as described in \cref{method:style-mixing}. We experiment with several architectures for semantic segmentation, i.e., HRNet\hrnet, SegFormer\segformer, and DeepLab v2/v3+\newcite{chen2017deeplabv2,chen2018deeplabv3plus}. 
The baseline segmentation models are trained with their default configurations and using the standard augmentation, i.e., random scaling and horizontal flipping. 

%%%%%%%%%%%%%%%%%%%%%%%%%%%%%%%%%%%%%
\subsection{Masked Noise Encoder}\label{sec: exp_gan_inversion}

\vspace{-0.5em}
\paragraph{Reconstruction quality} 
\Cref{tab:gan-inversion-comparison} shows that our masked noise encoder considerably outperforms two strong \mbox{StyleGAN2} inversion baselines pSp\newcite{pspencoder} and Feature-Style encoder\newcite{feature_style} in all three evaluation metrics. The achieved low values of MSE, LPIPS\newcite{zhang2018lpips} and FID\newcite{heusel2017fid} indicate its high-quality reconstruction. 
Both the masked noise encoder and the Feature-Style encoder adopt the adversarial loss $\mathcal{L}_{adv}$  
and regularization using synthesized images with ground truth latent codes $w_{gt}$.
Therefore, we also add them to train pSp and note this version as $\text{pSp}^\dagger$. While $\text{pSp}^\dagger$ improves over pSp in MSE and FID, it still underperforms compared to the others. This confirms that inverting into the extended latent space {\wplus} only allows limited reconstruction quality on Cityscapes.
The Feature-Style encoder\newcite{feature_style} replaces the prediction of the low level latent codes with feature prediction, which results in better reconstruction without severely harming style editability. 
However, its reconstruction on Cityscapes is still not satisfying and underperforms to our masked noise encoder.
As noted in \newcite{feature_style},
the feature size of the Feature-Style encoder is restricted. Using a larger feature map to improve reconstruction quality can only be done as a replacement of more latent code predictions. Consequently, it largely reduces the expressiveness of the latent embedding and leads to extremely poor editability, being no longer suitable for downstream applications, e.g., style mixing data augmentation.

The visual comparison across $\text{pSp}^\dagger$, the Feature-Style encoder and our masked noise encoder is shown in \cref{fig:encoder-visual} and is aligned with the quantitative results in \Cref{tab:gan-inversion-comparison}.  \changed{Caption of Table 1 is changed due to simplified loss formulation!}
$\text{pSp}^\dagger$ has overall poor reconstruction quality. The Feature-Style encoder cannot faithfully reconstruct small objects and restore fine details. In comparison, our masked noise encoder offers high-quality reconstruction, preserving the semantic layout and fine details of each class. Having a high-quality reconstruction is an important requirement for using the encoder for data augmentation. Unfortunately, neither $\text{pSp}^\dagger$ nor the Feature-Style encoder achieve satisfactory reconstruction quality. For instance, they both fail at capturing the red traffic light in \cref{fig:encoder-visual}. Using such images for data augmentation can confuse the semantic segmentation model, leading to performance degradation.

\begin{table}[t]
\begin{center}
{\small
    {
    \begin{tabular}{l|lcc}
        Method & MSE $\downarrow$ & LPIPS $\downarrow$  & FID $\downarrow$  \\
    \midrule
        pSp\newcite{pspencoder} & 0.078  & 0.348 & 130.62  \\
        pSp${}^\dagger$\newcite{pspencoder} & 0.049  & 0.339 & 14.60  \\
        Feature-Style\newcite{feature_style} & 0.025  & 0.220 & 7.14   \\
        \textbf{Ours} & \textbf{0.011}  & \textbf{0.124} & \textbf{3.94}  \\
    \end{tabular}
    }
}
\end{center}
\caption{Reconstruction quality on Cityscapes at the resolution $128\times 256$. MSE, LPIPS\newcite{zhang2018lpips} and FID\newcite{heusel2017fid} respectively measure the pixel-wise reconstruction difference, perceptual difference, 
and distribution difference between the real and reconstructed images. The proposed masked noise encoder (Ours) consistently outperforms pSp, $\text{pSp}^\dagger$ and the feature-style encoder. Note, $\text{pSp}^\dagger$ is introduced by us, by training pSp with an additional discriminator and incorporating synthesized images for better initialization. 
}
\label{tab:gan-inversion-comparison}
\vspace{-0.6em}
\end{table}

%%%%% Intra-mixing augmentation with and without masking %%%%%%%
\begin{table}[t]
\begin{center}
{\small%\small \footnotesize%
    {
    \begin{tabular}{@{}l|c|cccc@{}}
        %\toprule
        Method & CS & ACDC & BDD & DarkZ  \\ \midrule
        Baseline   & \textbf{70.47}  & 41.48 & 45.66  & 15.25  \\
        %\midrule
        {\ourstyle} w/o masking  & 69.68 & 44.63 & 46.45 & 17.36  \\
        {\ourstyle} w/- masking  & 69.48  &\textbf{ 47.43} & \textbf{47.87} & \textbf{26.10}  \\
        %\bottomrule
    \end{tabular}
    }
}
\end{center}
\caption{The effect of random noise masking on improving domain generalization via {\ourstyle}. We report the mean Intersection over Union (mIoU) of HRNet\newcite{wang2020hrnet} trained on Cityscapes at the resolution $256\times 512$. BDD100K (BDD), ACDC, and Dark Z\"urich (DarkZ) represent different domain shifts from Cityscapes.}
\label{tab:mask-cs-da}
%\vspace{-0.1em}
\end{table}

%%%%% SemSeg + Data augmetnation - Cityscapes%%%%
\begin{table*}[t]
\begin{center}
{\small 
    \begin{tabular}{@{}l|c|ccccc||c|ccccc @{}}
         & \multicolumn{6}{c||}{HRNet\hrnet} & \multicolumn{6}{c}{SegFormer\newcite{xie2021segformer}} \\ 
        Method & CS & Rain & Fog & Snow & Night & Avg. & CS & Rain & Fog & Snow & Night & Avg.\\ \midrule
        
        Baseline  & 70.47 & 44.15 & 58.68 & 44.20 & 18.90 & 41.48  & 67.90 & 50.22 & 60.52 & 48.86 & 28.56 & 47.04\\ 
        \midrule
        CutMix\newcite{yun2019cutmix}   & \textbf{72.68} & \underline{42.48} & \underline{58.63} & 44.50 & \underline{17.07} & \underline{40.67} & \textbf{69.23} & \underline{49.53} & 61.58 & \underline{47.42} & \underline{27.77} & \underline{46.57} \\
        Weather\newcite{hendrycks2018benchmarking}   & 69.25 & \textbf{50.78} & 60.82 & \underline{38.34} & 22.82 & 43.19  & 67.41 & 54.02 & 64.74 & 49.57 & 28.50 & 49.21 \\
        StyleMix\newcite{hong2021stylemix} & 57.40 & \underline{40.59} & \underline{49.11}  & \underline{39.14} & 19.34 & \underline{37.04} & 65.30 & 53.54 & 63.86 & 49.98 & 28.93 & 49.08\\ 
        \textbf{{\ourstylebf} (Ours)}    & 70.30 & 50.62 & \textbf{66.09} & \textbf{53.30} & \textbf{30.18} & \textbf{50.05} & 67.52 & \textbf{55.91} & \textbf{67.46} & \textbf{53.19} & \textbf{33.23} & \textbf{52.45}  \\
        \midrule
        Oracle & 70.29 & 65.67 & 75.22 & 72.34 & 50.39 & 65.90  & 68.24 & 63.67 & 74.10 & 67.97 & 48.79 & 63.56 \\
    \end{tabular}
}
\end{center}
\caption{
Comparison of data augmentation for improving domain generalization, i.e., from Cityscapes (train) to ACDC (unseen). The mean Intersection over Union (mIoU) is reported on Cityscapes (CS), four individual scenarios of ACDC (Rain, Fog, Snow and Night) and the whole ACDC (Avg.).
Oracle indicates the supervised training on both Cityscapes and ACDC, serving as an upper bound on ACDC for the other methods. Note, it is not supposed to be an upper bound on Cityscapes. Underline denotes worse results than the baseline on ACDC. {\ourstyle} performs the best and consistently improves the mIoU in all four scenarios of ACDC using both HRNet and SegFormer.}
\label{tab:hrnet-segformer-cityscapes-dg}
\vspace{-0.7em}
\end{table*}

\paragraph{Ablation on the masking effect} 
In \cref{fig:masking-ablation} and \cref{fig:intra-mix-example}, we visually observe that random masking offers a stronger perceivable style mixing effect compared to the model trained without masking. 
Next, we test the effect of masking on improving the domain generalization for the semantic segmentation task. In particular, we employ the encoder that is trained with and without masking to perform {\ourstyle}. In \cref{tab:mask-cs-da}, while slightly degrading the source domain performance of the baseline model on Cityscapes, {\ourstyle} improves the domain generalization performance on BDD100K, ACDC and Dark Z\"urich. As {\ourstyle} with masked noise encoder is more effective at diversifying the training set and reducing the style-content correlation, it achieves more pronounced gains in \cref{tab:mask-cs-da}, e.g., more than $10\%$ improvement in mIoU from Cityscapes to Dark Z\"urich. 

% 2x2
\begin{figure}[t]
    \begin{centering}
    \setlength{\tabcolsep}{0.0em}
    \renewcommand{\arraystretch}{0}
    \par\end{centering}
    \begin{centering}
    \vspace{-0.4em}
    \hfill{}
	\begin{tabular}{@{}c@{}c}
        \centering
		\small Content & \small Style  \tabularnewline
		\includegraphics[width=0.47\linewidth]{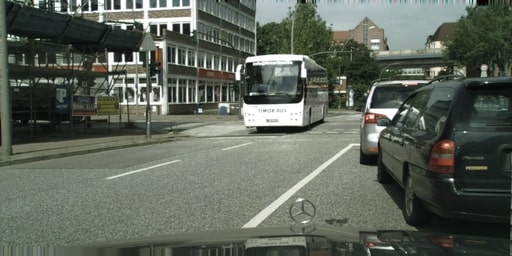} &
		{\footnotesize{}} \includegraphics[width=0.47\linewidth]{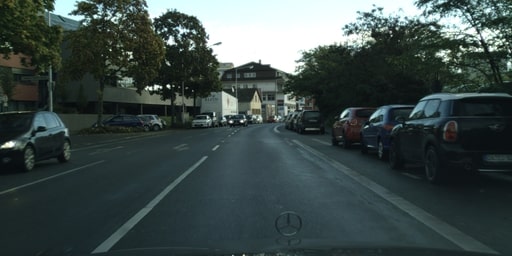}  \tabularnewline
		
		\small StyleMix\newcite{hong2021stylemix} & \small ISSA (Ours) \tabularnewline
		\includegraphics[width=0.47\linewidth]{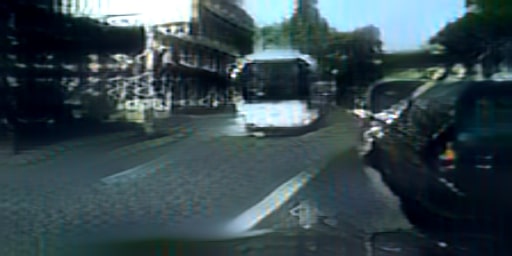} 
		&{\footnotesize{}} \includegraphics[width=0.47\linewidth]{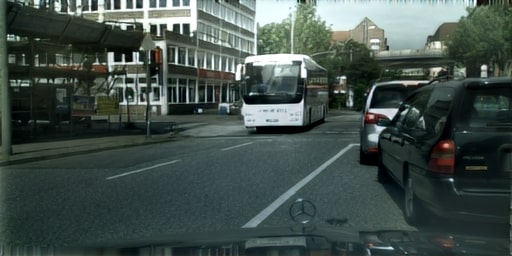}
		\tabularnewline
		\end{tabular}
\hfill{}
\par\end{centering}
\caption{Comparison of StyleMix\newcite{hong2021stylemix} and ISSA. StyleMix has rather low fidelity, while {\ourstyle} can preserve more details. }
\label{fig:stylemix-sample-1}
\vspace{-1.0em}
\end{figure}

%%%%%%%%%%%%%%%%%%%%%%%%%%%%%%%%%%
\subsection{Domain Generalization}\label{sec:exp_dg}

\paragraph{Comparison with data augmentation methods}
\cref{tab:hrnet-segformer-cityscapes-dg} reports the mIoU scores of Cityscapes to ACDC domain generalization using two semantic segmentation models, i.e., HRNet{\hrnet} and SegFormer\segformer. 
{\ourstyle} is compared with three representative data augmentations methods, i.e., CutMix\newcite{yun2019cutmix}, Hendrycks's weather corruptions\newcite{hendrycks2018benchmarking}, and StyleMix\newcite{hong2021stylemix}. Remarkably, our {\ourstyle} is the top performing method, consistently improving mIoU in both models and across all four different scenarios of ACDC, i.e., rain, fog, snow and night. Compared to HRNet, SegFormer is more robust against the considered domain shifts.

In contrast to the others, CutMix mixes up the content rather than the style. It improves the in-distribution performance on Cityscapes, but this gain does not extend to domain generalization. Hendrycks's weather corruptions can be seen as the synthetic version of Cityscapes under the rain, fog, and snow weather conditions. While already mimicking ACDC at training, it can still degrade ACDC-snow by more than $5.8\%$ in mIoU using HRNet. More results on the other corruption types can be found in the supp.~material. StyleMix\newcite{hong2021stylemix} also seeks to mix up styles. However, it does not work well for scene-centric datasets, such as Cityscapes. Its poor synthetic image quality (see \cref{fig:stylemix-sample-1}) leads to the performance drop over the HRNet baseline in many cases, e.g., on Cityscapes to ACDC-fog from $58.68\%$ to $49.11\%$ mIoU.

%%%%% SotA DG + Data augmetnation - Cityscapes %%%%
\begin{table}[t]
\begin{center}
{\small
    \begin{tabular}{@{}l|c|cccc@{}}
        Method  & CS & ACDC & BDD & DarkZ   \\ \midrule
        Baseline\newcite{chen2017deeplabv2} & 61.73  & 30.86 & 34.30  &  11.62 \\ 
    \midrule
        MixStyle\newcite{zhou2021mixstyle} & 59.01 & 36.97 & 36.27  & 9.38  \\
        DSU \newcite{li2022dsu}   & 59.59 & 38.31 & 35.53  & 12.29   \\
        \textbf{{\ourstylebf} (Ours)} & \textbf{62.20} & \textbf{43.21} & \textbf{42.60} & \textbf{21.56} \\ 
    \midrule
        MixStyle\newcite{zhou2021mixstyle} + \ourstyle  & 60.17 & 41.81  &42.17  &  20.56\\ 
        DSU \newcite{li2022dsu} + \ourstyle  & 60.20 & 43.31 & 42.24 & 24.63  \\ 
    \end{tabular}
}
\vspace{-0.5em}
\end{center}
\caption{Comparison with feature-level augmentation methods on domain generalization performance of Cityscapes 
as the source.
Following DSU\newcite{li2022dsu}, we conduct experiments using DeepLab v2\newcite{chen2017deeplabv2} as the baseline for fair comparison.}
\label{tab:feature-aug-da}
\vspace{-0.6em}
\end{table}
\paragraph{Comparison with domain generalization techniques}
We further compare {\ourstyle} with two advanced feature space style mixing methods designed to improve domain generalization performance: MixStyle\newcite{zhou2021mixstyle} and DSU\newcite{li2022dsu}. Both extract the style information at certain normalization layers of CNNs. MixStyle\newcite{zhou2021mixstyle} mixes up styles by linearly interpolating the feature statistics, i.e., mean and variance, of different images, while DSU\newcite{li2022dsu} models the feature statistics as a distribution and randomly draws samples from it. 

We adopt the experimental setting of DSU with default hyperparameters, using DeepLab v2\newcite{chen2017deeplabv2}
segmentation network with ResNet101 backbone.
\cref{tab:feature-aug-da} shows that {\ourstyle} outperforms both MixStyle and DSU by a large margin. 
We also observe that there is a slight performance drop on the source domain (CS) when applying DSU and MixStyle.
As they operate at the feature-level, there is no guarantee that the semantic content stays unchanged after the random perturbation of feature statistics. Thus, the changes in feature statistics might negatively affect the performance, as also indicated in\newcite{li2022dsu}. Note that, in contrast, {\ourstyle} operates on the image space.
Combining {\ourstyle} with MixStyle and DSU leads to a strong boost in performance of these methods. 

%%%%% SotA DG + Data augmetnation - Cityscapes %%%%

\begin{table}[t]
\begin{center}
{\small
    \begin{tabular}{@{}l|c|cccc@{}}
       % \toprule
         Method & CS & ACDC & BDD & DarkZ   \\ \midrule
        Baseline  & 69.01  & 44.23  & 43.27 & 16.03  \\
        RobustNet\newcite{robustnet_2021}   & \textbf{69.47}  & 47.25 & 46.94 & 20.11  \\
        + \ourstyle  & 69.45 &  \textbf{47.55} & \textbf{48.44} & \textbf{23.09}  \\ 
       % \bottomrule
    \end{tabular}
}
\vspace{-0.5em}
\end{center}
\caption{Combination of {\ourstyle} and RobustNet\newcite{robustnet_2021}. We adopt the experimental setting of RobustNet and use DeepLab v3+\newcite{chen2018deeplabv3plus} as the baseline. Our {\ourstyle} is complementary to RobustNet and further improves its generalization performance. }
\label{tab:robustnet-cs-da}
\vspace{-0.6em}
\end{table}

Being model-agnostic, {\ourstyle} can be combined with other networks designed specifically for the domain generalization of semantic segmentation. To showcase its complementary nature, we add {\ourstyle} on top of RobustNet\newcite{robustnet_2021}, which proposed a novel instance whitening loss to selectively remove domain-specific style information.
Although color transformation has already been used for augmentation in RobustNet, {\ourstyle} can introduce more natural style shifts, thus helping to further remove style-specific biases. 
\Cref{tab:robustnet-cs-da} verifies the effectiveness of {\ourstyle}. It brings extra gains for RobustNet, especially in the challenging day to night scenario, i.e., Cityscapes to Dark Z\"urich, boosting the performance from $20.11\%$ to $23.09\%$ in mIoU.

\section{Conclusion}

In this paper, we propose a GAN inversion based data augmentation method {\ourstyle} for learning domain generalized semantic segmentation using restricted training data from a single source domain. The key enabler for {\ourstyle} is the masked noise encoder, which is capable of preserving fine-grained content details and allows style mixing between images without affecting the semantic content. 
Extensive experimental results demonstrate the effectiveness of {\ourstyle} on domain generalization across different datasets and network architectures.

%\clearpage
%\newpage
{\small
\bibliographystyle{ieee_fullname}
\bibliography{docs/reference}
}

%%%%%%%%% Appendix %%%%%%%%
\clearpage
\appendix
% reset the counter
\renewcommand{\thetable}{S.\arabic{table}}  
\renewcommand{\thefigure}{S.\arabic{figure}} 
\renewcommand{\theequation}{S.\arabic{equation}}
\renewcommand{\thesection}{S.\arabic{section}}
\refstepcounter{figure} 
\refstepcounter{table} 
\refstepcounter{equation}
\setcounter{figure}{0}  
\setcounter{table}{0}  
\setcounter{equation}{0}

% Set title for the supplementary material
\newcommand{\abstractTitle}{
\begin{strip}   
\begin{center}
       \Large \bf Intra-Source Style Augmentation for Improved Domain Generalization \\
        \Large \bf Supplementary Material \\
    \vspace*{24pt}
    \large
    \lineskip .5em
        %Anonymous WACV 2023 {\color[rgb]{.9,.1,0.1} \fbox{Algorithms Track}} submission \\
    \vspace*{1pt}
\end{center}
\end{strip}    
}
% Comment this to compile the paper
\abstractTitle

% Overview
This supplementary material to the main paper is structured as follows:
\begin{itemize}
    \item In \cref{sec:appendix-encoder} Masked Noise Encoder, we include an ablation study on noise masking, and noise map resolution. More qualitative results of the encoder comparison are provided to supplement Fig. \textcolor{red}{2} in the main text. Additionally, we provide information regarding the computational complexity. 
    \item In \cref{sec:appendix-dg} Domain Generalization, we present a detailed quantitative comparison with other data augmentation techniques, as well as visual results of the semantic segmentation. {\ourstyle} improves the model's generalization performance, supporting the results of Table \textcolor{red}{3} in the main paper. We also demonstrate the plug-n-play ability of ISSA.
    
    \item In \cref{sec:appendix-uda} Comparison with Unsupervised Domain
Adaptation Methods, we show that {\ourstyle} is competitive with unsupervised domain adaptation methods, even though it does not have access to the target domain data.
   
    \item In \cref{sec:appendix-limitations} Limitations and Future Work, we provide the discussion on the limitations and future directions of the proposed method.
\end{itemize}

\section{Masked Noise Encoder} \label{sec:appendix-encoder}

%%%%%%%%%%%%%%%%%%%%%%%%%%%%%%%%%%%
\paragraph{Ablation on noise random masking} 
We conduct an ablation study on the mask patch size $P$ and masking ratio $\rho$, shown in \cref{tab:gan-inversion-mask-ablation}. We observe that the patch size $P=4$ with a masking ratio $\rho= 25\%$ achieves the best reconstruction performance. Therefore, we use
the encoder trained with this parameter combination for our data augmentation {\ourstyle}.

\paragraph{Ablation on the noise map resolution}
\newchange{
We investigate the effect of noise map size and experimentally observed that the reconstruction quality benefits the most from using the noise map at the intermediate feature space with one fourth of the input resolution.
As shown in~\cref{tab:noise-map}, using $32 \times 64 $ noise, i.e., one fourth of the image resolution, achieves better reconstruction quality than using lower resolution noise maps.
Higher resolution noise map, e.g., full image resolution, in contrast, can be too expressive and encode nearly all perceivable details. This results in worse style mixing capability, as shown in \cref{fig:noise-res-comp2}. 
Therefore, we employ the intermediate noise map at one fourth of the input resolution in all of our experiments.
}

%%%%% Masked Encoder Ablation Study %%%%
\begin{table}[t]
\begin{center}
{\small
    \begin{tabular}{@{}cc|ccc@{}}
        \multicolumn{1}{l}{Patch size} & Ratio & MSE $\downarrow$ & LPIPS $\downarrow$ & FID $\downarrow$ \\ \midrule
        \multirow{2}
        {*}{2} & 25\% & 0.005 & 0.090 & 1.50 \\
               & 50\% & 0.008 & 0.127 & 2.02 \\
        \midrule
        \multirow{2}
        {*}{4} & 25\% & \textbf{0.004} & \textbf{0.089} & \textbf{1.41} \\
               & 50\% & 0.009 & 0.129 & 2.01 \\ 
    \end{tabular}
}
\vspace{-0.5em}
\end{center}
\caption{Ablation on the mask patch size and masking ratio. The influence of patch size is minor on the reconstruction, while masking ratio is more important, i.e., higher masking ratio has negative impact. }
\label{tab:gan-inversion-mask-ablation}
\vspace{-0.5em}
\end{table}

\begin{table}[t]
\setlength{\tabcolsep}{0.5em}
\renewcommand{\arraystretch}{0.8}
\begin{center}
{\small
    {
    \begin{tabular}{l|lcc}
        Noise scale & MSE $\downarrow$ & LPIPS $\downarrow$  & FID $\downarrow$  \\
    \midrule
        $4\times8 \sim 8\times16$ & 0.041 & 0.317 & 14.90  \\
        $32\times64$ & 0.008  & 0.101 & 2.30  \\
    \end{tabular}
    }
}
\end{center}
\vspace{-0.5em}
\caption{Effect of noise map resolution on reconstruction quality. Experiments are done on Cityscapes, $128\times 256$ resolution.
}
\label{tab:noise-map}
\vspace{-0.5em}
\end{table}

\begin{figure*}[t]
\begin{centering}
\setlength{\tabcolsep}{0.0em}
\renewcommand{\arraystretch}{0}
\par\end{centering}
\begin{centering}
\hfill{}
	\begin{tabular}{@{}c@{}c@{}c@{}c@{}c}
			\centering
	Content & Style 
	& $\frac{H}{16}\times \frac{W}{16}$  
	& $\frac{H}{4}\times \frac{W}{4}$ (Ours)
	& $H\times W$
	\vspace{0.02cm} \tabularnewline
	\includegraphics[width=0.195\linewidth]{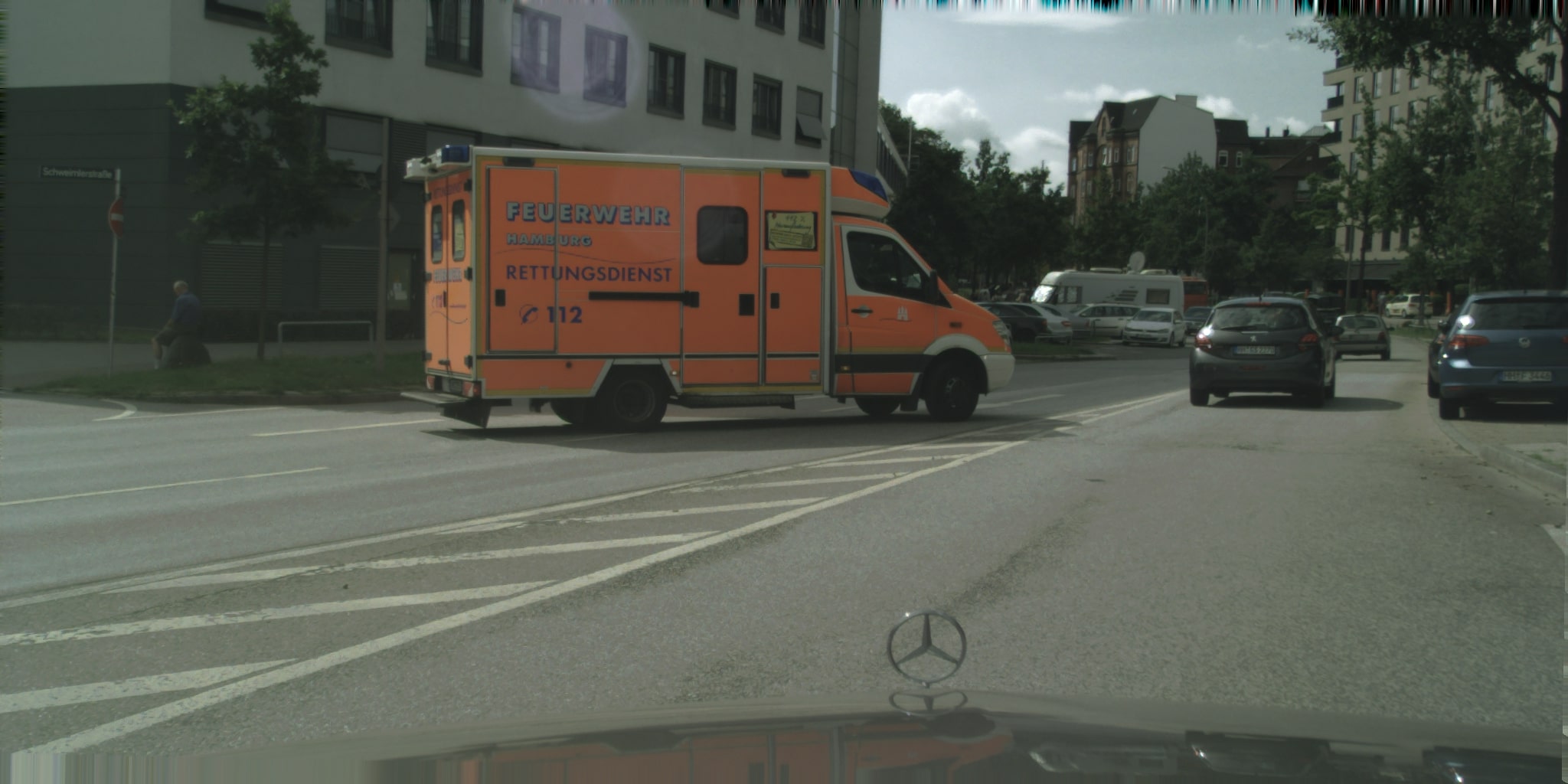}
	      & {\footnotesize{}}
	\includegraphics[width=0.195\linewidth]{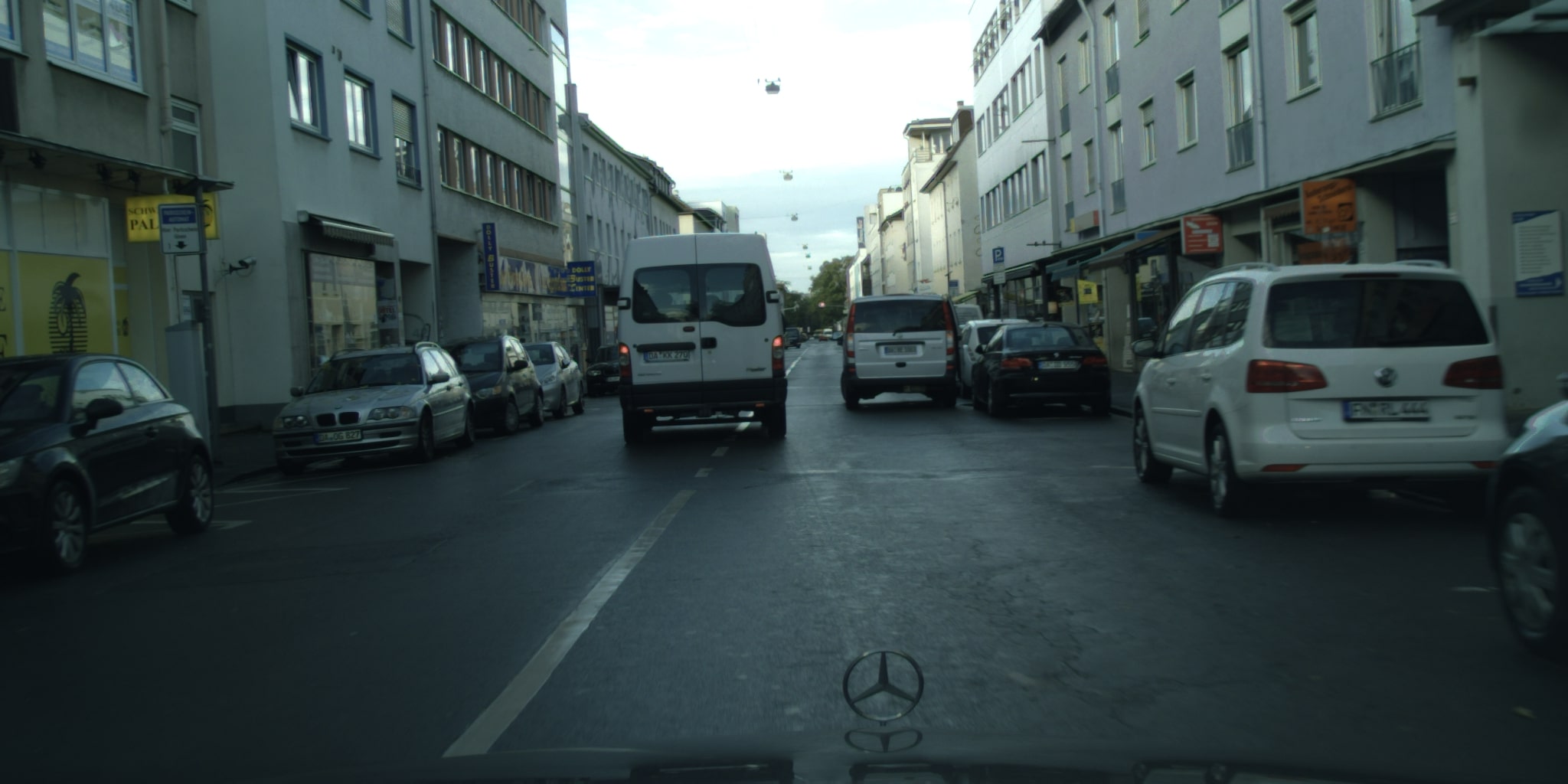}
	        & {\footnotesize{}}
	\includegraphics[width=0.195\linewidth]{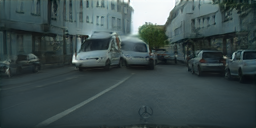} 
		    & {\footnotesize{}}
	\includegraphics[width=0.195\linewidth]{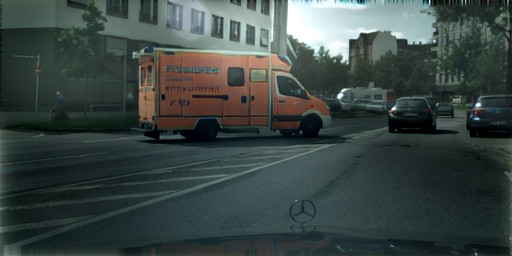}
		& {\footnotesize{}}
	\includegraphics[width=0.195\linewidth]{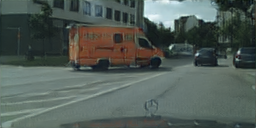}
	 \tabularnewline
	\end{tabular}
\hfill{}
\par\end{centering}
\caption{Influence of the  noise map resolution on style-mixing ability. Using higher resolution noise map, e.g, $H\times W$, leads to poor style-mixing ability. While too low resolution, e.g., $\frac{H}{16}\times \frac{W}{16}$, cannot reconstruct the scene faithfully.
} 
\label{fig:noise-res-comp2}
\vspace{-0.7em}
\end{figure*}
\begin{figure*}[t]
\begin{centering}
\setlength{\tabcolsep}{0.0em}
\renewcommand{\arraystretch}{0}
\par\end{centering}
\begin{centering}
\hfill{}
	\begin{tabular}{@{\hspace{-0.3em}}c@{\hspace{0.1em}}c@{}c@{}c@{}c}
			\centering
		&   &   &   &\vspace{0.01cm} \tabularnewline
		
	\multirow{1}{*}{ \rotatebox{90}{\hspace{3.5em}  Input \hspace{-3.5em} }} &	    
	\begin{tikzpicture}
            \node [
	        above right,
	        inner sep=0] (image) at (0,0) {\includegraphics[width=0.239\textwidth,]{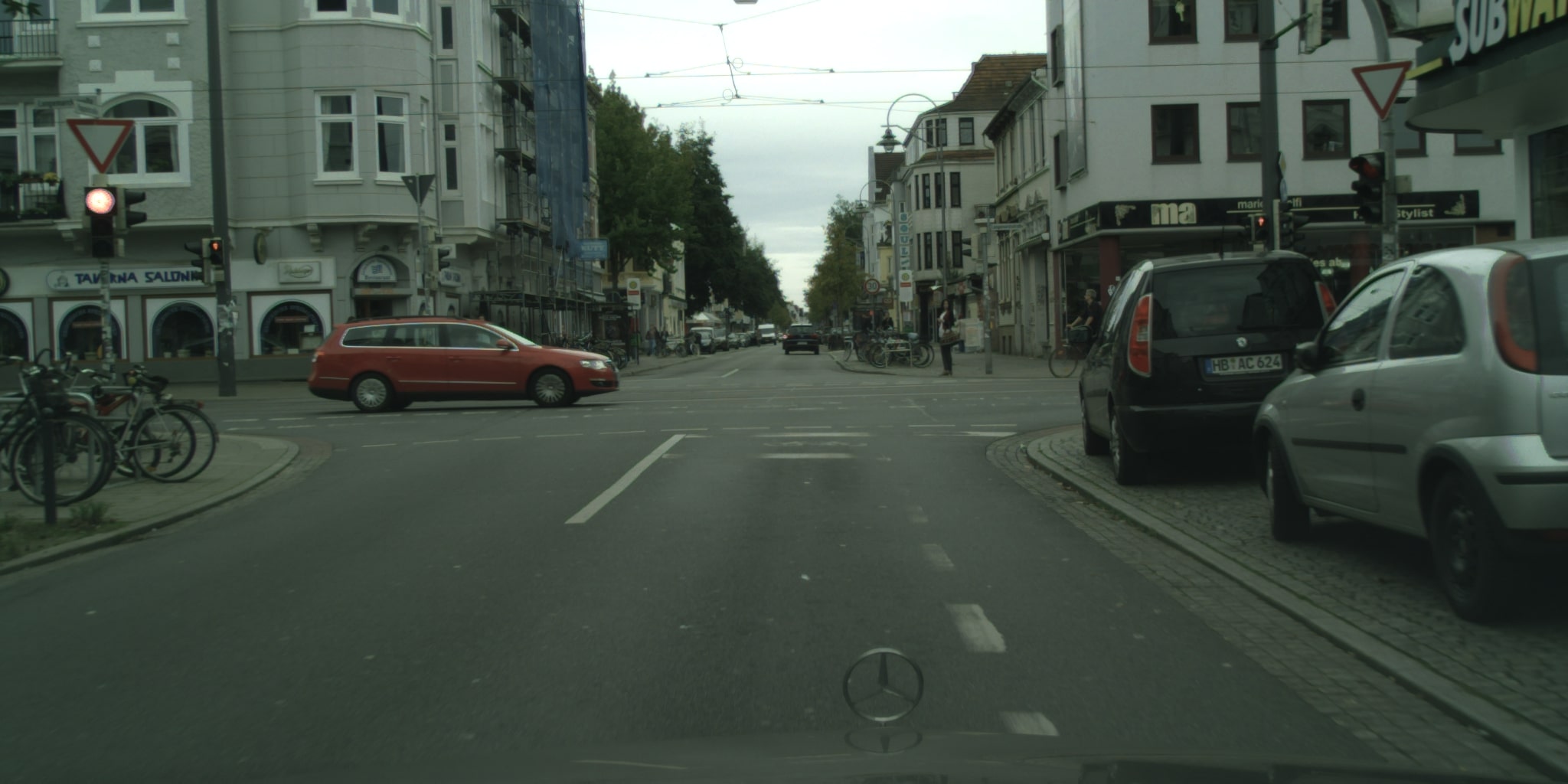} };
            \begin{scope}[
            x={($0.1*(image.south east)$)},
            y={($0.1*(image.north west)$)}]
            \draw[thick,green] (0.01,3) rectangle (4.2,8.4) ;
        \end{scope}
    \end{tikzpicture}
	        & {\footnotesize{}}
	 \begin{tikzpicture}
            % Include the image in a node
            \node [
	        above right,
	        inner sep=0] (image) at (0,0) {\includegraphics[width=0.239\textwidth,]{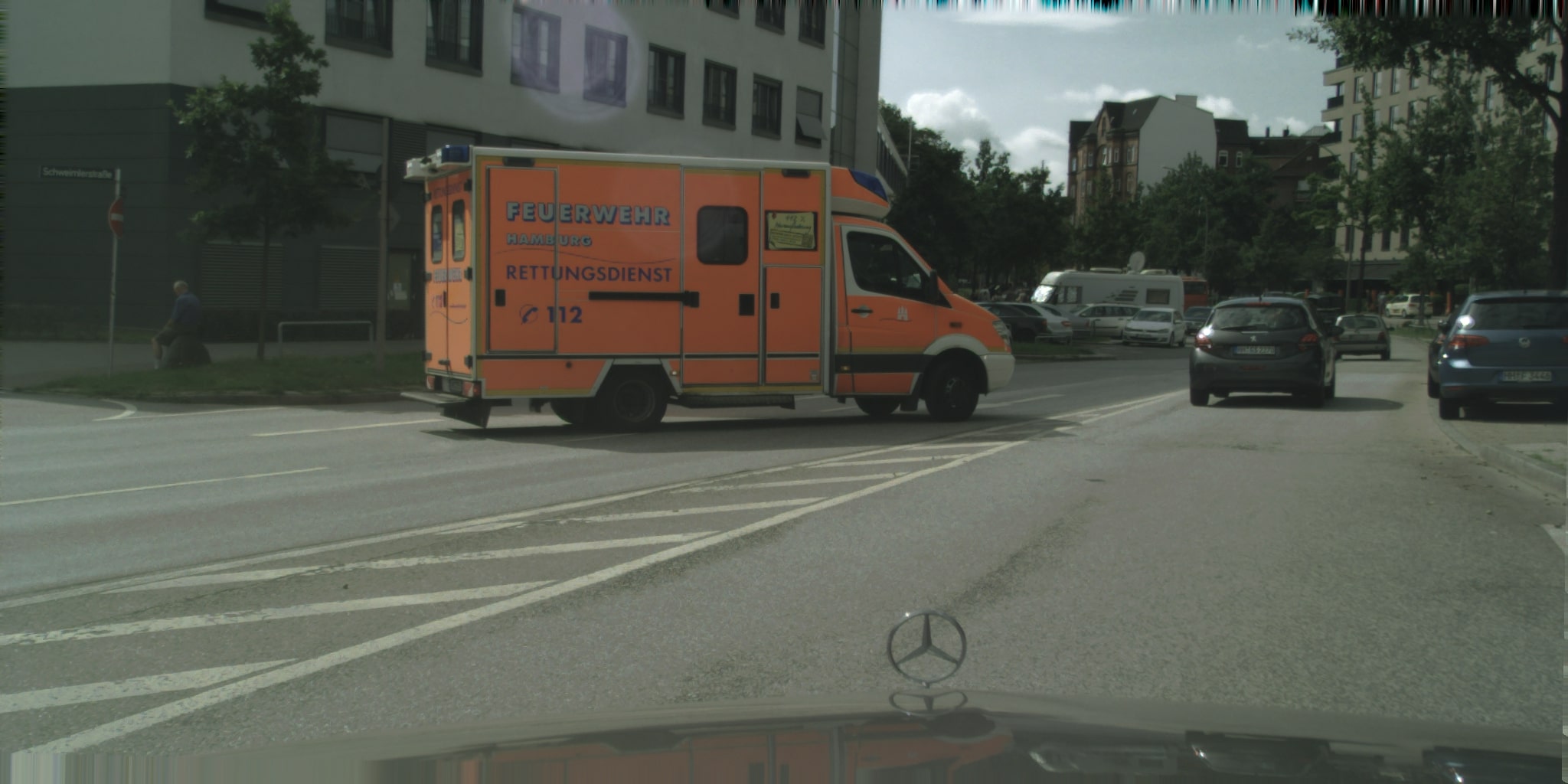}  };
            % Create scope with normalized axes
            \begin{scope}[
            x={($0.1*(image.south east)$)},
            y={($0.1*(image.north west)$)}]
            \draw[thick,green] (2.5,4.2) rectangle (6.7,8.5) ;
        \end{scope}
    \end{tikzpicture}       
	        & {\footnotesize{}}
	        
	 \begin{tikzpicture}
            % Include the image in a node
            \node [
	        above right,
	        inner sep=0] (image) at (0,0) {\includegraphics[width=0.239\textwidth,]{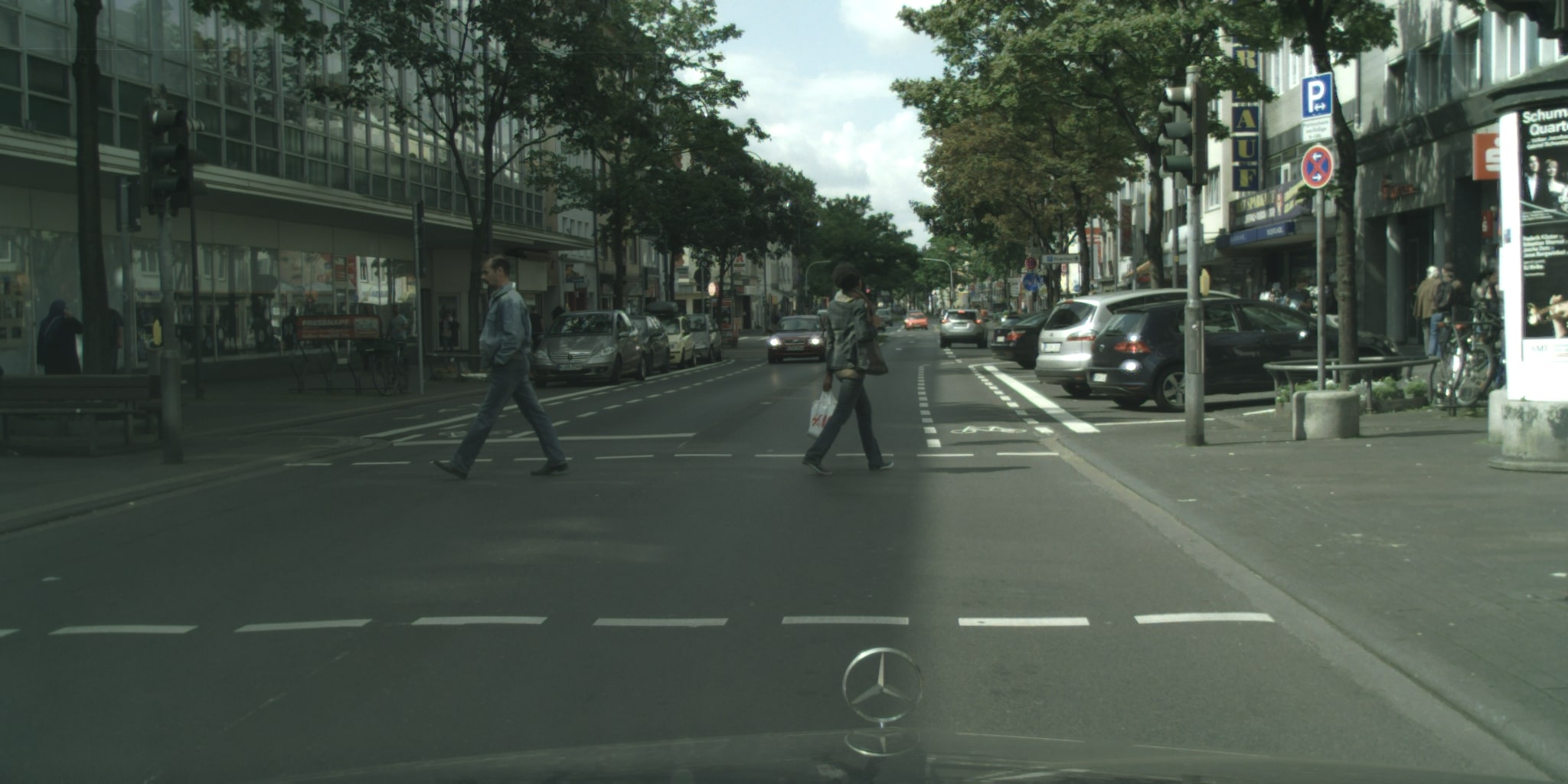}  };
            % Create scope with normalized axes
            \begin{scope}[
            x={($0.1*(image.south east)$)},
            y={($0.1*(image.north west)$)}]
            \draw[thick,green] (2.5,3.5) rectangle (6.7,7) ;
        \end{scope}
    \end{tikzpicture}   
		    & {\footnotesize{}}
		    
    \begin{tikzpicture}
            % Include the image in a node
            \node [
	        above right,
	        inner sep=0] (image) at (0,0) {\includegraphics[width=0.239\textwidth,]{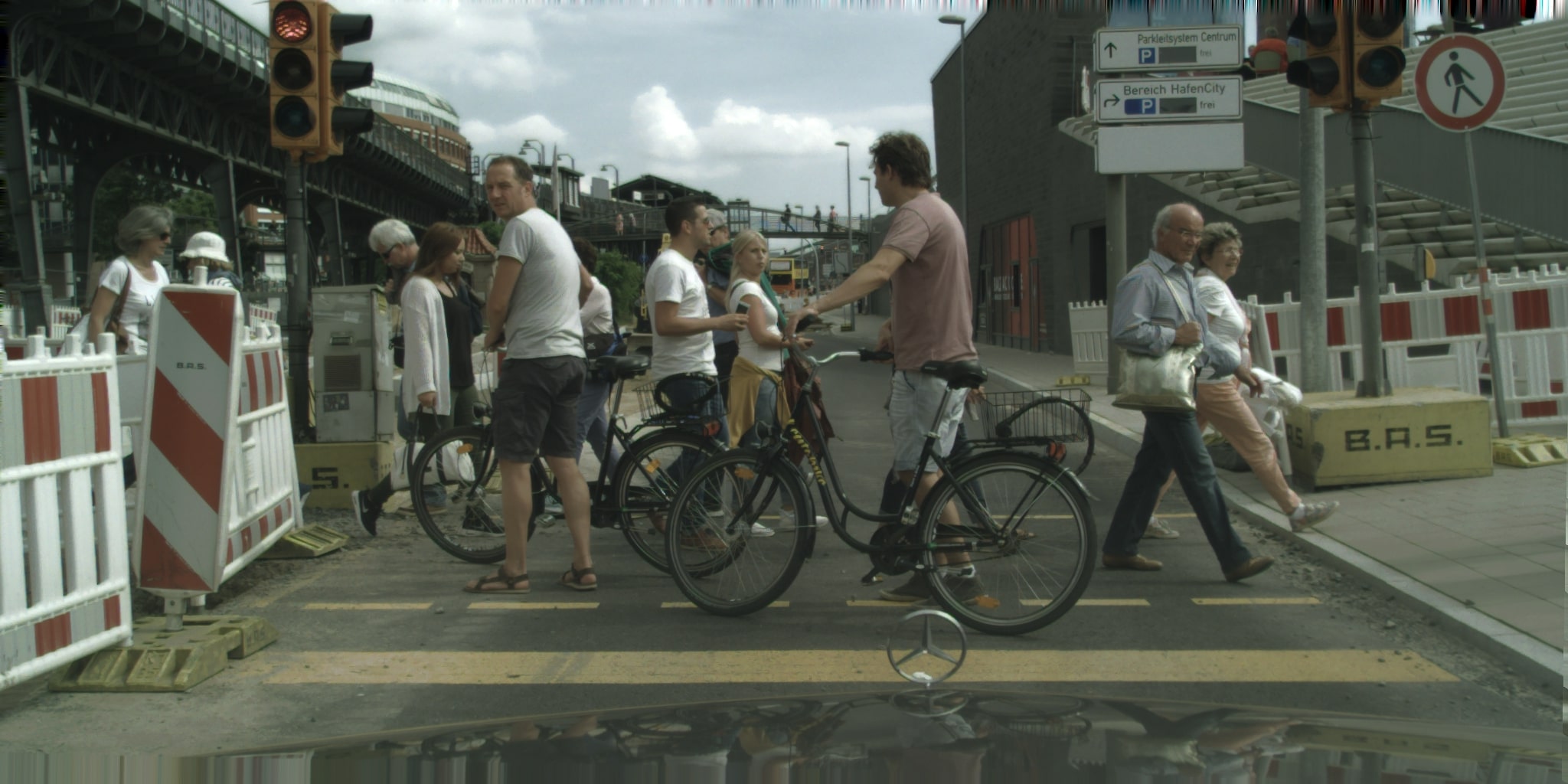}  };
            % Create scope with normalized axes
            \begin{scope}[
            x={($0.1*(image.south east)$)},
            y={($0.1*(image.north west)$)}]
            \draw[thick,green] (1.5,7.8) rectangle (2.5,9.99) ;
            \draw[thick,green] (8.7,8) rectangle (9.9,9.7) ;
        \end{scope}
    \end{tikzpicture} 
	 \tabularnewline	
	 
	 \multirow{1}{*}{ \rotatebox{90}{\hspace{3.5em}  pSp \hspace{-3.5em} }} &	 
	 \begin{tikzpicture}
            % Include the image in a node
            \node [
	        above right,
	        inner sep=0] (image) at (0,0) {\includegraphics[width=0.239\textwidth,]{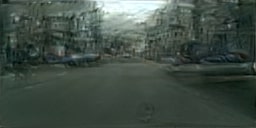}  };
            % Create scope with normalized axes
            \begin{scope}[
            x={($0.1*(image.south east)$)},
            y={($0.1*(image.north west)$)}]
            \draw[thick,red] (0.01,3) rectangle (4.2,8.4) ;
        \end{scope}
    \end{tikzpicture}
	        & {\footnotesize{}}
\begin{tikzpicture}
            % Include the image in a node
            \node [
	        above right,
	        inner sep=0] (image) at (0,0) {\includegraphics[width=0.239\textwidth,]{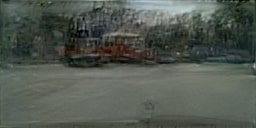}  };
            % Create scope with normalized axes
            \begin{scope}[
            x={($0.1*(image.south east)$)},
            y={($0.1*(image.north west)$)}]
            \draw[thick,red] (2.5,4.2) rectangle (6.7,8.5) ;
        \end{scope}
    \end{tikzpicture}	  
	        & {\footnotesize{}}
\begin{tikzpicture}
            % Include the image in a node
            \node [
	        above right,
	        inner sep=0] (image) at (0,0) {\includegraphics[width=0.239\textwidth,]{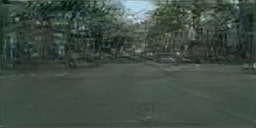} };
            % Create scope with normalized axes
            \begin{scope}[
            x={($0.1*(image.south east)$)},
            y={($0.1*(image.north west)$)}]
            \draw[thick,red] (2.5,3.5) rectangle (6.7,7) ;
        \end{scope}
    \end{tikzpicture}
		    & {\footnotesize{}}
    \begin{tikzpicture}
            % Include the image in a node
            \node [
	        above right,
	        inner sep=0] (image) at (0,0) {		\includegraphics[width=0.239\textwidth,]{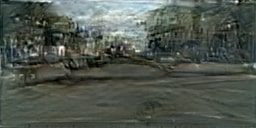}};
            % Create scope with normalized axes
            \begin{scope}[
            x={($0.1*(image.south east)$)},
            y={($0.1*(image.north west)$)}]
            \draw[thick,red] (1.5,7.8) rectangle (2.5,9.99) ;
            \draw[thick,red] (8.7,8) rectangle (9.9,9.7) ;
        \end{scope}
    \end{tikzpicture}
	 \tabularnewline
	
	\multirow{1}{*}{ \rotatebox{90}{\hspace{3.5em}  pSp${}^\dagger$ \hspace{-3.5em} }} &
	\begin{tikzpicture}
            % Include the image in a node
            \node [
	        above right,
	        inner sep=0] (image) at (0,0) {	\includegraphics[width=0.239\textwidth,]{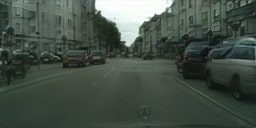} };
            % Create scope with normalized axes
            \begin{scope}[
            x={($0.1*(image.south east)$)},
            y={($0.1*(image.north west)$)}]
            \draw[thick,red] (0.01,3) rectangle (4.2,8.4) ;
        \end{scope}
    \end{tikzpicture}
	        & {\footnotesize{}}
	  \begin{tikzpicture}
            % Include the image in a node
            \node [
	        above right,
	        inner sep=0] (image) at (0,0) {\includegraphics[width=0.239\textwidth,]{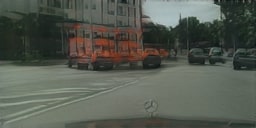} };
            % Create scope with normalized axes
            \begin{scope}[
            x={($0.1*(image.south east)$)},
            y={($0.1*(image.north west)$)}]
            \draw[thick,red] (2.5,4.2) rectangle (6.7,8.5) ;
        \end{scope}
    \end{tikzpicture}
	        & {\footnotesize{}}
	 \begin{tikzpicture}
            \node [
	        above right,
	        inner sep=0] (image) at (0,0) {\includegraphics[width=0.239\textwidth,]{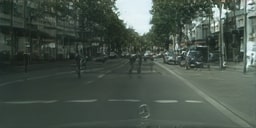}};
            \begin{scope}[
            x={($0.1*(image.south east)$)},
            y={($0.1*(image.north west)$)}]
            \draw[thick,red] (2.5,3.5) rectangle (6.7,7) ;
        \end{scope}
    \end{tikzpicture}
		    & {\footnotesize{}}
	\begin{tikzpicture}
            \node [
	        above right,
	        inner sep=0] (image) at (0,0) {	\includegraphics[width=0.239\textwidth,]{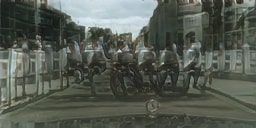}};
            \begin{scope}[
            x={($0.1*(image.south east)$)},
            y={($0.1*(image.north west)$)}]
            \draw[thick,red] (1.5,7.8) rectangle (2.5,9.99) ;
            \draw[thick,red] (8.7,8) rectangle (9.9,9.7) ;
        \end{scope}
    \end{tikzpicture}
	
	 \tabularnewline
		
	\multirow{1}{*}{ \rotatebox{90}{\hspace{3.5em}  \small{Feature-Style} \hspace{-5.5em} }} &
	\begin{tikzpicture}
            \node [
	        above right,
	        inner sep=0] (image) at (0,0) {\includegraphics[width=0.239\textwidth,]{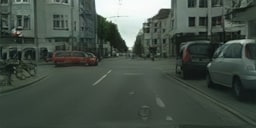}   };
            \begin{scope}[
            x={($0.1*(image.south east)$)},
            y={($0.1*(image.north west)$)}]
            \draw[thick,red] (0.01,3) rectangle (4.2,8.4) ;
        \end{scope}
    \end{tikzpicture}
	
	        & {\footnotesize{}}
	  \begin{tikzpicture}
            \node [
	        above right,
	        inner sep=0] (image) at (0,0) {\includegraphics[width=0.239\textwidth,]{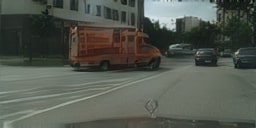} };
            \begin{scope}[
            x={($0.1*(image.south east)$)},
            y={($0.1*(image.north west)$)}]
            \draw[thick,red] (2.5,4.2) rectangle (6.7,8.5) ;
        \end{scope}
    \end{tikzpicture}
	        & {\footnotesize{}}
	 \begin{tikzpicture}
            \node [
	        above right,
	        inner sep=0] (image) at (0,0) {\includegraphics[width=0.239\textwidth,]{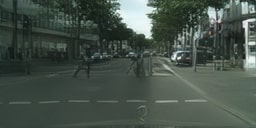}  };
            \begin{scope}[
            x={($0.1*(image.south east)$)},
            y={($0.1*(image.north west)$)}]
            \draw[thick,red] (2.5,3.5) rectangle (6.7,7) ;
        \end{scope}
    \end{tikzpicture}
		    & {\footnotesize{}}
	\begin{tikzpicture}
            \node [
	        above right,
	        inner sep=0] (image) at (0,0) {\includegraphics[width=0.239\textwidth,]{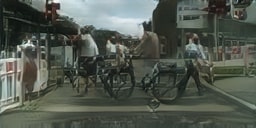} };
            \begin{scope}[
            x={($0.1*(image.south east)$)},
            y={($0.1*(image.north west)$)}]
            \draw[thick,red] (1.5,7.8) rectangle (2.5,9.99) ;
            \draw[thick,red] (8.7,8) rectangle (9.9,9.7) ;
        \end{scope}
    \end{tikzpicture}
		
	 \tabularnewline
	 
	\multirow{1}{*}{ \rotatebox{90}{\hspace{3.5em}  \small{Ours} \hspace{-3.5em} }} &	        
	\begin{tikzpicture}
            \node [
	        above right,
	        inner sep=0] (image) at (0,0) {\includegraphics[width=0.239\textwidth,]{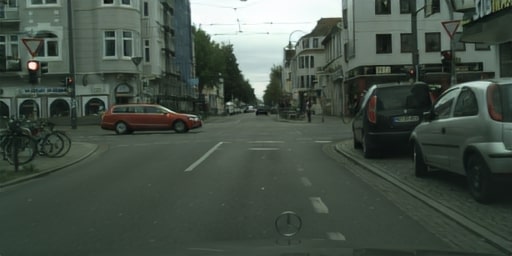}};
            \begin{scope}[
            x={($0.1*(image.south east)$)},
            y={($0.1*(image.north west)$)}]
            \draw[thick,green] (0.01,3) rectangle (4.2,8.4) ;
        \end{scope}
    \end{tikzpicture}
	        & {\footnotesize{}}
	 \begin{tikzpicture}
            \node [
	        above right,
	        inner sep=0] (image) at (0,0) {\includegraphics[width=0.239\textwidth,]{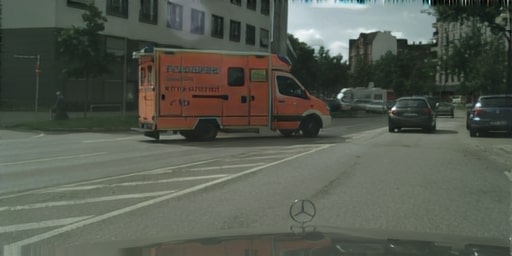} };
            \begin{scope}[
            x={($0.1*(image.south east)$)},
            y={($0.1*(image.north west)$)}]
            \draw[thick,green] (2.5,4.2) rectangle (6.7,8.5) ;
        \end{scope}
\end{tikzpicture}
	        & {\footnotesize{}}
	\begin{tikzpicture}
            \node [
	        above right,
	        inner sep=0] (image) at (0,0) {\includegraphics[width=0.239\textwidth,]{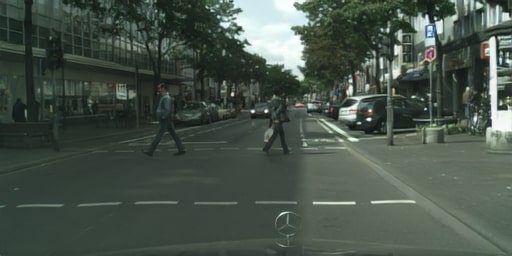} };
            \begin{scope}[
            x={($0.1*(image.south east)$)},
            y={($0.1*(image.north west)$)}]
            \draw[thick,green] (2.5,3.5) rectangle (6.7,7) ;
        \end{scope}
    \end{tikzpicture}
		    & {\footnotesize{}}
	\begin{tikzpicture}
            \node [
	        above right,
	        inner sep=0] (image) at (0,0) {\includegraphics[width=0.239\textwidth,]{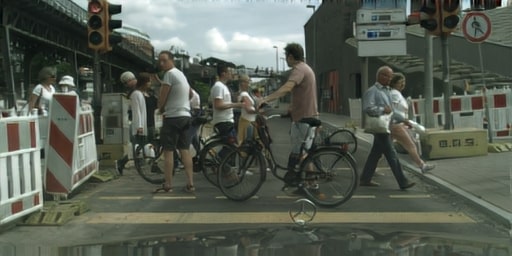}};
            \begin{scope}[
            x={($0.1*(image.south east)$)},
            y={($0.1*(image.north west)$)}]
            \draw[thick,green] (1.5,7.8) rectangle (2.5,9.99) ;
            \draw[thick,green] (8.7,8) rectangle (9.9,9.7) ;
        \end{scope}
\end{tikzpicture}
	 \tabularnewline
	\end{tabular}
\hfill{}
\par\end{centering}
\vspace{-0.5em}
\caption{Qualitative comparison between our masked noise encoder and other StyleGAN2 inversion encoders on Cityscapes (best view in color and zoom  in). Note, $\text{pSp}^\dagger$ is obtained by us, training pSp with an additional discriminator and incorporate synthesized images for better initialization. Evidently, our masked noise encoder achieves the highest fidelity and successfully reconstruct small objects such pedestrians and traffic signs. This is consistent with the observation in Fig. \textcolor{red}{2} of the main text.} 
\label{fig:encoder-visual-appendix}
\vspace{-0.5em}
\end{figure*}
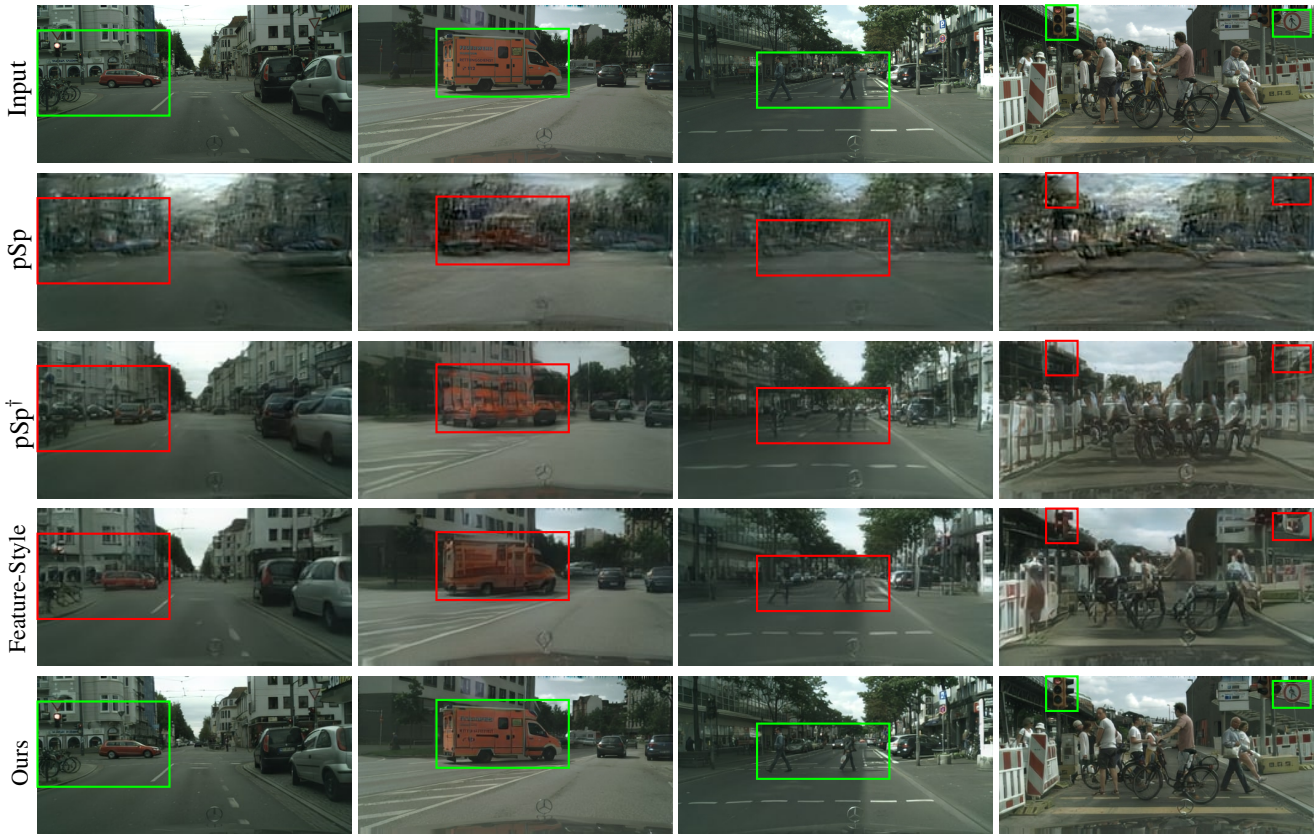

%%%%%%%%%%%%%%%%%%%%%%%%%%%%%%%%%%%
\paragraph{Additional qualitative results}
In \cref{fig:encoder-visual-appendix}
we provide more visual results of the comparison among pSp\newcite{pspencoder}, $\text{pSp}^\dagger$, feature-style encoder\newcite{feature_style} and our masked noise encoder. Note that,
$\text{pSp}^\dagger$ is an improved version obtained by us, which is trained with an additional discriminator and synthesized images for better initialization. It is evident that our masked noise encoder is capable of preserving more fine details and high-quality reconstruction, which is consistent with the observation in Fig. \textcolor{red}{2} in the main text.

%%%%%%%%%%%%%%%%%%%%%%%%%%%%%%%%%%%
\paragraph{Computational complexity} 
We provide more details on the time and memory usage required by using the masked noise encoder. %
It takes around $7$ days to train the masked noise encoder on $256 \times 512$ resolution using $2$ GPUs. A similar amount of time is required for the StyleGAN2 training. Nonetheless, for data augmentation, it only concerns the inference time of our encoder, which is much faster, i.e., 0.1 seconds, compared to optimization based methods such as PTI\newcite{roich2021pti} that takes 55.7 seconds per image. %
Furthermore, stylized images by {\ourstyle} can be pre-generated and pre-stored instead of being generated on-the-fly for data augmentation, reducing the memory usage during the semantic segmentation network training.

%%%%%%%%%%%%%%%%%%%%%%%%%%%%%%%%%%%
\section{Domain Generalization}\label{sec:appendix-dg}

%%%%%%%%%%%%%%%%%%%%%%%%%%%%%%%%%%%
\paragraph{Comparison with data augmentation methods}
\cref{tab:hrnet-segformer-cityscapes-dg-appendix} provides the full comparison on Cityscapes to ACDC domain generalization between  {\ourstyle} and other data augmentation methods, e.g.,  CutMix\newcite{yun2019cutmix}, Hendrycks corruptions\newcite{hendrycks2018benchmarking} and StyleMix\newcite{hong2021stylemix}. Two semantic segmentation models  HRNet{\hrnet} and SegFormer{\segformer} are used. We report more generalization results on BDD100K and Dark Z\"urich in \cref{tab:hrnet-segformer-cityscapes-all-dg-appendix}. Supporting results in Table \textcolor{red}{3} of the main paper, {\ourstyle} has shown consistent improvements on models' generalization capability across datasets and network architectures. We also observe that, among different Hendrycks corruption types, noise and blur corruptions have larger negative impact on the performance, while weather and digital corruptions can offer little help on the generalization performance.

%%%%% SemSeg + Data augmetnation - Cityscapes%%%%
\begin{table*}[t]
\begin{center}
{\small %\footnotesize
    \begin{tabular}{@{}l|c|ccccc||c|ccccc @{}}
        %\toprule
         & \multicolumn{6}{c||}{HRNet\hrnet} & \multicolumn{6}{c}{SegFormer\newcite{xie2021segformer}} \\ 
        Method & CS & Rain & Fog & Snow & Night & Avg. & CS & Rain & Fog & Snow & Night & Avg.\\ \midrule
        
        Baseline  & 70.47 & 44.15 & 58.68 & 44.20 & 18.90 & 41.48  & 67.90 & 50.22 & 60.52 & 48.86 & 28.56 & 47.04\\ 
        \midrule
        CutMix\newcite{yun2019cutmix}   & \textbf{72.68} & 42.48 & 58.63 & 44.50 & 17.07 & 40.67 & \textbf{69.23} & 49.53 & 61.58 & 47.42 & 27.77 & 46.57 \\
        Weather\newcite{hendrycks2018benchmarking}   & 69.25 & \textbf{50.78} & 60.82 & 38.34 & 22.82 & 43.19  & 67.41 & 54.02 & 64.74 & 49.57 & 28.50 & 49.21 \\
        Noise\newcite{hendrycks2018benchmarking}     & 65.78 & 42.45 & 54.60 & 41.64 & 16.31 & 38.75 & 65.89 & 53.15 & 63.88 & 46.63 & 27.66 & 47.83 \\
        Digital\newcite{hendrycks2018benchmarking}   & 69.13 & 50.13 & 65.71 & 49.22 & 24.81 & 47.47 & 67.57 & 55.53 & 66.46 & 49.92 & 30.33 & 50.56\\
        Blur\newcite{hendrycks2018benchmarking}      & 65.95 & 44.05 & 51.22 & 40.19 & 16.83 & 38.07 & 66.15 & 51.17 & 61.57 & 45.71 & 27.49 & 46.48  \\
        Common\newcite{hendrycks2018benchmarking}    & 68.68 & 52.00 & 62.33 & 43.42 & 21.78 & 44.88 & 67.26 & 55.63 & 66.78 & 48.50 & 32.63 & 50.89\\
        StyleMix\newcite{hong2021stylemix} & 57.40 & 40.59 & 49.11  & 39.14 & 19.34 & 37.04 & 65.30 & 53.54 & 63.86 & 49.98 & 28.93 & 49.08\\ 
        {\ourstylebf} \textbf{(Ours)}  & 70.30 & 50.62 & \textbf{66.09} & \textbf{53.30} & \textbf{30.18} & \textbf{50.05} & 67.52 & \textbf{55.91} & \textbf{67.46} & \textbf{53.19} & \textbf{33.23} & \textbf{52.45}  \\
        \midrule
        {\ourstyle}+CutMix    & 72.37 & 53.42 & 68.88 & 53.82 & 30.10 & 51.55 & 68.43 & 55.85 & 68.70 & 52.98 & 33.82 & 52.84\\
        \midrule
        Oracle & 70.29 & 65.67 & 75.22 & 72.34 & 50.39 & 65.90  & 68.24 & 63.67 & 74.10 & 67.97 & 48.79 & 63.56 \\
    \end{tabular}
}
\end{center}
\caption{
Comparison of data augmentation for improving domain generalization, i.e., from Cityscapes (train) to ACDC (unseen). The mean Intersection over Union (mIoU) is reported on Cityscapes (CS), four individual scenarios of ACDC (Rain, Fog, Snow and Night) and the whole ACDC (Avg.). %
Oracle indicates the supervised training on both Cityscapes and ACDC, serving as an mIoU upper bound on ACDC for the other methods. Note, it is not supposed to be an upper bound on Cityscapes. {\ourstyle} performs the best on ACDC using both HRNet and SegFormer, consistently improving the mIoU in all four scenarios of ACDC. This table complements Table \textcolor{red}{3} of the main paper with additional types of Hendrycks' corruption types, i.e., noise, digital and blur. Additionally, we combine {\ourstyle} with CutMix to diversify both styles and content of the training samples, where CutMix brings performance gain on the source domain. 
}
\label{tab:hrnet-segformer-cityscapes-dg-appendix}
\end{table*}
%%%%% SemSeg + Data augmetnation - Cityscapes%%%%
\begin{table*}[t]
\begin{center}
{\small 
    \begin{tabular}{@{}l|c|ccc||c|ccc @{}}
         & \multicolumn{4}{c||}{HRNet\hrnet} & \multicolumn{4}{c}{SegFormer\newcite{xie2021segformer}} \\ 
        Method & CS & ACDC & BDD100K & Dark Z\"urich  & CS & ACDC & BDD100K & Dark Z\"urich \\ \midrule
        Baseline & 70.47 & 41.48  & 45.66 & 15.50 & 67.90 &  47.04 & 49.35 & 24.20 \\ 
        \midrule
        CutMix\newcite{yun2019cutmix}   & \textbf{72.68} & 40.67 &45.57 & 15.34 & \textbf{69.23} & 46.57 & 48.93 & 22.98\\
        Weather\newcite{hendrycks2018benchmarking}   & 69.25 & 43.19 & 44.53  & 18.71   & 67.41 & 49.21 & 49.84 & 23.44 \\
        Noise\newcite{hendrycks2018benchmarking}     & 65.78 & 38.75 & 44.13 & 12.40 & 65.89 &  47.83 & 49.55 & 22.50 \\
        Digital\newcite{hendrycks2018benchmarking}   & 69.13 &  47.47 & 47.60 & 22.32 & 67.57 & 50.56 & 51.11 & 25.11\\
        Blur\newcite{hendrycks2018benchmarking}  & 65.95 & 38.07 & 37.16 & 15.05 & 66.15 & 46.48 & 48.89 & 22.82 \\
        Common\newcite{hendrycks2018benchmarking}  & 68.68 & 44.88& 46.31 & 18.30 & 67.26 & 50.89 & 51.53 & 27.11\\
        StyleMix\newcite{hong2021stylemix} & 57.40 & 37.04 & 39.30 & 15.85 & 65.30 &  49.08 & 50.49 & 23.50 \\ 
        {\ourstylebf} \textbf{(Ours)}    & 70.30 &  \textbf{50.05} & \textbf{50.29} & \textbf{27.24} & 67.52 & \textbf{52.45} & \textbf{51.92} & \textbf{27.39}\\
        \midrule
        {\ourstyle}+CutMix    & 72.37 & 51.55  & 50.06 & 26.24 & 68.43  & 52.84 & 51.89 & 28.29\\
    \end{tabular}
}
\end{center}
\caption{
Comparison of data augmentation for improving domain generalization, i.e., from Cityscapes (train) to ACDC, BDD100K and Dark Z\"urich (unseen). The mean Intersection over Union (mIoU) is reported.
This table supplements the results in Table \textcolor{red}{3} of the main paper.
{\ourstyle} consistently outperforms the other data augmentation techniques across different datasets and network architectures.  
We additionally combine {\ourstyle} with CutMix to diversify both styles and content of the training samples, where CutMix brings performance gain on the source domain.
}
\label{tab:hrnet-segformer-cityscapes-all-dg-appendix}
\vspace{-1em}
\end{table*}
%%%%% SemSeg + Data augmetnation - BDD100K Subset 
\begin{table}[t]
\begin{center}
{\small
    \begin{tabular}{@{}l|c|cccc@{}}
        Method & BDD100K & ACDC-Night & DarkZ\"urich   \\ \midrule
        Baseline\hrnet  & 52.97  & 23.52 & 23.63  \\
        \midrule
        CutMix\newcite{yun2019cutmix}          & \textbf{54.03}  & 24.37 & 23.99  \\
        Weather\newcite{hendrycks2018benchmarking}          & 52.10  & 23.79 &  24.21 \\
        Noise\newcite{hendrycks2018benchmarking}           & 49.25 & 19.69 & 19.31 \\
        Blur\newcite{hendrycks2018benchmarking}   & 50.92  & 20.68 & 20.08  \\
        Digital\newcite{hendrycks2018benchmarking}          & 52.10   & 24.17 & 23.24 \\
        Common\newcite{hendrycks2018benchmarking}        & 51.34 & 23.76 & 23.62 \\
        StyleMix\newcite{hong2021stylemix} & 46.33 & 19.13 & 19.27 \\ 
        \ourstylebf \textbf{(Ours)} & 53.37 & \textbf{25.93} & \textbf{26.55} \\
    \end{tabular}
}
\end{center}
\caption{Comparison of data augmentation techniques for improving domain generalization using HRNet\hrnet, i.e., from BDD100K-Daytime to ACDC-Night and Dark Z\"urich. BDD100K-Daytime is a subset of BDD100K, which contains $2526$ images in daytime under various weather conditions, but not in dawn/nighttime. Here, we evaluate the domain generalization with respect to day to night.}
\label{tab:hrnet-bdd-daytime-da}
\end{table}
%%%%% BDDGAN-ISSA on Cityscapes %%%%
\begin{table}[t]
\setlength{\tabcolsep}{0.5em}
\renewcommand{\arraystretch}{1.0}
\begin{center}
{\small
    \begin{tabular}{@{}l|c|ccccc@{}}
        Method  & CS & Rain & Fog & Snow & Night & Avg.   \\ \hline
        Baseline & \textbf{70.5} & 44.2 & 58.7  &44.2 &  18.9 & 41.5 \\ \hline
        ISSA: CS-G-E & 70.3 & 50.6 & 66.1 & \textbf{53.3} & 30.2 & 50.1  \\
        ISSA: BDD-G-E & 70.3 & \textbf{52.2} & \textbf{66.3} & 52.2 & \textbf{31.0} & \textbf{50.4}
    \end{tabular}
}
\end{center} 
\caption{Comparison on Cityscapes to ACDC generalization using ISSA with generator and encoder trained on Cityscapes (CS-G-E) and BDD100K (BDD-G-E), respectively.
Despite never seeing Cityscapes samples, ISSA with BDD-G-E is still highly effective.
}
\label{tab:bddgan-cityscapes-dg-small}
\vspace{-0.7em}
\end{table}

%%%%% Compare with UDA methods %%%%
\begin{table}[t]
\begin{center}
{\footnotesize 
    {
    \begin{tabular}{l|c|ccc}
        Method & Network & Use Target & mIoU \\
    \midrule
        Baseline & \multirow{10}{*}{DeepLabv2\deeplabvtwo} & \textbf{---} & 30.9 
        \\ \midrule
        BDL\newcite{li2019bdl}  & & \cmark & 32.7 \\
        CRST \newcite{zou2019confidence}  &  & \cmark & 32.8 \\
        AdaptSegNet\newcite{tsai2018AdaptSegNet}  &  & \cmark & 33.4 \\
        SIM\newcite{wang2020differential}  &  & \cmark & 34.6 \\
        MRNet \newcite{zheng2021rectifying}  &   & \cmark & 36.1 \\
        ADVENT\newcite{tsai2019advent}  &  & \cmark & 37.7 \\
        CLAN\newcite{luo2019clan}  &   & \cmark & 39.0 \\
        FDA\newcite{yang2020fda}  &   & \cmark & 45.7 \\
        \ourstyle(Ours) &   & \xmark & 43.2 \\
        \midrule
        DAFormer\newcite{hoyer2022daformer} & DAFormer\newcite{hoyer2022daformer}  & \cmark & 55.4\\
        \ourstyle(Ours) & SegFormer\segformer  & \xmark & 52.5\\
    \end{tabular}
    }
}
\end{center}
\caption{Quantitative comparison on Cityscapes $\rightarrow$ ACDC with UDA methods. Remarkably, our domain generalization method (without access to the target domain, neither images nor labels), is on-par or better than unsupervised domain adaptation (UDA) methods, which requires knowledge of the target domain during training. Results of UDA methods are from\acdc.}
\label{tab:uda-comparison}
\vspace{-1em}
\end{table}
\begin{figure}[t]
    \begin{centering}
    \setlength{\tabcolsep}{0.0em}
    \renewcommand{\arraystretch}{0}
    \par\end{centering}
    \begin{centering}
    \hfill{}%
	\begin{tabular}{@{}c@{}c@{}c}
        \centering
		 Content & Style & Mixed  \tabularnewline
    \includegraphics[width=0.3\linewidth]{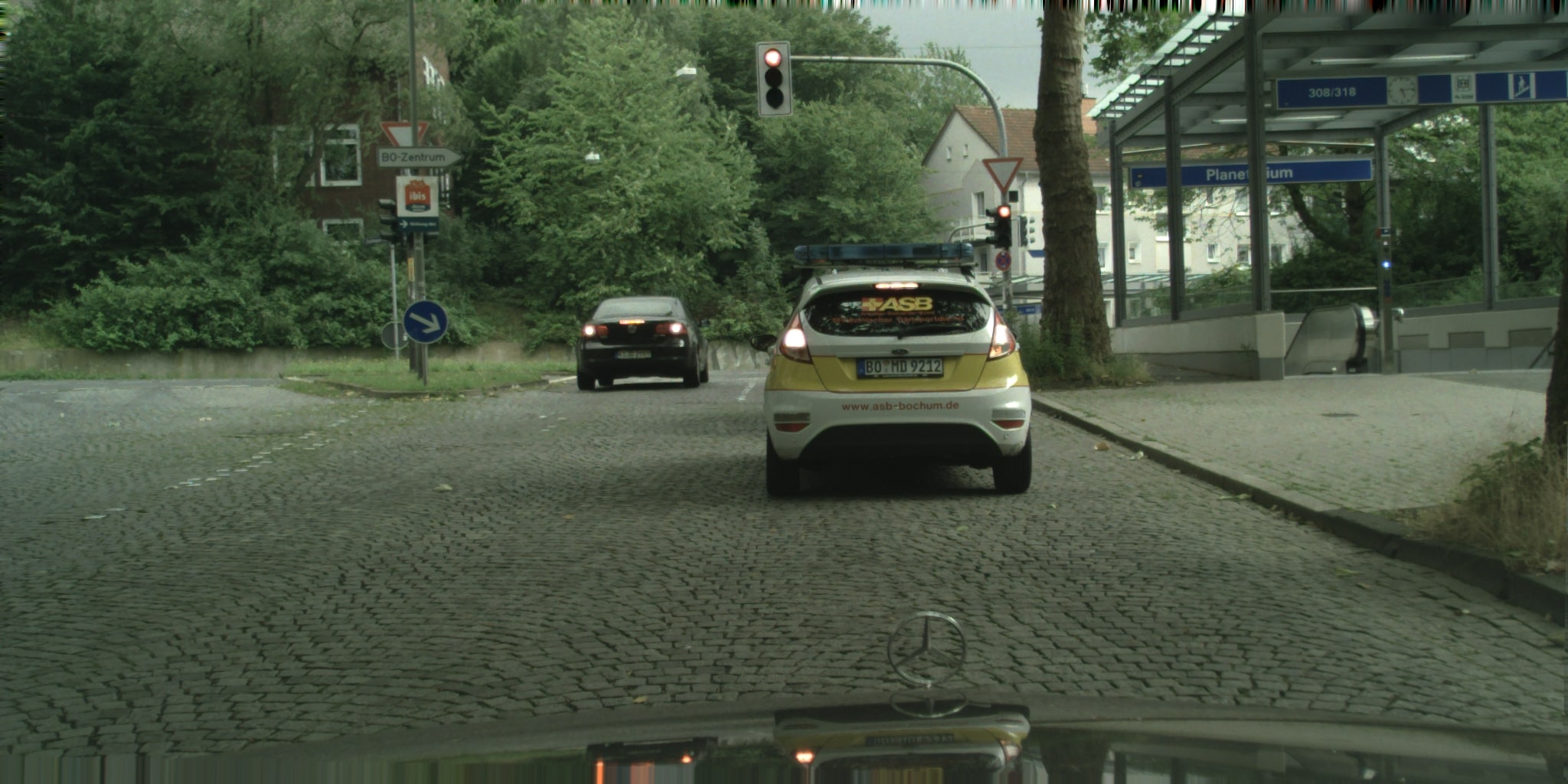} &
	{\footnotesize{}} \includegraphics[width=0.3\linewidth]{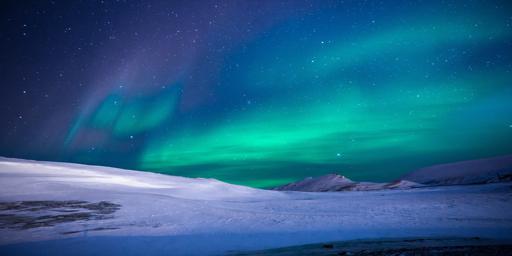} & {\footnotesize{}}
	\includegraphics[width=0.3\linewidth]{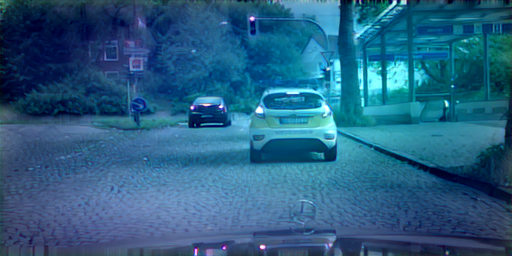} 
	\tabularnewline
	\includegraphics[width=0.3\linewidth]{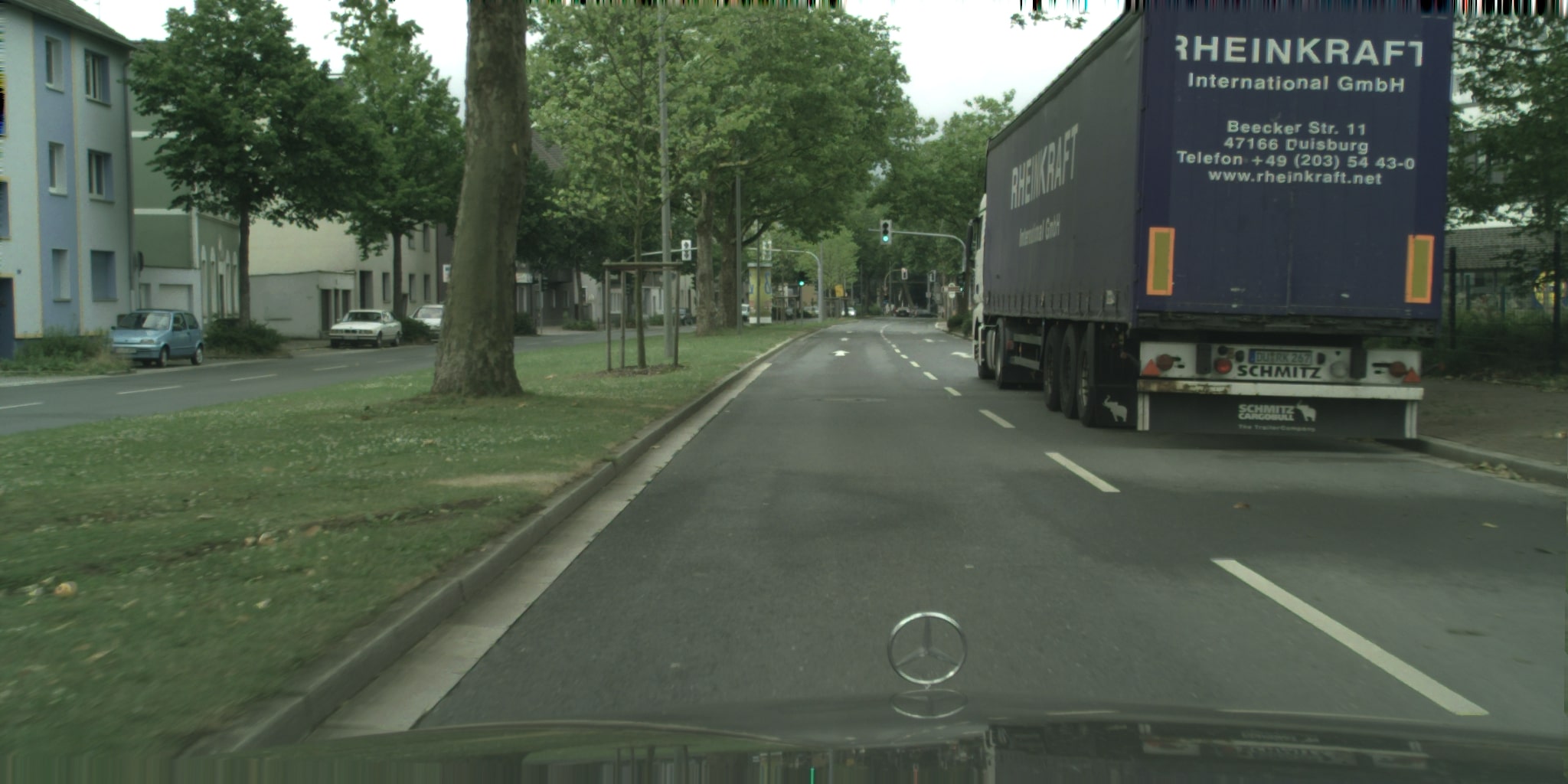} &
    {\footnotesize{}} \includegraphics[width=0.3\linewidth]{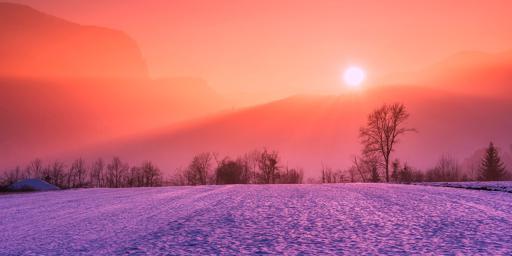} & {\footnotesize{}}
	\includegraphics[width=0.3\linewidth]{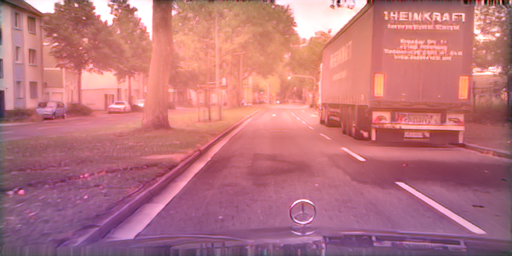} 
	\tabularnewline
	
	\includegraphics[width=0.3\linewidth]{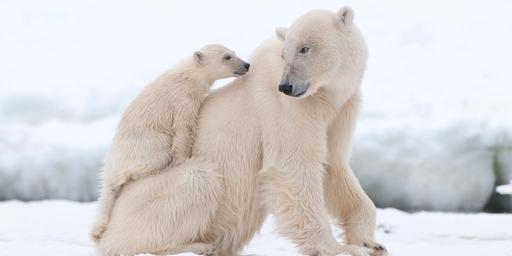} &
	{\footnotesize{}} \includegraphics[width=0.3\linewidth]{appendix/figs/landscape/purple_512x256.jpg} & {\footnotesize{}}
	\includegraphics[width=0.3\linewidth]{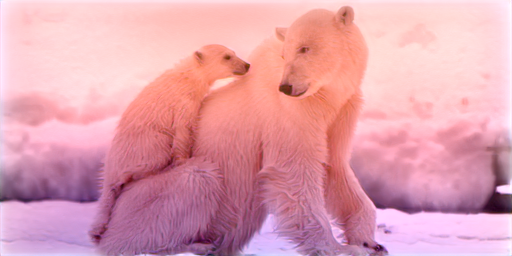}
	\tabularnewline
		
	\includegraphics[width=0.3\linewidth]{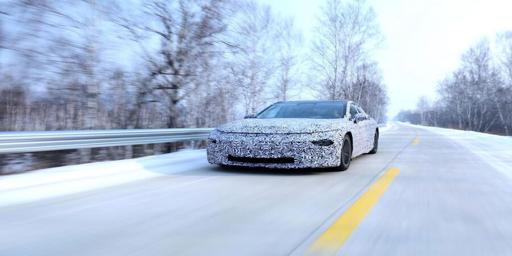} &
	{\footnotesize{}} \includegraphics[width=0.3\linewidth]{appendix/figs/landscape/purple_512x256.jpg} & {\footnotesize{}}
	\includegraphics[width=0.3\linewidth]{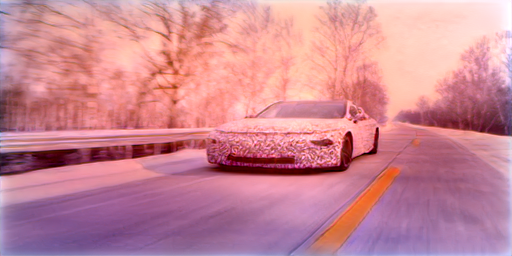}
	\tabularnewline
		
	\includegraphics[width=0.3\linewidth]{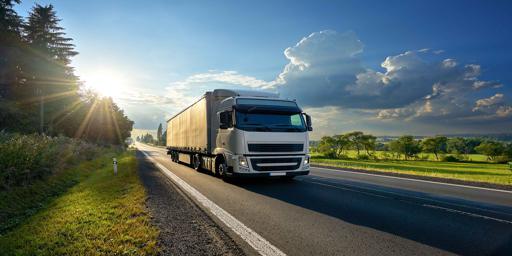} & {\footnotesize{}}
	\includegraphics[width=0.3\linewidth]{appendix/figs/landscape/driving_512x256.jpg} &
	{\footnotesize{}} 
	\includegraphics[width=0.3\linewidth]{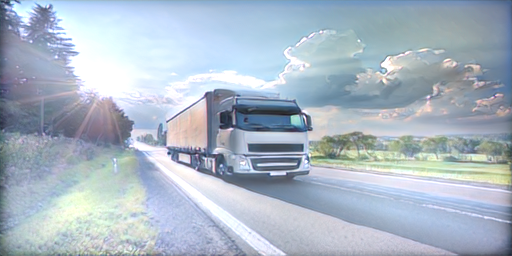}
	\tabularnewline
	
	\includegraphics[width=0.3\linewidth]{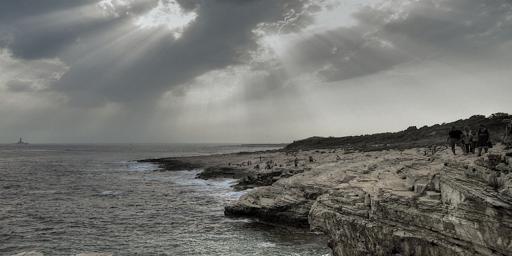} &
	{\footnotesize{}} \includegraphics[width=0.3\linewidth]{appendix/figs/landscape/green_512x256.jpg} & {\footnotesize{}}
	\includegraphics[width=0.3\linewidth]{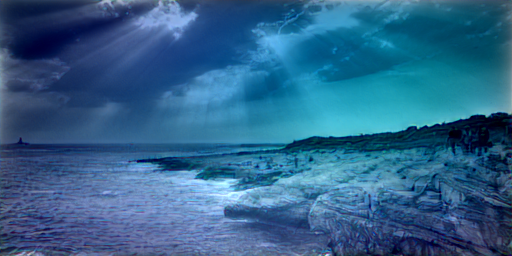} 
	\tabularnewline
	
	\includegraphics[width=0.3\linewidth]{appendix/figs/landscape/green_512x256.jpg} & {\footnotesize{}}
    \includegraphics[width=0.3\linewidth]{appendix/figs/landscape/gray_512x256.jpg} &
	{\footnotesize{}}
	\includegraphics[width=0.3\linewidth]{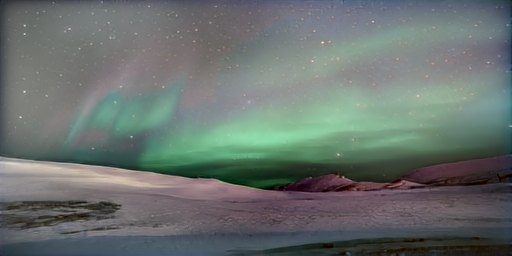} 
	\tabularnewline		
\end{tabular}
\hfill{}
\par\end{centering}
\caption{
Style-mixing using web-crawled images, where
the generator and encoder are only trained on Cityscapes. Except for the content images of the first 2 rows, all the others are web-crawled images.
}
\label{fig:landscape-example}
\vspace{-1.5em}
\end{figure}

Besides, we consider BDD100K-Daytime as the source domain, ACDC and Dark Z\"urich as the unseen target domains. We report the quantitative results in \cref{tab:hrnet-bdd-daytime-da}. As BDD100K already covers different times of day and diverse weather conditions, we only use a subset, i.e., 2526 daytime images of BDD100K for training, to allow for a more representative domain generalization evaluation. In this case, we specifically report ACDC-Night performance, since only nighttime images are not included in the training set. {\ourstyle} still outperforms the other data augmentation methods on unseen domains, being coherent with the other experimental results.

\begin{figure*}[t]
\begin{centering}
\setlength{\tabcolsep}{0.0em}
\renewcommand{\arraystretch}{0}
\par\end{centering}
\begin{centering}
\hfill{}%
	\begin{tabular}{@{}c@{}c@{}c@{}c}
			\centering
		   &   &   & \vspace{0.01cm} \tabularnewline
	\includegraphics[width=0.241\textwidth,]{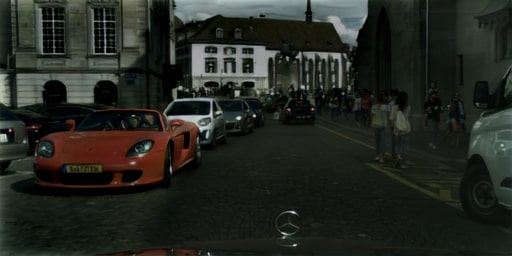} 
	        & {\footnotesize{}}
	    \includegraphics[width=0.241\textwidth,]{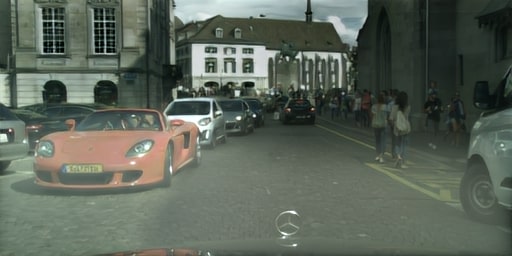} 
	        & {\footnotesize{}}
		\includegraphics[width=0.241\textwidth,]{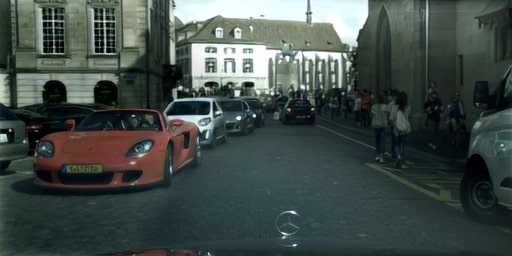} 
		    & {\footnotesize{}}
		\includegraphics[width=0.241\textwidth,]{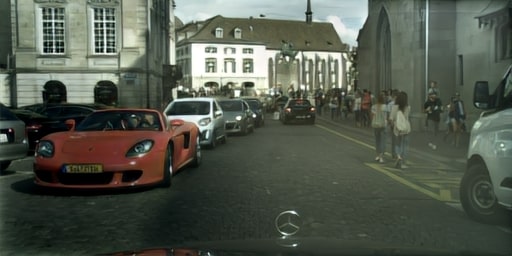} 
	 \tabularnewline
	\includegraphics[width=0.241\textwidth,]{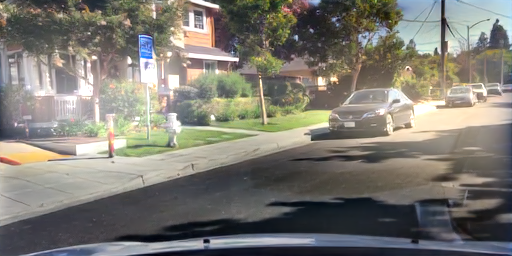} 
	        & {\footnotesize{}}
	   \includegraphics[width=0.241\textwidth,]{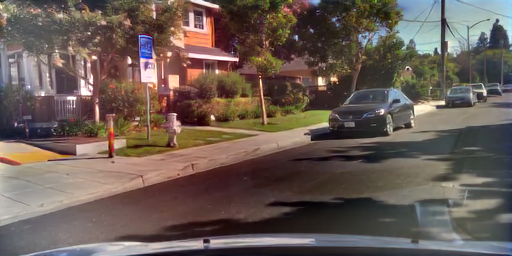} 
	   & {\footnotesize{}}
		\includegraphics[width=0.241\textwidth,]{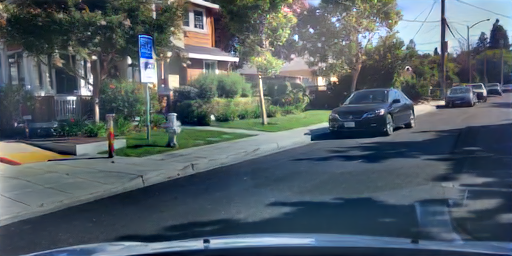} 
		    & {\footnotesize{}}
		\includegraphics[width=0.241\textwidth,]{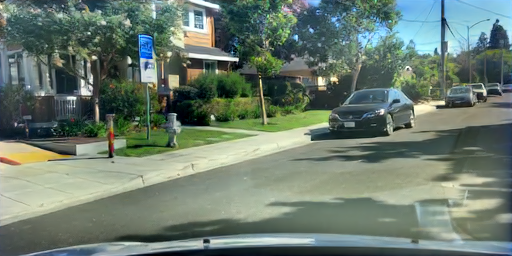} 
	 \tabularnewline
	 
	\includegraphics[width=0.241\textwidth,]{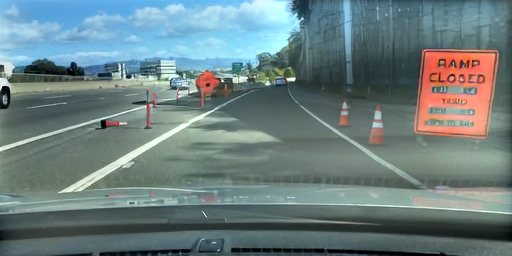} 
	        & {\footnotesize{}}
	   \includegraphics[width=0.241\textwidth,]{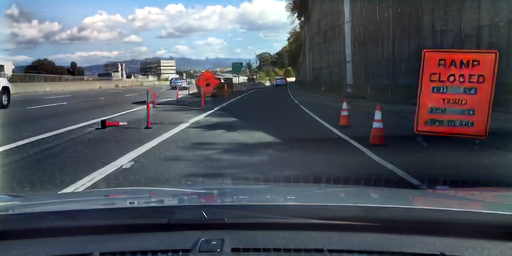} 
	   & {\footnotesize{}}
		\includegraphics[width=0.241\textwidth,]{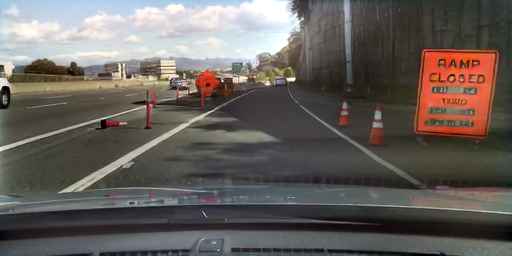} 
		    & {\footnotesize{}}
		\includegraphics[width=0.241\textwidth,]{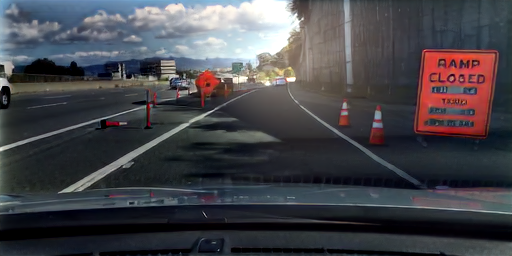} 
	 \tabularnewline
	 
	\includegraphics[width=0.241\textwidth,]{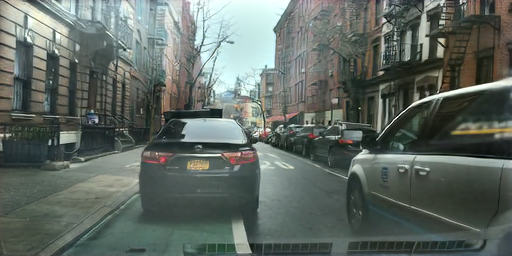} 
	  & {\footnotesize{}}
	   \includegraphics[width=0.241\textwidth,]{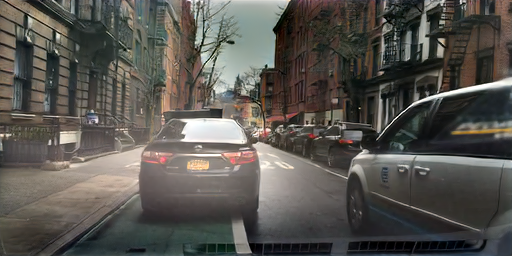} 
	   & {\footnotesize{}}
		\includegraphics[width=0.241\textwidth,]{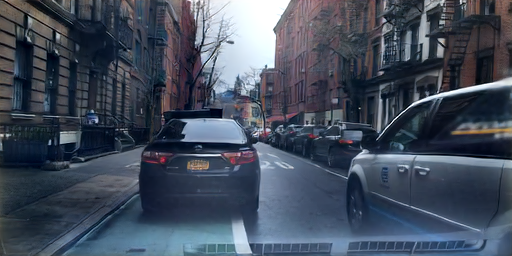} 
		    & {\footnotesize{}}
		\includegraphics[width=0.241\textwidth,]{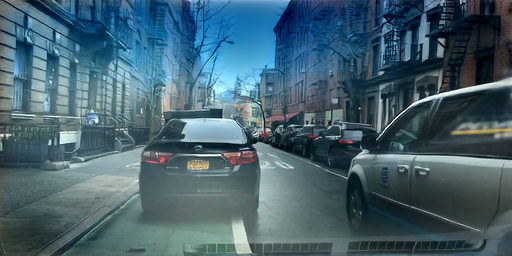} 
	 \tabularnewline
	 
	 \includegraphics[width=0.241\textwidth,]{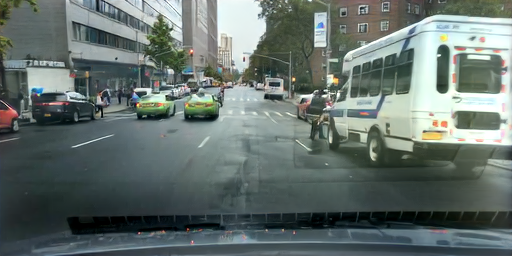} 
	  & {\footnotesize{}}
	   \includegraphics[width=0.241\textwidth,]{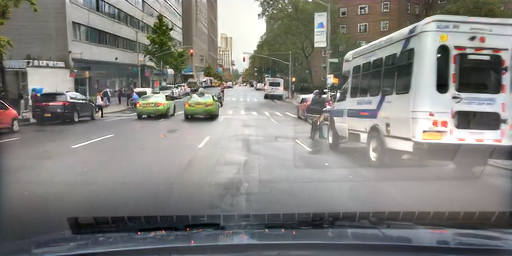}  
	   & {\footnotesize{}}
		\includegraphics[width=0.241\textwidth,]{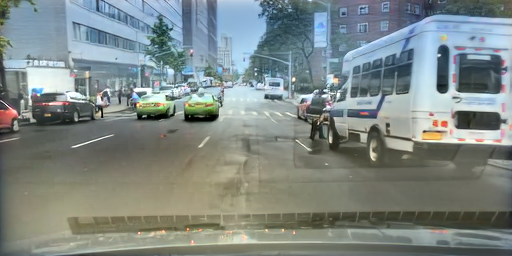} 
		    & {\footnotesize{}}
		\includegraphics[width=0.241\textwidth,]{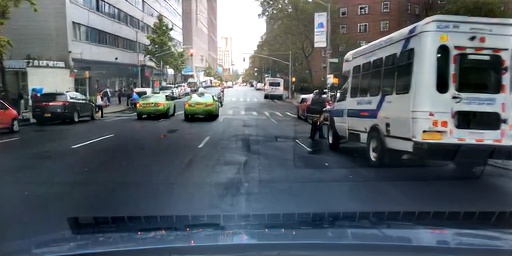} 
	 \tabularnewline
	 
	 \includegraphics[width=0.241\textwidth,]{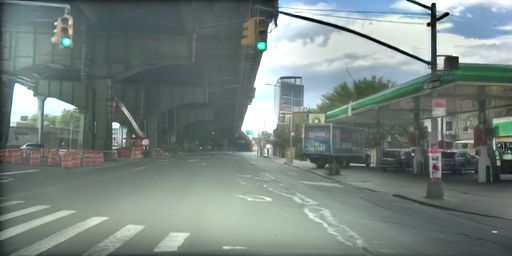} 
	  & {\footnotesize{}}
	   \includegraphics[width=0.241\textwidth,]{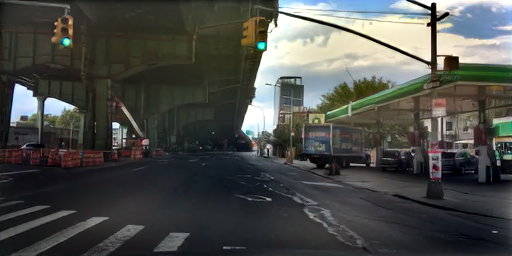}  
	   & {\footnotesize{}}
		\includegraphics[width=0.241\textwidth,]{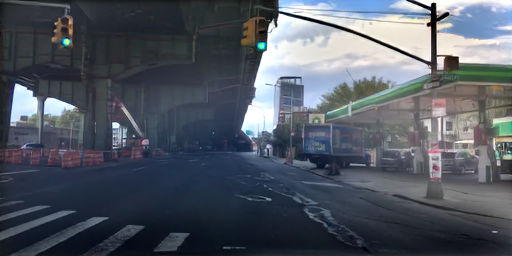}
		    & {\footnotesize{}}
		\includegraphics[width=0.241\textwidth,]{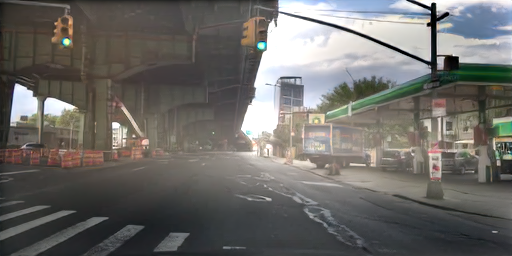}
	 \tabularnewline
	\end{tabular}
\hfill{}
\par\end{centering}
\caption{Examples of augmented images by our intra-source style augmentation ({\ourstyle}).
Each row presents randomly stylized samples of the same content using  {\ourstyle}, where both content and styles come from the source domain only, i.e., Cityscapes for the 1st row and BDD100K-Daytime for the remaining rows.  } 
\label{fig:intra-mix-bdd-appendix}
\end{figure*}

\begin{figure*}[t]
\begin{centering}
\setlength{\tabcolsep}{0.0em}
\renewcommand{\arraystretch}{0}
\par\end{centering}
\begin{centering}
\hfill{}%
	\begin{tabular}{@{}c@{}c@{}c@{}c}
			\centering
		 Image  & Ground truth  & Baseline & Ours \vspace{0.01cm} \tabularnewline
	\includegraphics[width=0.241\textwidth,height=0.1205\textwidth,]{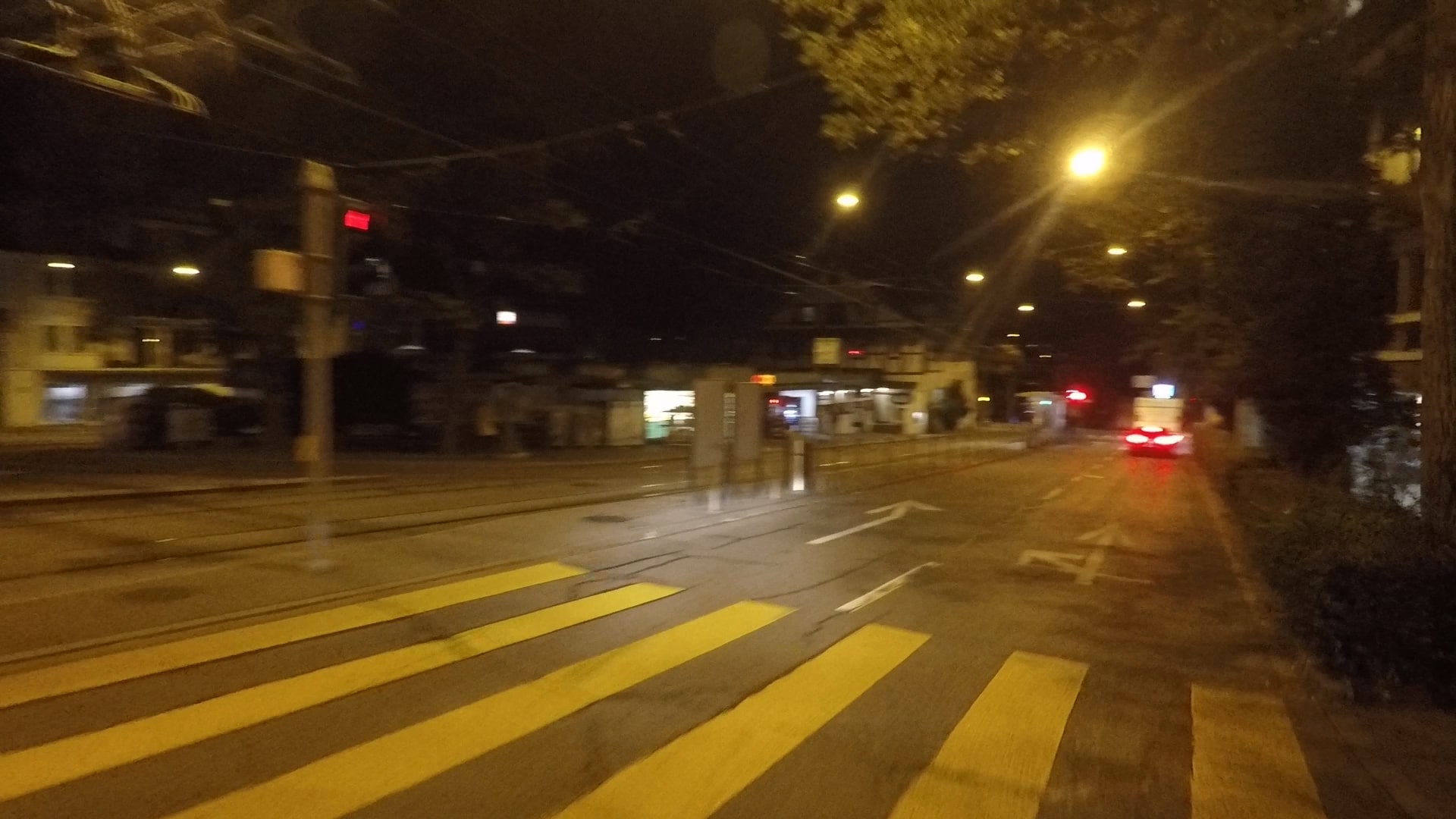} 
	        & {\footnotesize{}}
	  \begin{tikzpicture}
            \node [
	        above right,
	        inner sep=0] (image) at (0,0) {\includegraphics[width=0.241\textwidth,height=0.1205\textwidth]{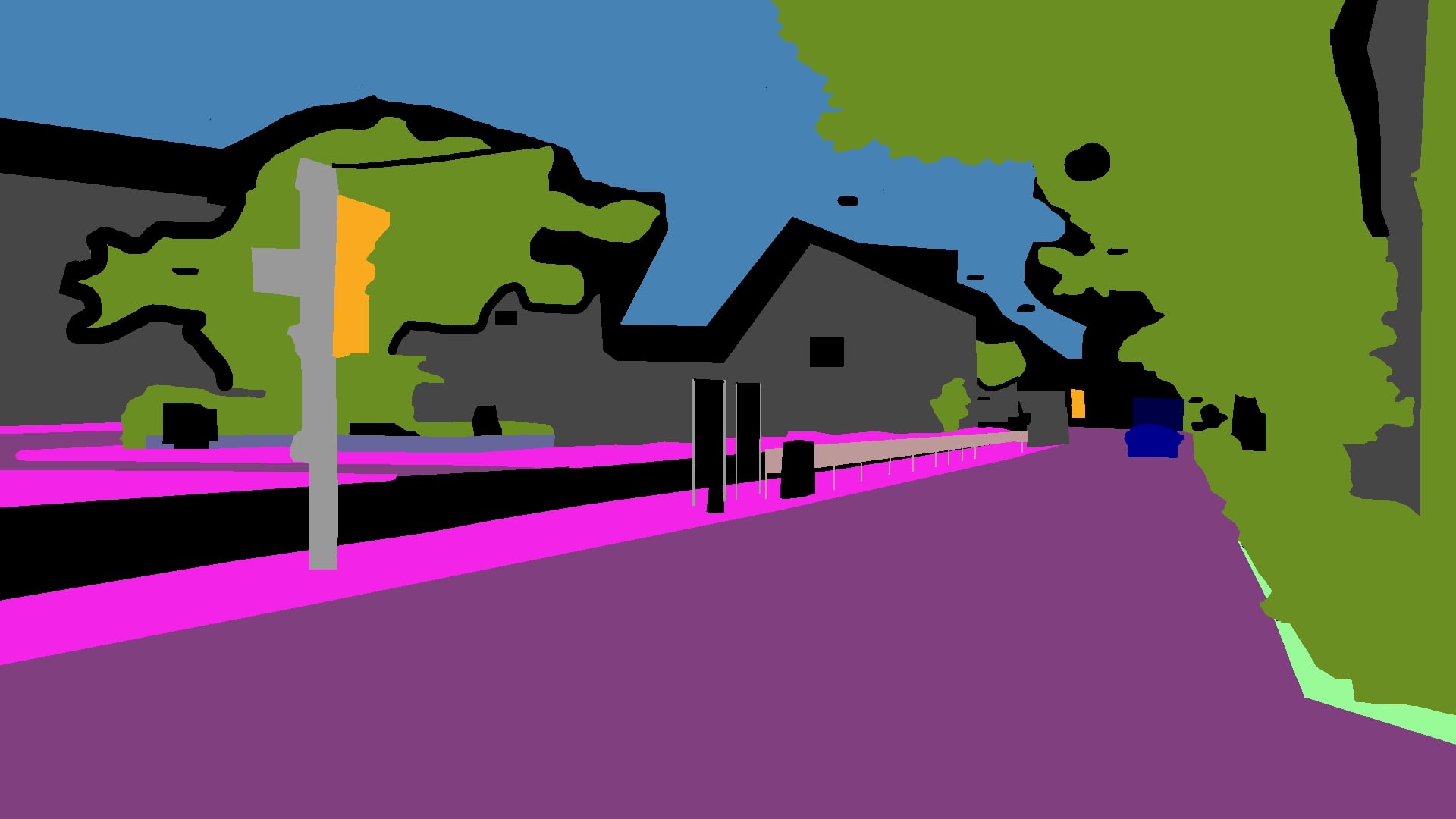} };
            \begin{scope}[
            x={($0.1*(image.south east)$)},
            y={($0.1*(image.north west)$)}]
            \draw[thick,green] (1.5,2.5) rectangle (3,8.4) ;
        \end{scope}
    \end{tikzpicture}
	        & {\footnotesize{}}
	  \begin{tikzpicture}
            % Include the image in a node
            \node [
	        above right,
	        inner sep=0] (image) at (0,0) {\includegraphics[width=0.241\textwidth,]{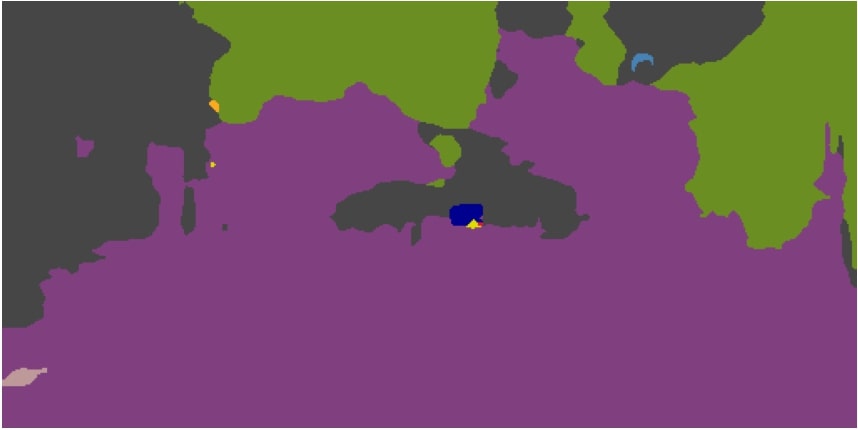} };
            % Create scope with normalized axes
            \begin{scope}[
            x={($0.1*(image.south east)$)},
            y={($0.1*(image.north west)$)}]
            \draw[thick,red] (1.5,2.5) rectangle (3,8.4) ;
        \end{scope}
    \end{tikzpicture}
		    & {\footnotesize{}}
    \begin{tikzpicture}
            % Include the image in a node
            \node [
	        above right,
	        inner sep=0] (image) at (0,0) {\includegraphics[width=0.241\textwidth,]{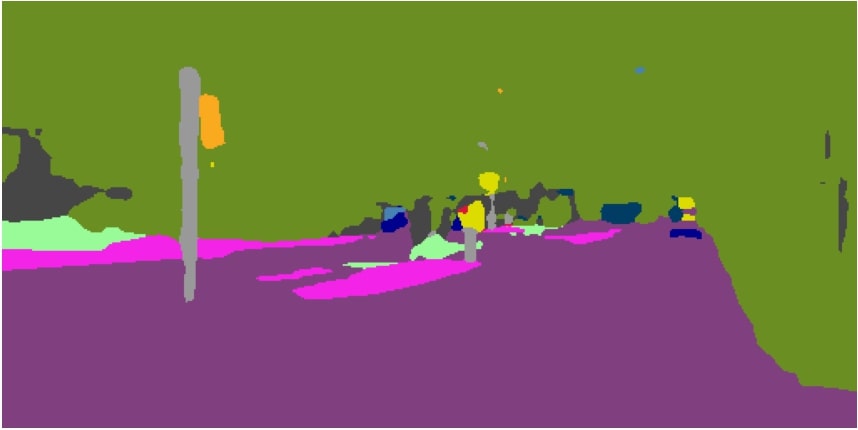}  };
            % Create scope with normalized axes
            \begin{scope}[
            x={($0.1*(image.south east)$)},
            y={($0.1*(image.north west)$)}]
            \draw[thick,green] (1.5,2.5) rectangle (3,8.4) ;
        \end{scope}
    \end{tikzpicture}
	 \tabularnewline	
	 
	 	\includegraphics[width=0.241\textwidth,height=0.1205\textwidth]{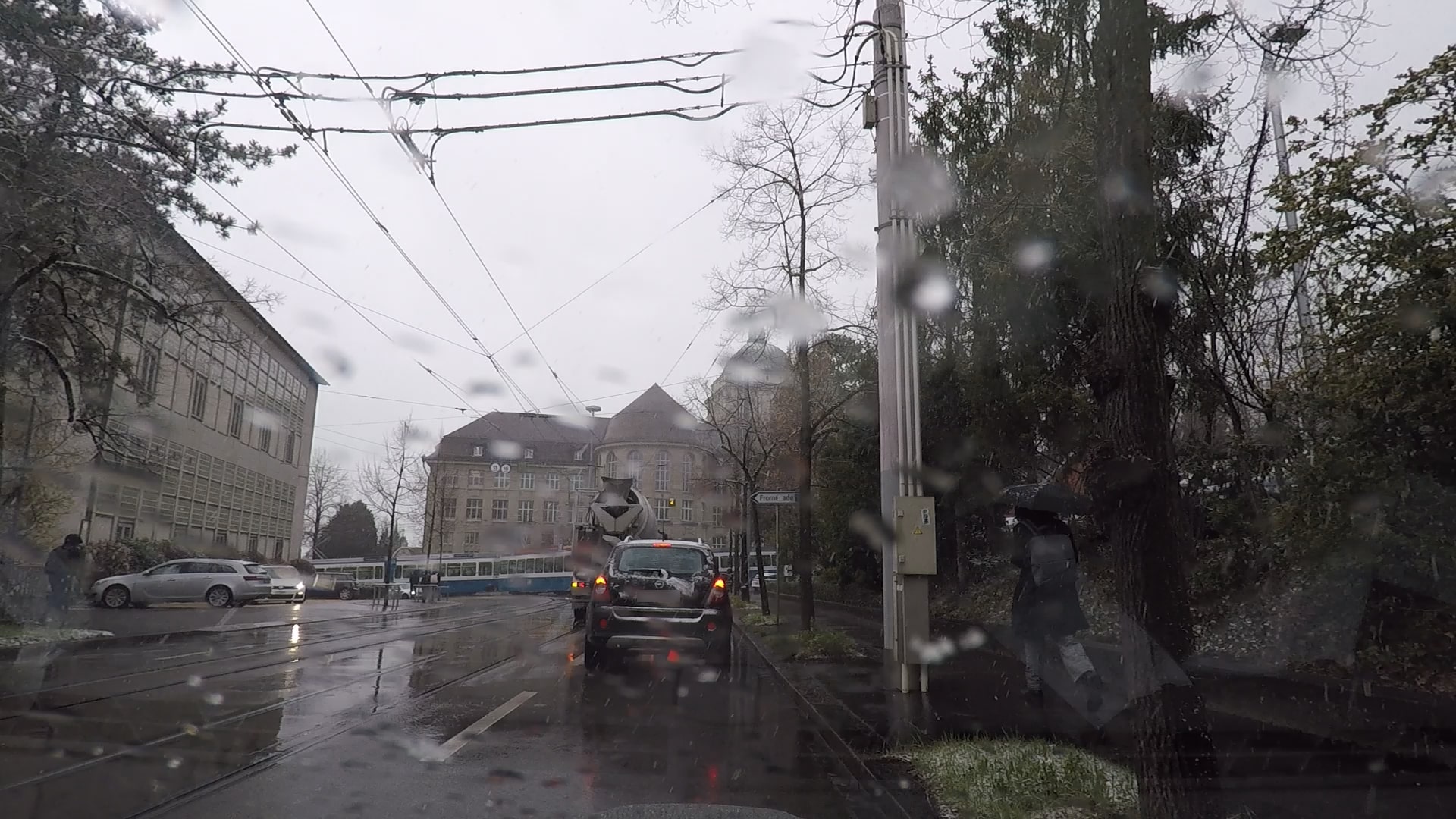} 
	        & {\footnotesize{}}
	  \begin{tikzpicture}
            % Include the image in a node
            \node [
	        above right,
	        inner sep=0] (image) at (0,0) {\includegraphics[width=0.241\textwidth,height=0.1205\textwidth]{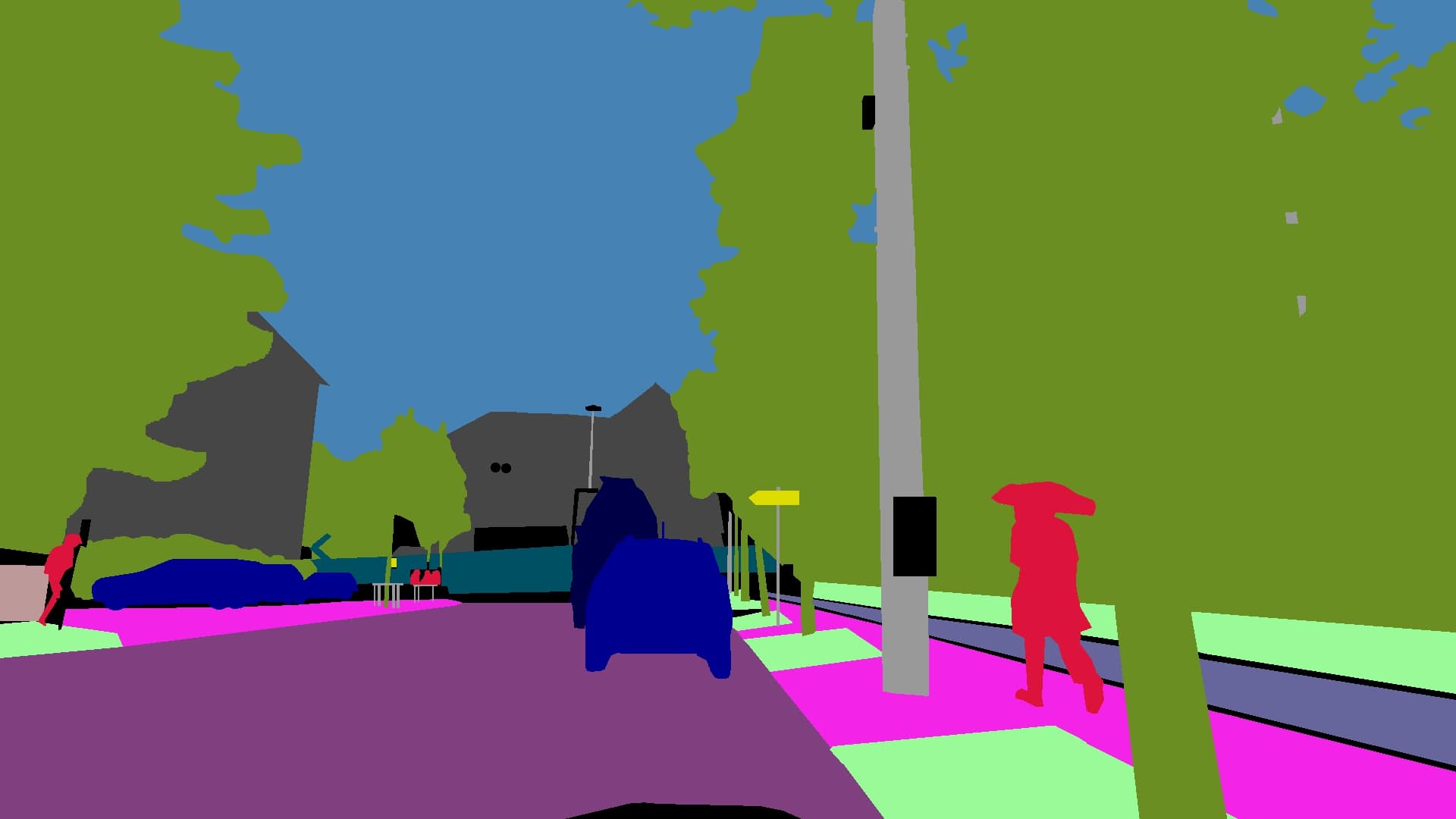}};
            % Create scope with normalized axes
            \begin{scope}[
            x={($0.1*(image.south east)$)},
            y={($0.1*(image.north west)$)}]
            \draw[thick,green] (6.5,1.0) rectangle (8,4.5) ;
        \end{scope}
    \end{tikzpicture}
	   & {\footnotesize{}}
	   \begin{tikzpicture}
            % Include the image in a node
            \node [
	        above right,
	        inner sep=0] (image) at (0,0) {\includegraphics[width=0.241\textwidth,]{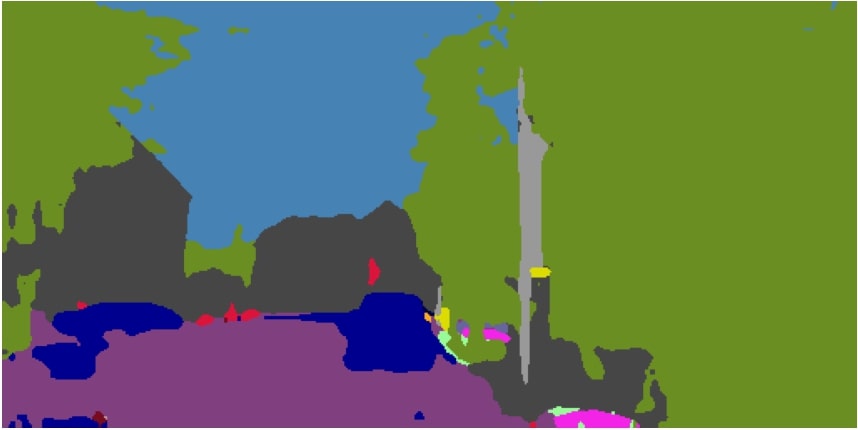}  };
            % Create scope with normalized axes
            \begin{scope}[
            x={($0.1*(image.south east)$)},
            y={($0.1*(image.north west)$)}]
            \draw[thick,red] (6.5,1.0) rectangle (8,4.5) ;
        \end{scope}
    \end{tikzpicture}
		    & {\footnotesize{}}
	\begin{tikzpicture}
            % Include the image in a node
            \node [
	        above right,
	        inner sep=0] (image) at (0,0) {\includegraphics[width=0.241\textwidth,]{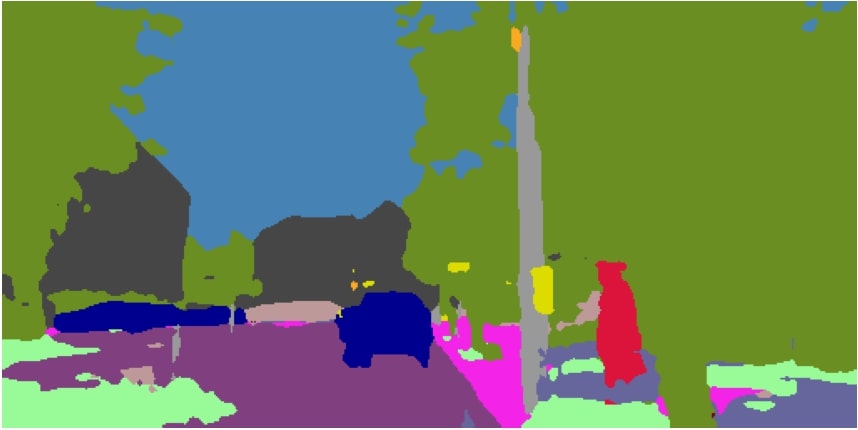}   };
            % Create scope with normalized axes
            \begin{scope}[
            x={($0.1*(image.south east)$)},
            y={($0.1*(image.north west)$)}]
            \draw[thick,green] (6.5,1.0) rectangle (8,4.5) ;
        \end{scope}
    \end{tikzpicture}
	 \tabularnewline
	 
	\includegraphics[width=0.241\textwidth,height=0.1205\textwidth]{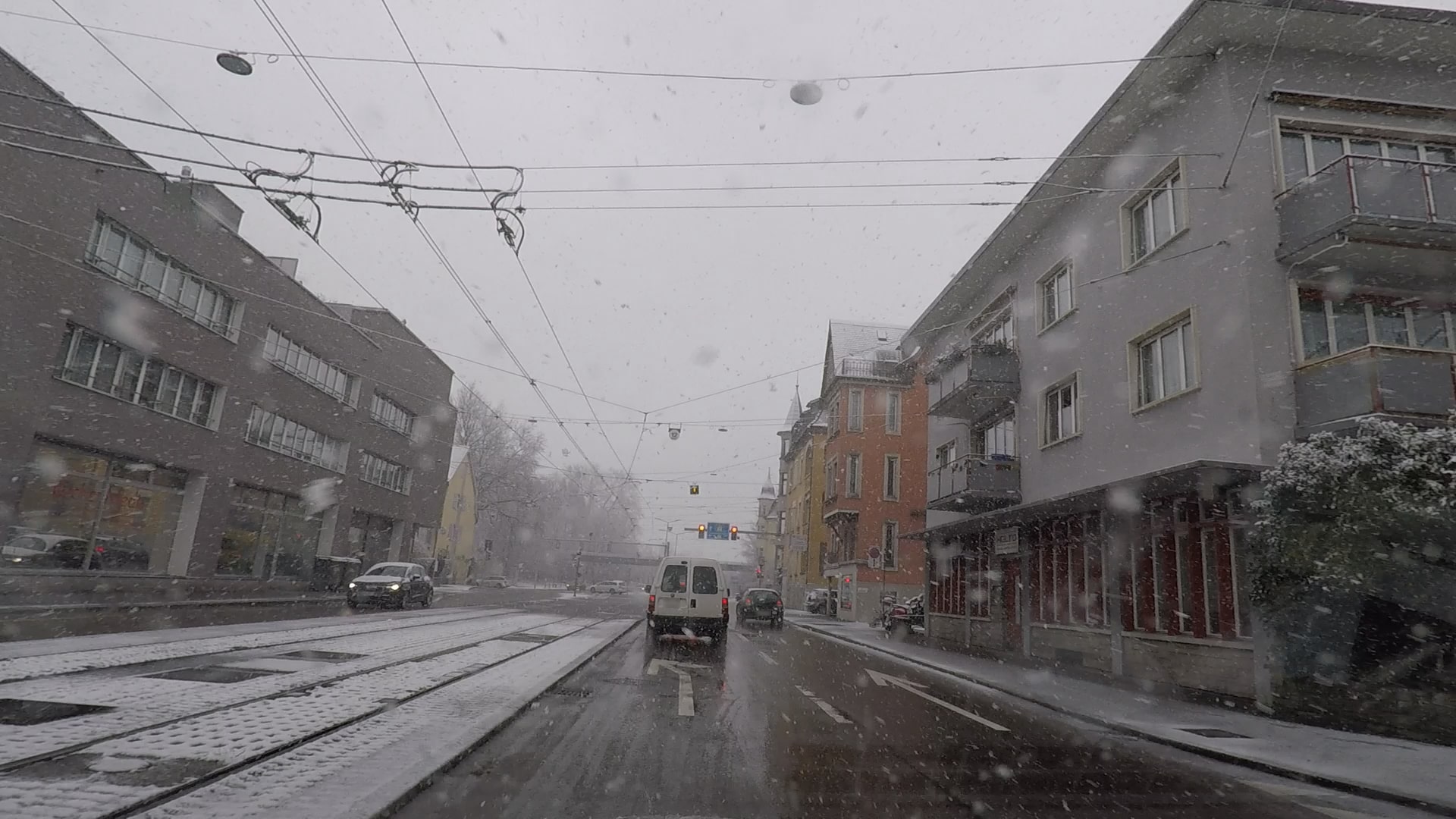} 
	        & {\footnotesize{}}
	   \begin{tikzpicture}
            % Include the image in a node
            \node [
	        above right,
	        inner sep=0] (image) at (0,0) {\includegraphics[width=0.241\textwidth,height=0.1205\textwidth]{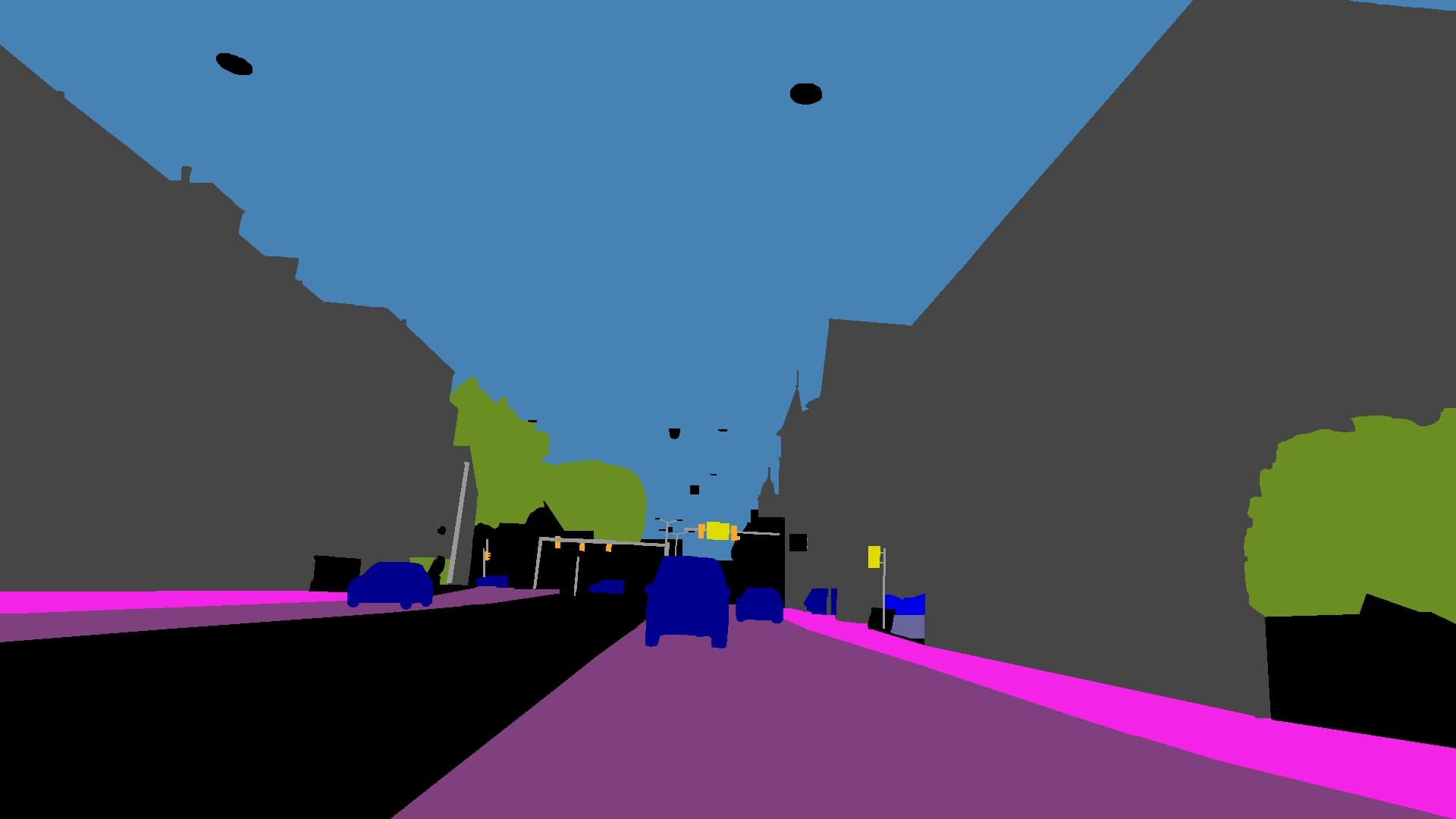}};
            % Create scope with normalized axes
            \begin{scope}[
            x={($0.1*(image.south east)$)},
            y={($0.1*(image.north west)$)}]
           \draw[thick,green] (1.1,1.8) rectangle (4.5, 9.2) ;
        \end{scope}
    \end{tikzpicture}
	   & {\footnotesize{}}
	  \begin{tikzpicture}
            % Include the image in a node
            \node [
	        above right,
	        inner sep=0] (image) at (0,0) {\includegraphics[width=0.241\textwidth,]{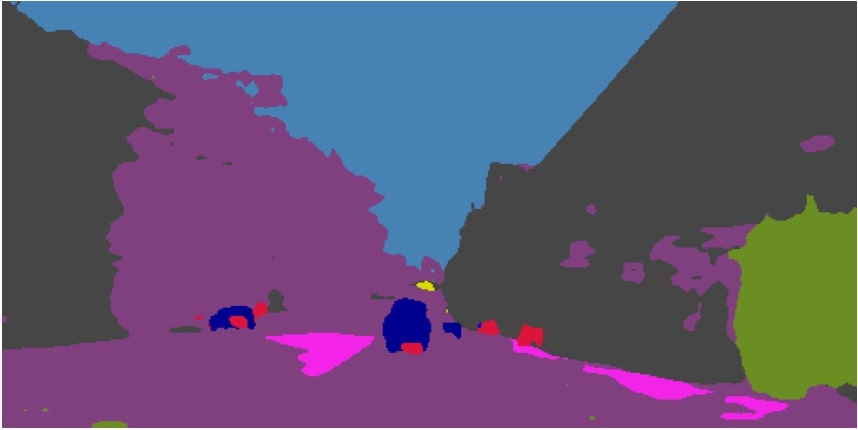}  };
            % Create scope with normalized axes
            \begin{scope}[
            x={($0.1*(image.south east)$)},
            y={($0.1*(image.north west)$)}]
            % Grid
	        %\draw[lightgray,step=0.5] (image.south west) grid (image.north east);
            % Axes' labels
        	%\foreach \x in {0,1,...,10} { \node [below] at (\x,0) {\x}; }
        	%\foreach \y in {0,1,...,10} { \node [left] at (0,\y) {\y};}
            % Labels
            \draw[thick,red] (1.1,1.8) rectangle (4.5, 9.2) ;
        \end{scope}
    \end{tikzpicture}
		    & {\footnotesize{}}
	\begin{tikzpicture}
            % Include the image in a node
            \node [
	        above right,
	        inner sep=0] (image) at (0,0) {\includegraphics[width=0.241\textwidth,]{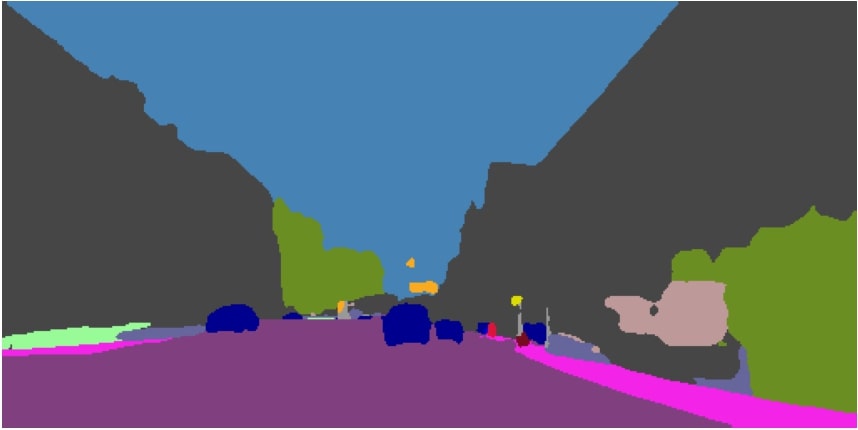} };
            % Create scope with normalized axes
            \begin{scope}[
            x={($0.1*(image.south east)$)},
            y={($0.1*(image.north west)$)}]
           \draw[thick,green] (1.1,1.8) rectangle (4.5, 9.2) ;
        \end{scope}
    \end{tikzpicture}
	 \tabularnewline
	 
	 \includegraphics[width=0.241\textwidth,height=0.1205\textwidth]{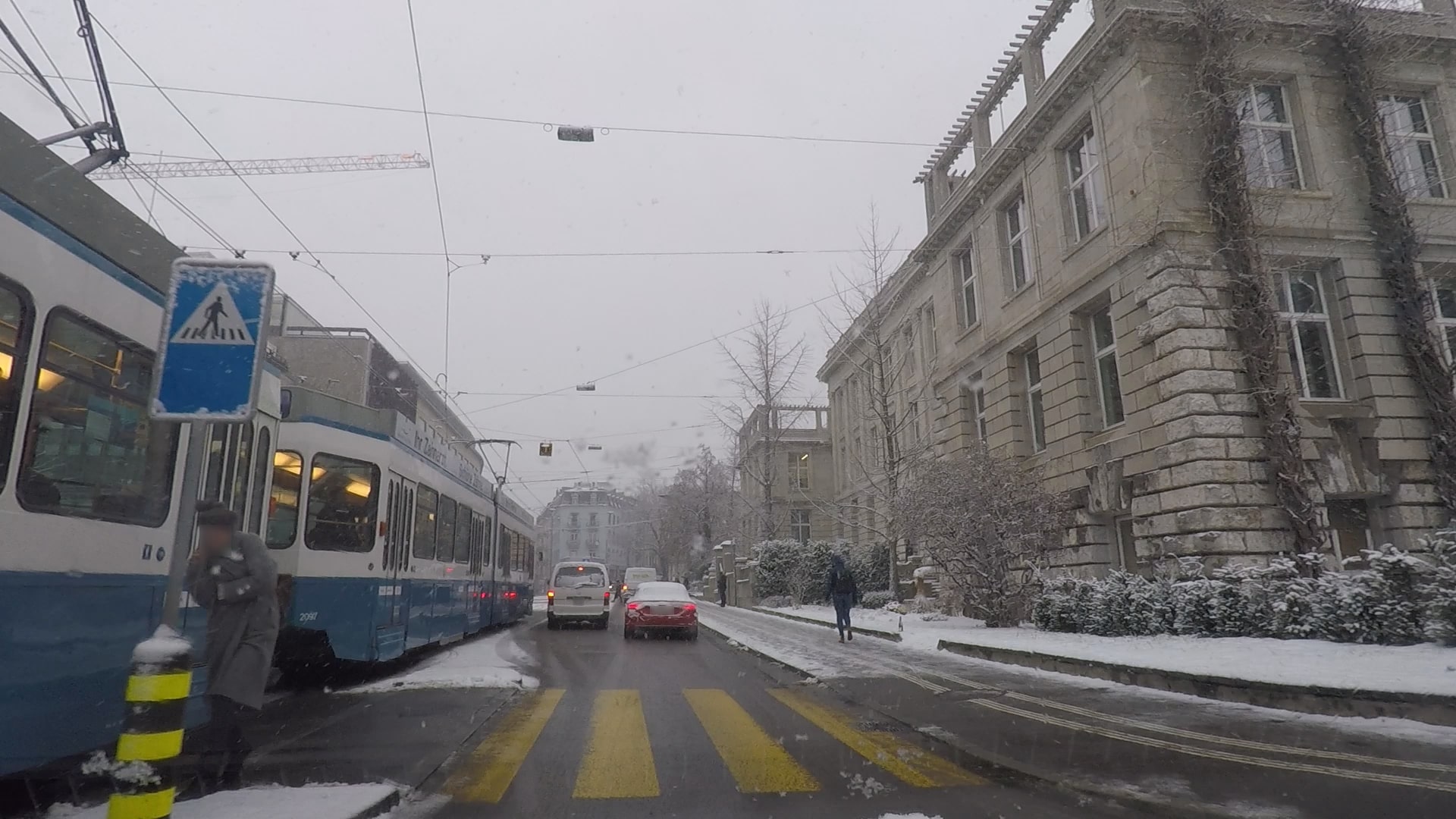} 
	        & {\footnotesize{}}
	        
	  \begin{tikzpicture}
            % Include the image in a node
            \node [
	        above right,
	        inner sep=0] (image) at (0,0) {\includegraphics[width=0.241\textwidth,height=0.1205\textwidth]{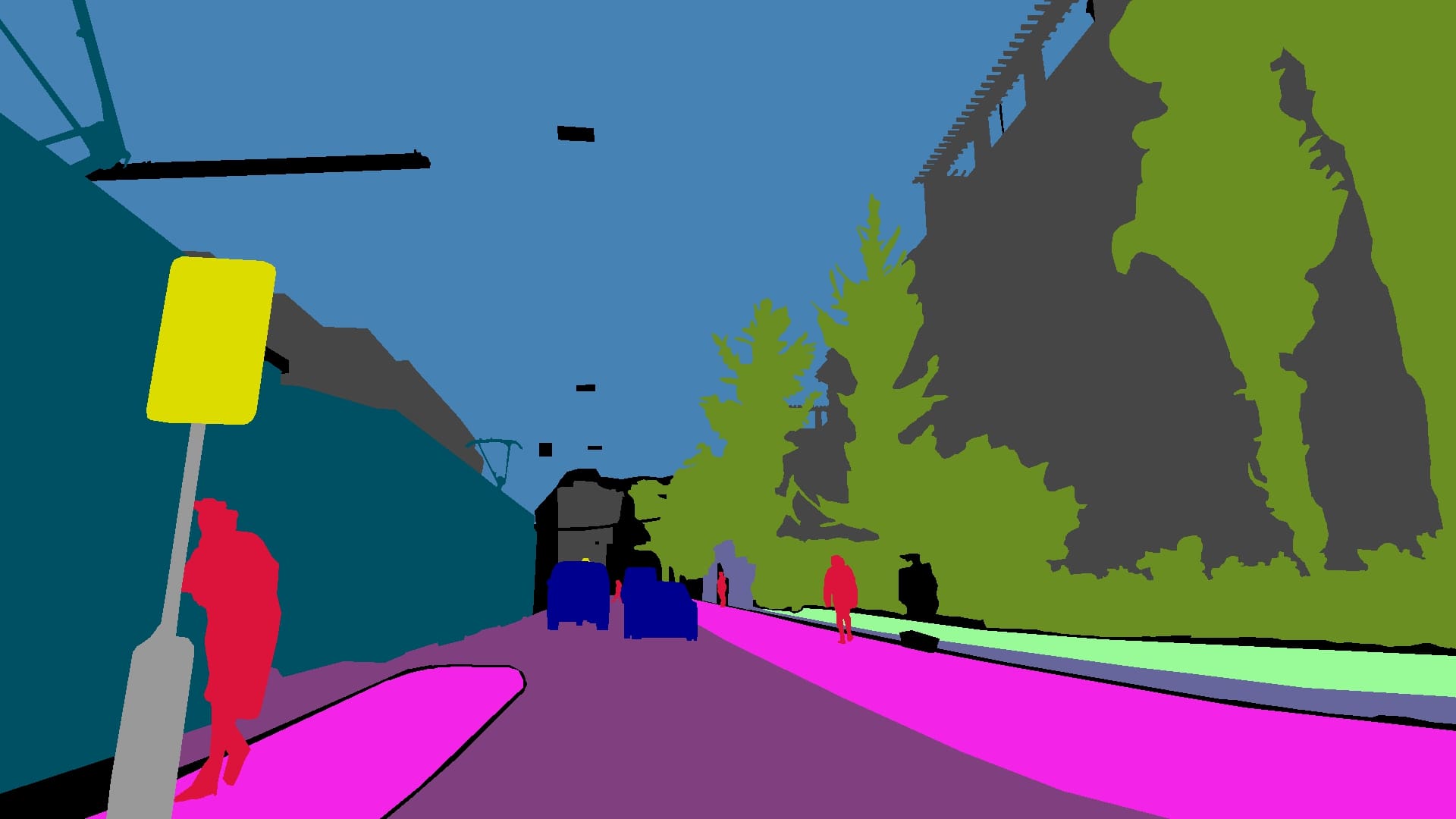}  };
            % Create scope with normalized axes
            \begin{scope}[
            x={($0.1*(image.south east)$)},
            y={($0.1*(image.north west)$)}]
            % Grid
	        %\draw[lightgray,step=0.5] (image.south west) grid (image.north east);
            % Axes' labels
        	%\foreach \x in {0,1,...,10} { \node [below] at (\x,0) {\x}; }
        	%\foreach \y in {0,1,...,10} { \node [left] at (0,\y) {\y};}
            % Labels
            \draw[thick,green] (0.5,0.3) rectangle (2.5, 7.2) ;
        \end{scope}
    \end{tikzpicture}      
	   & {\footnotesize{}}
	  \begin{tikzpicture}
            % Include the image in a node
            \node [
	        above right,
	        inner sep=0] (image) at (0,0) {\includegraphics[width=0.241\textwidth,]{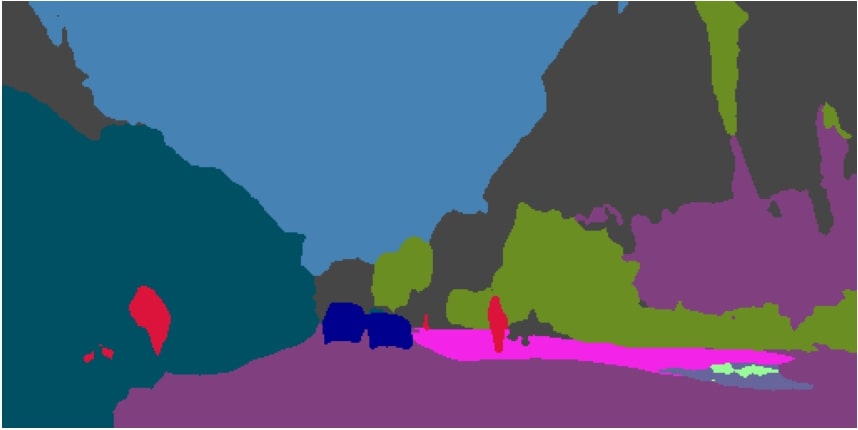}};
            % Create scope with normalized axes
            \begin{scope}[
            x={($0.1*(image.south east)$)},
            y={($0.1*(image.north west)$)}]
           \draw[thick,red] (0.5,0.3) rectangle (2.5, 7.2) ;
        \end{scope}
    \end{tikzpicture}
		    & {\footnotesize{}}
    \begin{tikzpicture}
            % Include the image in a node
            \node [
	        above right,
	        inner sep=0] (image) at (0,0) {	\includegraphics[width=0.241\textwidth,]{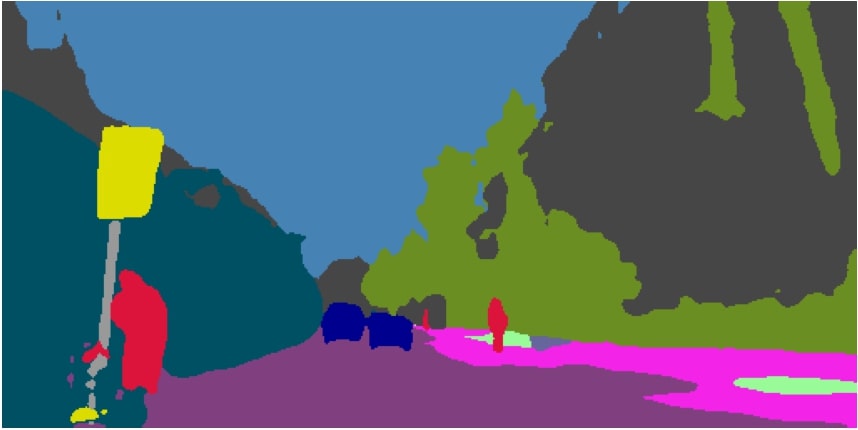}};
            % Create scope with normalized axes
            \begin{scope}[
            x={($0.1*(image.south east)$)},
            y={($0.1*(image.north west)$)}]
           \draw[thick,green] (0.5,0.4) rectangle (2.5, 7.3) ;
        \end{scope}
    \end{tikzpicture}
	 \tabularnewline
	 
	\end{tabular}
\hfill{}
\par\end{centering}
\caption{Semantic segmentation results of Cityscapes $\rightarrow$ ACDC generalization using HRNet. The HRNet is trained on Cityscapes only.} 
\label{fig:dg-semseg}
\end{figure*}
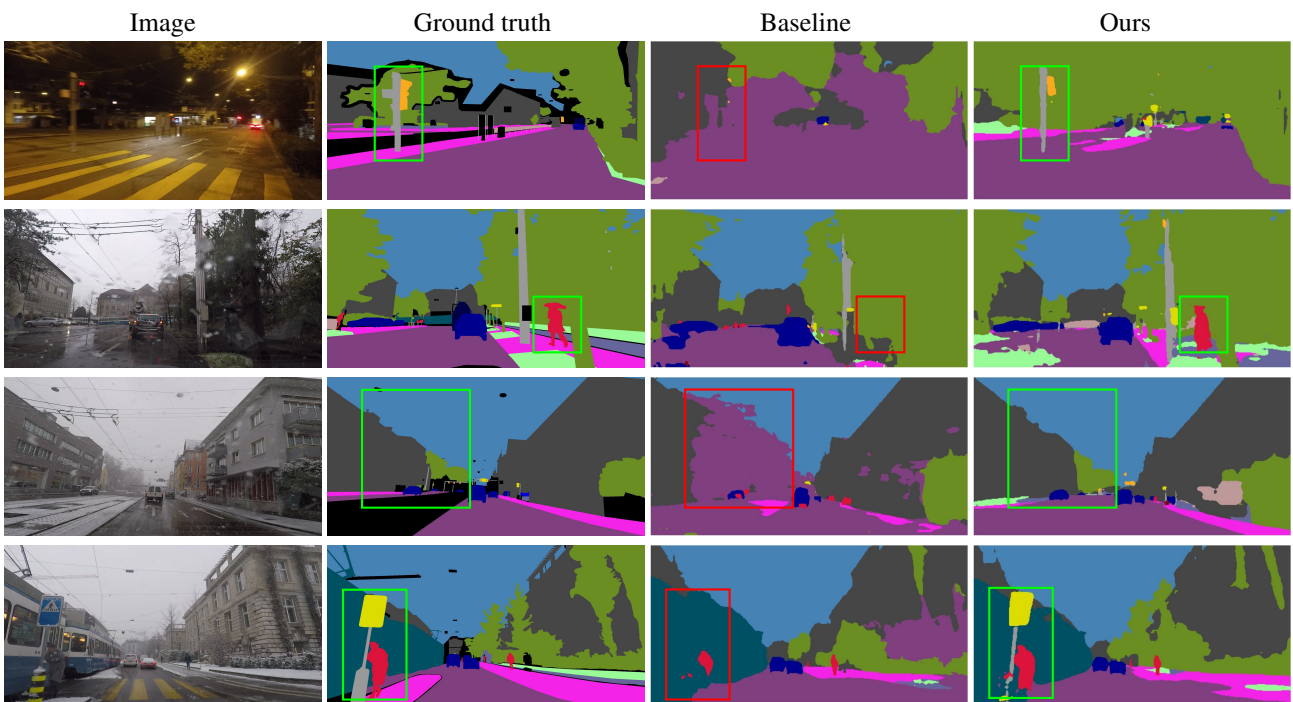

%%%%%%%%%%%%%%%%%%%%%%%%%%%%%%%%%%%
\paragraph{Qualitative results of \textbf{ISSA}}
We present visual examples of our {\ourstyle} in 
\cref{fig:intra-mix-bdd-appendix}. Images in each row have the same content with random styles extracted from the source domain, i.e., Cityscapes for the 1st row and BDD100K-Daytime for the remaining rows.
Besides, some qualitative semantics segmentation results on Cityscapes to ACDC generalization are demonstrated in \cref{fig:dg-semseg}.

%%%%%%%%%%%%%%%%%%%%%%%%%%%%%%%%%%%
\paragraph{Plug-n-play ability} 
\newchange{Training GAN and encoder could take considerable computational resources, therefore we investigate the plug-n-play ability of our pipeline. We observe that 
ISSA can still be effective even when encoder and generator are trained on a different dataset of a similar task, and re-training is not required. 
As shown in \cref{tab:bddgan-cityscapes-dg-small}, when training the segmenter on Cityscapes using ISSA, we can directly use generator and encoder trained on BDD100K without fine-tuning. The effectiveness of ISSA is not compromised even though the model has never seen Cityscapes samples. Visual examples in \cref{fig:landscape-example} show the plug-n-play style-mixing ability of our encoder
on web-crawled images, where the model is only trained on Cityscapes.
}

%%%%%%%%%%%%%%%%%%%%%%%%%%%%%%%%
\section{Comparison with Unsupervised Domain Adaptation Methods}\label{sec:appendix-uda}
We compare our method with multiple unsupervised domain adaptation (UDA) techniques, which not only have access to the source domain, but also use extra unlabeled samples of the target domain. The quantitative comparison of Cityscapes to ACDC adaptation/generalization is shown in \Cref{tab:uda-comparison}. Our method has presented competitive performance, even without using images from the target domain.

%%%%%%%%%%%%%%%%%%%%%%%%%%%%%%%%%%%%
\section{Limitations and Future Work} \label{sec:appendix-limitations}
One limitation of {\ourstyle} is that our style mixing is a global transformation, which cannot specifically alter the style of local objects, e.g., adjusting vehicle color from red to black, though when changing the image globally, local areas are inevitably modified. 

In the future, it is challenging yet interesting to extend our work with class-aware style mixing. Also, by exploiting the pre-trained language-vision model such as CLIP\newcite{radford2021clip}, we can synthesize styles conditioned on text rather than an image. For instance, by providing a text condition ``snowy road",  ideally we would want to obtain an image where there is snow on the road and other semantic classes remain unchanged.

\end{document}